\documentclass[10pt]{article}
\usepackage[english]{babel}
\usepackage[left=1in,right=1in,top=1in,bottom=1.5in]{geometry} 
\usepackage[skip=5pt, indent=10pt]{parskip}
\usepackage{amssymb,amsmath, amsthm}
\usepackage{hyperref}
\usepackage{xcolor}
\usepackage[margin=1cm]{caption}
\usepackage{natbib}
\usepackage[makeroom]{cancel}
\usepackage{bbm}
\usepackage{mathrsfs}
\usepackage{algorithm}
\usepackage{algpseudocode}
\usepackage{times}
\usepackage{subcaption}
\usepackage{multirow}
\usepackage{diagbox}
\usepackage{siunitx}

\usepackage{mathtools}
\mathtoolsset{showonlyrefs=true}

\definecolor{darkred}{HTML}{880000}
\definecolor{darkblue}{HTML}{000088}

\hypersetup{
  colorlinks,
  linkcolor={darkred},
  citecolor={darkblue}
}

\theoremstyle{plain}
\newtheorem{Th}{Theorem}
\newtheorem{Co}{Corollary}
\newtheorem{As}{Assumption}
\newtheorem{Lem}{Lemma}[section]
\newtheorem{Prop}{Proposition}

\theoremstyle{definition}

\renewcommand{\le}{\leqslant}
\renewcommand{\leq}{\leqslant}
\renewcommand{\ge}{\geqslant}
\renewcommand{\geq}{\geqslant}

\newcommand{\eps}{\varepsilon}

\def\argmin{\operatornamewithlimits{argmin}}

\newcommand{\dd}{{\mathrm{d}}}

\newcommand{\ED}{\mathrm{ED}}
\newcommand{\MMD}{\mathrm{MMD}}
\newcommand{\supp}{\mathrm{supp}}

\newcommand{\Tr}{\mathrm{Tr}}
\newcommand{\opterr}{\Delta_{\mathrm{opt}}}

\newcommand{\sfP}{\mathsf{P}}
\newcommand{\sfp}{\mathsf{p}}
\newcommand{\sfq}{\mathsf{q}}
\newcommand{\sfQ}{\mathsf{Q}}
\newcommand{\J}{{\mathsf{J}}}
\newcommand{\KL}{{\mathsf{KL}}}
\newcommand{\NN}{\mathsf{NN}}

\newcommand{\E}{\mathbb E}
\newcommand{\bbM}{\mathbb M}
\newcommand{\N}{\mathbb N}
\newcommand{\p}{\mathbb P}
\newcommand{\R}{\mathbb R}
\newcommand{\Z}{\mathbb Z}

\newcommand{\cA}{{\mathcal{A}}}
\newcommand{\cB}{{\mathcal{B}}}

\newcommand{\cE}{{\mathcal{E}}}
\newcommand{\cF}{{\mathcal{F}}}

\newcommand{\cK}{{\mathcal{K}}}
\newcommand{\cL}{{\mathcal{L}}}
\newcommand{\cM}{{\mathcal{M}}}
\newcommand{\cN}{{\mathcal{N}}}
\newcommand{\cO}{{\mathcal{O}}}
\newcommand{\cR}{{\mathcal{R}}}
\newcommand{\cT}{{\mathcal{T}}}
\newcommand{\cV}{{\mathcal{V}}}

\newcommand{\m}{{\mathfrak m}}
\newcommand{\z}{{\mathfrak z}}

\newcommand{\ttA}{\mathtt A}
\newcommand{\ttb}{\mathtt b}
\newcommand{\ttB}{\mathtt B}

\newcommand{\ttT}{\mathtt T}
\newcommand{\ttv}{\mathtt v}

\newcommand{\relu}{\mathrm{ReLU}}

\newcommand{\myendproof}{\hfill$\square$}

\title{On accuracy of density estimation in generative models\\based on stochastic optimal control}

\author{%
Nikita Puchkin\thanks{HSE University, Russian Federation}
\and
Denis Suchkov\footnotemark[1]
\and
Alexey Naumov\footnotemark[2]
\and
Denis Belomestny\thanks{Duisburg-Essen University, Germany}\hspace{0.16cm}
}

\date{}

\begin{document}

\maketitle

\begin{abstract}
    We study the statistical accuracy of Schr\"odinger bridge models based on an Ornstein--Uhlenbeck reference process. Given a known, compactly supported initial distribution and independent samples from an unknown target distribution, we estimate a right Schr\"odinger log-potential by minimizing empirical Kullback--Leibler (KL) risk over a prescribed function class. For bounded sub-Gaussian target densities, potentially with unbounded support, we establish a non-asymptotic high-probability oracle inequality for the KL-divergence between the target and estimated terminal marginals. Under additional two-sided Gaussian bounds and smoothness assumptions on the target log-density, we derive approximation guarantees for sparse ReLU networks and establish a KL convergence rate of $\widetilde{\mathcal O}(n^{-2s / (2s + d)})$ for exact empirical risk minimization. This matches, up to logarithmic factors, the classical nonparametric KL-risk convergence exponent, while allowing target densities on $\mathbb R^d$ whose tails are controlled by Gaussian envelopes.
\end{abstract}

\section{Introduction}

Generative modelling methods based on discretization of SDEs \citep{song19, song21} and ODEs \citep{lipman23, liu23} have recently gained considerable attention due to their ability to produce high-quality and diverse images. Huge empirical success of these models
fostered substantial methodological advances. For this reason, the Schr\"odinger bridge problem \citep{Schrodinger1932} attracted the attention of many machine learners and statisticians (for instance, \citep{Tzen2019, DeBortoli2021, Shi2023, Korotin2024, Gushchin2024a, Rapakoulias2024} to name a few). In its original form, given a reference process $\{X_t^0 : 0 \leq t \leq T\}$ (also called prior or base process) and two distributions $\sfP_0$ and $\sfP_T^*$, one seeks a random process $\{X_t^* : 0 \leq t \leq T\}$ such that $X_0^* \sim \sfP_0$, $X_T^* \sim \sfP_T^*$, and
the Kullback--Leibler (KL) divergence between the path measures of $\{X_t^* : 0 \leq t \leq T\}$ and $\{X_t^0 : 0 \leq t \leq T\}$ is as small as possible. This problem can be interpreted as the most likely evolution of a probability distribution between two given endpoint marginals. The standard candidate for the reference dynamics is the Wiener process $\{W_t : 0 \leq t \leq T\}$. This choice is usually motivated by the connections of the Schr\"odinger bridge problem with entropic optimal transport (see, for instance, \cite{Leonard2014}). However, in our view, the non-ergodicity and long-term memory of the initial distribution render the Wiener process a suboptimal choice for generative modelling. Instead, we look at the Schr\"odinger bridge problem through the lens of stochastic optimal control \citep{Daipra1991} and consider the Ornstein--Uhlenbeck process
\begin{equation}
    \label{eq:ou_process}
    \dd X_t^0 = -b \left( X_t^0 - m \right) \dd t + \Sigma^{1/2} \dd W_t, \quad 0 < t < T,
\end{equation}
as the base one. Here $b > 0$ controls the drift rate, $m \in \mathbb{R}^d$ represents the mean-reversion level, and $\Sigma \in \mathbb{R}^{d \times d}$ is a positive definite symmetric matrix, allowing us to take data anisotropy into account. It is known that the conditional distribution of $X_t^0$ given $X_0^0 = x$ is Gaussian $\cN\big(m_t(x), \Sigma_t \big)$ with
\begin{equation}
    \label{eq:m_t_sigma_t}
    m_t(x) = (1 - e^{-bt}) m + e^{-bt} x
    \quad \text{and} \quad
    \Sigma_t = \sigma_t^2 \, \Sigma,
    \quad \text{where} \quad
    \sigma_t^2 = \frac{1 - e^{-2bt}}{2b}.
\end{equation}
This implies geometric ergodicity of the reference process \eqref{eq:ou_process}.

In what follows, we assume for simplicity that both $\sfP_0$ and $\sfP_T^*$ are absolutely continuous with the densities $\rho_0$ and $\rho_T^*$, respectively.
To construct the solution for the Schr\"odinger bridge problem, \cite{Daipra1991} uses the observation of \cite{Beurling1960} (see also Theorem 2.2 in \cite{Daipra1991}) that there exist unique (up to multiplicative constants) non-negative measurable functions $\nu_0$ and $\nu_T$ such that the coupling
\begin{equation}
    \label{schroe1}
    \pi(x, y) = \nu_0(x) \, \sfq_T(y \,|\, x) \, \nu_T(y),
\end{equation}
where
\begin{equation}
    \label{eq:ou_transition_density}
    \sfq_t(y \,\vert\, x) = \frac{1}{(2\pi)^{d/2} \sqrt{\det(\Sigma_t)}} \exp\left\{-\frac12 \left\|\Sigma_t^{-1/2} \big(y - m_t(x) \big) \right\|^2 \right\},
    \quad 0 < t \leq T,
\end{equation}
is the transition density of the process \eqref{eq:ou_process}, has the marginals $\rho_0(x)$ and $\rho_T^*(y)$. The functions $\nu_0$ and $\nu_T$ are referred to as left and right \emph{Schr\"odinger potentials}, respectively. Based on this fact, under some additional technical assumptions, \cite{Daipra1991} deduces that the process
\begin{equation}
    \label{eq:opt_control_sde}
    \dd X_t^* = -b (X_t^* - m) \, \dd t + \Sigma \nabla \log \cT_{T - t}[\nu_T](X_t^*) \, \dd t + \Sigma^{1/2} \, \dd W_t,
    \quad
    0 < t < T,
\end{equation}
where
\[
    \cT_{T - t}[\nu_T](x) = \int\limits_{\R^d} \sfq_{T - t}(y \,\vert\, x) \, \nu_T(y) \, \dd y,
    \quad x \in \R^d,
\]
and $X_0^* \sim \rho_0$,
solves the minimal action problem, and, in view of Proposition 6 from \cite{Leonard2014}, the Schr\"odinger bridge problem as well\footnote{A reader is also encouraged to check Theorem 2.1 from \cite{Daipra1991} and Theorem 2.6 from \cite{Leonard2014} to ensure that the transition kernel of the process \eqref{eq:opt_control_sde} coincides with the one of the solution of the Schr\"odinger bridge problem.}. In other words, the dynamics $\{X_t^* : 0 \leq t \leq T\}$ is completely determined by the right Schr\"odinger potential $\nu_T$.

In the present paper, we are interested in estimation of the Schr\"odinger log-potential $\varphi^*(y) = \log \nu_T(y)$ from $n$ i.i.d. samples $Y_1, \dots, Y_n \sim \rho_T^*$. With an estimate $\widehat\varphi$ of $\varphi^*$ at hand, one can consider the corresponding process
\begin{equation}
    \label{eq:sde_estimate}
    \widehat X_0 \sim \rho_0,
    \quad
    \dd \widehat X_t = -b\left(\widehat X_t - m\right) \dd t + \Sigma \nabla \log \cT_{T - t}\big[e^{\widehat\varphi}\big](\widehat X_t) \, \dd t + \Sigma^{1/2} \, \dd W_t,
    \quad
    0 < t < T,
\end{equation}
and, using SDE discretization schemes or the probability flow ODE technique (see Appendix D in \cite{song21}), produce new samples from a distribution, which is close to the target one. In general, for any measurable function $\varphi : \R^d \rightarrow \R$, such that $\cT_T[e^\varphi](x) < +\infty$ for all $x \in \R^d$, the process $\{X_t^\varphi : 0 \leq t \leq T\}$ driven by the SDE
\begin{equation}
    \label{eq:sde}
    \dd X_t^\varphi = -b\left(X_t^\varphi - m\right) \, \dd t + \Sigma \nabla \log \cT_{T - t}[e^{\varphi}](X_t^\varphi) \, \dd t + \Sigma^{1/2} \, \dd W_t,
    \quad
    0 < t < T,
\end{equation}
has the transition density of the form (see \cite{Daipra1991}, Theorem 2.1)
\begin{equation}
    \label{eq:transition_density}
    (X_t^\varphi \,|\, X_s^\varphi = x) \sim \frac{\cT_{T - t}[e^\varphi](y)} {\cT_{T - s}[e^\varphi](x)} \, \sfq_{t - s}(y \,|\, x),
    \quad 0 \leq s < t \leq T.
\end{equation}
As before, for any $g : \R^d \rightarrow \R$, $\cT_t [g](x)$ stands for the Ornstein--Uhlenbeck operator:
\begin{align*}
    \cT_t [g](x)
    &
    = \int\limits_{\R^d} \sfq_{t}(y \,\vert\, x) \, g(y) \, \dd y
    \\&
    = \frac{1}{(2\pi)^{d/2} \sqrt{\det(\Sigma_t)}} \int\limits_{\R^d} \exp\left\{-\frac12 \left\|\Sigma_t^{-1/2} \big(y - m_t(x) \big) \right\|^2 \right\} g(y) \, \dd y,
    \quad x \in \R^d.
\end{align*}
The formula \eqref{eq:transition_density} yields that the terminal density of the process \eqref{eq:sde} with the initial condition $X_0^\varphi \sim \rho_0$,
is given by
\begin{equation}
    \label{eq:rho_T_psi}
    \rho_T^\varphi(y) = \int\limits_{\R^d} \frac{e^{\varphi(y)} \sfq_T(y \,\vert\, x)}{\cT_T \big[e^{\varphi} \big](x)} \, \rho_0(x) \, \dd x.
\end{equation}
We measure the proximity of a log-potential $\varphi: \R^d \rightarrow \R$, $\cT_T[e^\varphi](x) < +\infty$ for all $x \in \R^d$, to $\varphi^*$ via the Kullback--Leibler divergence $\KL(\rho_T^*, \rho_T^\varphi)$. Since the target density $\rho_T^*$ is unknown, we replace the KL-divergence by its empirical counterpart
\begin{equation}
    \label{eq:hat_kl}
    \widehat\KL\left(\rho_T^*, \rho_T^\varphi \right)
    = \frac1n \sum\limits_{i = 1}^n \log \frac{\rho_T^*(Y_i)}{\rho_T^\varphi(Y_i)}
\end{equation}
and study properties of an approximate empirical risk minimizer 
\begin{equation}
    \label{eq:erm}
    \widehat \varphi \in \Phi :
    \quad
    \widehat\KL\left(\rho_T^*, \rho_T^{\widehat \varphi} \right)
    \leq \inf\limits_{\varphi \in \Phi} \widehat\KL\left(\rho_T^*, \rho_T^\varphi \right) + \opterr
\end{equation}
over a reference class of log-potentials $\Phi$ (for instance, a class of feed-forward neural networks). The parameter $\opterr$ reflects optimization error. We leave its study beyond the scope of the present paper and focus on statistical properties of the estimate \eqref{eq:erm}. Similarly to other approaches based on empirical risk minimization (see, for example, \cite{Korotin2024}, \cite{Gushchin2024a}, \cite{Rapakoulias2024}), it combines the full problem  (the marginals, transition densities, and the potential function) into one single optimization framework and aims to directly minimize the objective of matching the marginals at time $ T $ without separating the problem into smaller subproblems (in contrast to iterative proportional \citep{DeBortoli2021} and Markovian \citep{Shi2023} fitting, which update Schrödinger bridge estimates step by step). However, in \cite{Korotin2024}, \cite{Gushchin2024a}, \cite{Rapakoulias2024}, the authors focused primarily on methodological aspects of generative modelling and entropic optimal transport. For this reason, generalization ability of their estimates remains largely unexplored.

\paragraph*{Contribution}
The main contributions of the present paper are non-asymptotic statistical guarantees on proximity of the marginal endpoint density $\widehat \rho_T$, corresponding to $\widehat \varphi$ from \eqref{eq:erm}, to the target one.
\begin{itemize}
    \item In the case of a bounded sub-Gaussian density $\rho_T^*$, we show that, for any $\delta \in (0, 1)$, with probability at least $(1 - \delta)$, the KL-risk $\KL(\rho_T^*, \widehat\rho_T)$ satisfies the oracle inequality 
    \begin{equation}
        \label{eq:excess_kl_bound_simplified}
        \KL(\rho_T^*, \widehat\rho_T) - \inf\limits_{\varphi \in \Phi} \KL( \rho_T^*, \rho_T^\varphi)
        \lesssim \opterr + \sqrt{\Upsilon(n, \delta)  \, \inf\limits_{\varphi \in \Phi} \KL( \rho_T^*, \rho_T^\varphi )} 
        + \Upsilon(n, \delta),
    \end{equation}
    where $\opterr$ is the optimization error from \eqref{eq:erm} and
    \[
        \Upsilon(n, \delta) \lesssim \frac{\log^2(en) + \log(1/\delta)\log(en)}n.
    \]
    Here and further in the paper, the sign $\lesssim$ stands for an inequality up to a multiplicative constant. In contrast to \cite{Korotin2024}, the excess risk may decrease much faster than $\cO(n^{-1/2}$, provided that the class of log-potentials $\Phi$ is rich enough to approximate the target density $\rho_T^*$ and $\opterr$ is small.
    \item Furthermore, if the target log-density $\log \rho_T^*$ is $s$-smooth and exhibits a sub-quadratic behaviour, then the right Schrödinger log-potential $\varphi^*$ can be approximated via feedforward deep neural networks (Theorem~\ref{th:approximation_error_relu}). As a consequence, we show that (see Corollary~\ref{cor:combined_estimation_approximation_rate}), if a learner chooses a proper neural network architecture, then the terminal marginal density $\widehat \rho_T$, corresponding to the exact empirical risk minimizer (that is the estimate \eqref{eq:erm} with $\opterr = 0$), fulfils  
    \[
        \KL(\rho_T^*, \widehat\rho_T) = \widetilde\cO \left( n^{-2s / (2s + d)} + \frac{\log(1 / \delta)}n \right)
    \]
    with probability at least $(1 - \delta)$.
\end{itemize}
Both guarantees admit target densities, which are not separated from zero and have unbounded supports.

\paragraph*{Paper structure}
The rest of the paper is organized as follows. We provide a short review of related work in Section~\ref{sec:related_work} and introduce necessary definitions and notations in Section \ref{sec:notation}. In Section~\ref{sec:main_result} is devoted to our main results: Theorem~\ref{th:excess_kl_bound} establishes the non-asymptotic estimation error bounds, while Theorem~\ref{th:approximation_error_relu} and Corollary~\ref{cor:combined_estimation_approximation_rate} provide approximation error guarantees and convergence rates for clipped ReLU networks. Section~\ref{sec:numerical} demonstrates the practical performance of our approach through numerical experiments on Gaussian mixtures and photorealistic face generation. The proofs of the main theorems are detailed in Section~\ref{sec:th_excess_kl_bound_proof} and Section~\ref{sec:proof_approximation_error_relu}. Rigorous derivations and auxiliary technical results are deferred to the supplementary material.

\section{Related work}
\label{sec:related_work}

\paragraph*{Entropic optimal transport and Sinkhorn methods}
The Schr\"odinger bridge problem is closely related to
entropically regularized optimal transport and its dual
formulation \citep{Leonard2014,Chen2021}.
A classical computational approach is the Sinkhorn
algorithm \citep{Sinkhorn1967}, which alternately updates
two potentials to enforce the prescribed marginal
constraints. Its applications include data-driven
Schr\"odinger bridges \citep{Pavon2018}, generative modelling
\citep{Genevay2018}, and financial calibration
\citep{March2023}. Recent theoretical developments establish
stability properties of entropic transport plans and
convergence guarantees for Sinkhorn iterations under
various assumptions on the marginals
\citep{Conforti2024a,Chiarini2024}.

\paragraph*{Dynamic and stochastic-control approaches}
The connection between Schr\"odinger bridges and stochastic
optimal control \citep{Daipra1991,Chen2021} has motivated
several approaches to sampling and generative modelling
\citep{Tzen2019,Zhang2022}. Diffusion Schr\"odinger Bridge
\citep{DeBortoli2021} employs iterative proportional
fitting, whereas Diffusion Schr\"odinger Bridge Matching
\citep{Shi2023} introduces iterative Markovian fitting
based on alternating Markovian and reciprocal projections.
Further developments include Light and Optimal
Schr\"odinger Bridge Matching \citep{Gushchin2024a},
adversarial bridge matching \citep{Gushchin2024b},
and diffusion-based approaches to generative modelling
and image-to-image translation \citep{Wang2021,Liu2023}.

\paragraph*{Statistical guarantees.}
Statistical properties of Schr\"odinger bridge estimators
have been investigated in
\citep{Stromme2023,Pooladian2024,Korotin2024}.
A detailed comparison of their convergence guarantees
with our results is provided after
Theorem~\ref{th:excess_kl_bound}.
The results of the present paper have also been used
in subsequent work by \citet{puchkin26}, who study
Schr\"odinger potential estimation in unpaired data
translation. In particular, their analysis relies
on the comparison estimates established in Lemma~D.3.

\section{Notation}
\label{sec:notation}

Throughout the paper, we denote the Euclidean, the $\ell_\infty$-, and the $\ell_0$-norm of a vector $x$ by $\|x\|$, $\|x\|_\infty$, and $\|x\|_0$, respectively. For a matrix $A$, $\|A\|$, $\|A\|_\infty$, and $\|A\|_0$ stand for the operator norm, the Chebyshev norm, and the number of non-zero elements, respectively. The notation $f \lesssim g$ or $g \gtrsim f$ means that $f = \cO(g)$. Besides, we often replace $\max\{a, b\}$ and $\min\{a, b\}$ by shorter expressions $a \vee b$ and $a \wedge b$, respectively. Given a multi-index $\alpha = (\alpha_1, \dots, \alpha_k) \in \Z_+^k$ and a $|\alpha| = \alpha_1 + \ldots + \alpha_k$ times differentiable function $f$, we denote its partial derivative
\[
	\frac{\partial^{|\alpha|} f(x)}{\partial x_1^{\alpha_1} \dots \partial x_k^{\alpha_k}}
\]
by $\partial^\alpha f$ for short.

For any $s \geq 1$, the Orlicz $\psi_s$-norm of a random variable $\xi$ is defined as
\[  
    \|\xi\|_{\psi_s} = \inf\left\{ u > 0 : \E e^{|\xi|^s / u^s} \leq 2 \right\}.
\]
Given $p \geq 1$ and a probability density $\rho$, the weighted $L_p$-norm of a function $f$ is defined as $\|f\|_{L_p(\rho)} = \big(\E_{\xi \sim \rho} |f(\xi)|^p \big)^{1/p}$. We also write $\|f\|_{\psi_s(\rho)}$ for the Orlicz norm of $f(\xi)$ when $\xi\sim\rho$. Given two probability densities $\sfp \ll \sfq$ on $\R^d$, the Kullback--Leibler and Jeffreys divergence between them are defined as $\KL(\sfp, \sfq) = \E_{\xi \sim \sfp} \log\big( \sfp(\xi) / \sfq(\xi) \big)$ and $\J(\sfp,\sfq) = \KL(\sfp,\sfq) + \KL(\sfq,\sfp)$, respectively.
For an integrable function $g : \R^d \rightarrow \R$ and $t \in (0, +\infty)$ we define the Ornstein--Uhlenbeck operator
\[
    \cT_t [g](x)
    = \frac{1}{(2\pi)^{d/2} \sqrt{\det(\Sigma_t)}} \int\limits_{\R^d} \exp\left\{-\frac12 \left\| \Sigma_t^{-1/2} \big(y - m_t(x) \big) \right\|^2 \right\} g(y) \, \dd y,
    \quad x \in \R^d.
\]
Since $X_t^0$ from \eqref{eq:ou_process} is a geometrically ergodic Markov process with the limiting distribution $\cN\big( m, \Sigma / (2b) \big)$, it holds that $\cT_\infty [g](x) \equiv \cT_\infty[g]$ is a constant.

Finally, let us introduce a class of feedforward neural networks, which is often used in the literature on statistical properties of neural networks and deep generative models (see, for instance, \cite{schmidt-hieber20}, \cite{oko23}, and \cite{yakovlev25}). We would like to recall that the ReLU activation function is defined as $\relu(x) = 0 \vee x$.
For a vector $b \in \R^r$, let $\relu_b : \R^r \ni x \mapsto \smash{\big( \relu(x_1 - b_1), \dots, \relu(x_r - b_r) \big)}^\top$ stand for the shifted activation function. Given a number of layers $K \in \N$ and an architecture vector $W = (W_0, \dots, W_K) \in \R^{K + 1}$, a feedforward neural network $f : \R^{W_0} \to \R^{W_K}$ is a composition of linear and nonlinear transforms of the form
\begin{equation}
\label{eq:f_nn_def}
    f(x) = -b^{(K)} + A^{(K)} \circ \relu_{b^{(K - 1)}} \circ A^{(K - 1)} \circ \ldots \circ \relu_{b^{(1)}} \circ A^{(1)} \circ x, 
    \quad x \in \R^{W_0}.
\end{equation}
Here, $A^{(j)} \in \R^{W_j \times W_{j - 1}}$ is a weight matrix and $b^{(j)} \in \R^{W_j}$ is a bias vector for each $j \in \{1, \dots, K\}$. Throughout the paper
\begin{equation}
    \label{eq:nn_class}
    \NN(K, W, S, B) = \left\{f \text{ of the form } \eqref{eq:f_nn_def} : \substack{\sum\limits_{j = 1}^K \big( \|A_j\|_0 + \|b_j\|_0 \big) \leq S,\\
    \max\limits_{1 \leq j \leq K} \big\{ \|A_j\|_\infty \vee \|b_j\|_\infty \big\} \leq B} \right\}
\end{equation}
denotes the class of feedforward neural networks of depth $(K + 1)$ with at most $S \in \N$ non-zero weights, the architecture $\|W\|_\infty \in \N$, and the weight magnitude $B$.

\section{Main results}
\label{sec:main_result}

In the present section, we discuss theoretical properties of the approximate empirical risk minimizer $\widehat\varphi$ \eqref{eq:erm}. We derive non-asymptotic upper bounds on estimation and approximation errors (Theorems \ref{th:excess_kl_bound} and \ref{th:approximation_error_relu}, respectively) and combine them to establish explicit rates of convergence (Corollary \ref{cor:combined_estimation_approximation_rate}). We focus on statistical properties of $\widehat\varphi$ leaving the study of optimization and discretization errors, data-dependent reference process selection, and accuracy of Monte Carlo approximation of the Ornstein--Uhlenbeck operator out of the scope of the present paper.

\subsection{Assumptions}

Before we move to a high-probability Bernstein-type upper bound, we introduce necessary assumptions. First, as we announced before, we use the Ornstein--Uhlenbeck process $X_t^0$ as the reference one.

\begin{As}
\label{as:base-process}
The base process $X^0$ solves the SDE
\[
    \dd X_t^0 = b \left( m -  X_t^0 \right) \dd t + \Sigma^{1/2} \, \dd W_t,
    \quad 0 \leq t \leq T,
\]
where $b > 0$, $m \in \mathbb{R}^d,$ $\Sigma $ is a positive definite symmetric matrix of size $d\times d$, and $W$ is a $d$-dimensional Brownian motion.
\end{As}

Main properties of the Ornstein--Uhlenbeck process were discussed in the previous section.
Second, we suppose that the source and target densities meet the following requirements.

\begin{As}
    \label{as:bounded_support}
    The initial density $\rho_0$ has a bounded support and $\|\Sigma^{-1/2}(x - m)\| \leq R$ for some $R>0$ and all $x \in \supp(\rho_0)$.
\end{As}

\begin{As}
    \label{as:sub_gaussian_density}
     The target distribution at time $T$ admits a bounded density $\rho_T^*$ with respect to the Lebesgue measure such that
     \[
        \rho_T^*(x) \leq \rho_{\max} \quad \text{for all $x \in \R^d$.}
     \]
     Moreover, the target distribution $\rho_T^*$ is sub-Gaussian with variance proxy $\ttv^2$, that is,
     \begin{equation}
        \label{eq:rho_T_sub-gaussian}
        \E_{Y \sim \rho_T^*} e^{u^\top Y} \leq e^{\ttv^2 \|u\|^2 / 2}
        \quad \text{for any $u \in \R^d$.}
     \end{equation}
\end{As}

Assumption \ref{as:bounded_support} is not restrictive, since in unconditional generation problems a learner is free to choose the source distribution.
Assumption \ref{as:sub_gaussian_density} is also very mild. Despite the fact that we deal with logarithmic loss, we do not require $\rho_T^*$ to be bounded away from zero. In contrast to $\rho_0$, we do not even require its support to be compact. This significantly complicates the proof of the excess KL-bound and poses nontrivial technical challenges. 
Let us note that the condition \eqref{eq:rho_T_sub-gaussian} yields that $\E_{Y \sim \rho_T^*} Y = 0$. However, it does not diminish generality of our setup.

The remaining assumptions concern properties of the class of log-potentials $\Phi$. First, we assume that admissible log-potentials $\varphi(x)$ are bounded from above and behave as $\cO(\|x\|^2)$ as $x$ tends to infinity.

\begin{As}
    \label{as:log-potential_quadratic_growth}
    There exist non-negative constants $L$ and $M$ such that
    \[
        -L \left\| \Sigma^{-1/2} (x - m) \right\|^2 - M
        \leq \varphi(x)
        \leq M
        \quad \text{for all $x \in \R^d$ and $\varphi \in \Phi$.}
    \]
    Moreover, for any $\varphi \in \Phi$, it holds that $ \cT_\infty[\varphi] = \E\varphi(X_\infty^0) = 0$.
\end{As}

The condition $\cT_\infty \varphi = 0$ appears because of the fact that the Schr\"odinger potentials $\nu_0$ and $\nu_T$ (see \eqref{schroe1}) are defined up to a multiplicative constant. The requirement $\cT_\infty \varphi = 0$ is nothing but a normalization. 
Second, we assume that $\Phi$ is parametrized by a finite-dimensional parameter $\theta \in \R^D$:
\[
    \Phi = \left\{ \varphi_\theta : \theta \in \Theta \right\},
\]
where $\Theta$ is a subset of a $D$-dimensional cube $[-1, 1]^D$ and each function $\varphi_\theta$ maps $\R^d$ into $\mathbb{R}$. We suppose that the parametrization is sufficiently smooth in the following sense.
\begin{As}
    \label{as:lipschitz_parameterization}
    There exists $\Lambda \geq 1$ such that
    \[
        \left| \varphi_{\theta}(x) - \varphi_{\theta'}(x) \right| \leq \Lambda \left(1 + \left\|\Sigma^{-1/2}(x - m) \right\|^2 \right) \|\theta - \theta'\|_{\infty}
    \]
    for all $\theta, \theta' \in \Theta$ and all $x \in \R^d$.
\end{As}

Assumptions \ref{as:log-potential_quadratic_growth} and \ref{as:lipschitz_parameterization} are quite general. We provide three examples when they hold. First, in the case when $\rho_0$ and $\rho_T$ are Gaussian measures the log-potential $\varphi^*(x)$ is quadratic and meets these requirements under the proper choice of $b$, $T$, and $\Sigma$ (see, for instance, Lemma A.1 in \cite{patarusau26}). 
Second, consider the mixture form \eqref{eq:light_potential} used in the experiments. Fix $d,K$ and the reference parameters $b,T,m,\Sigma$. Suppose that for every component of every mixture,
$\lvert\log\alpha_k\rvert\le H$, $\|\Sigma^{-1/2}r_k\|\le R$, and $aI_d\preceq S_k\preceq qI_d$, with common finite $H,R$ and $0<a<q<1$.
Use the log-weights, the coordinates of the whitened centers, and the independent entries of the shape matrices as parameters, rescaled to lie in $[-1,1]$. Then the centered log-potentials $f-\cT_\infty[f]$, with $f$ given by \eqref{eq:light_potential}, satisfy Assumptions~\ref{as:log-potential_quadratic_growth} and~\ref{as:lipschitz_parameterization} with common constants $L,M,\Lambda$. Similar log-potentials were used by \cite{Korotin2024} to construct a computationally friendly baseline for approximate solution of the Schrödinger problem. Finally, Assumptions \ref{as:log-potential_quadratic_growth} and \ref{as:lipschitz_parameterization} will be satisfied for more complex classes of functions, such as neural networks with ReLU activations. While it is clear that one can enforce Assumption \ref{as:log-potential_quadratic_growth} by proper truncation (as it will be done later, see \eqref{eq:Phi}), the question, whether Assumption \ref{as:lipschitz_parameterization} holds, needs a discussion. 

Let $\bbM = \{ M^{(k)} : 1 \leq k \leq K\}$ be a collection of matrices $M^{(k)} \in \{0, 1\}^{W_k \times (W_{k - 1} + 1)}$. We denote
\begin{equation}
    \label{eq:nn_class_with_mask}
    \cF(K, W, \bbM, B) = \left\{f \text{ of the form } \eqref{eq:f_nn_def} : \substack{A_{i, j}^{(k)} \neq 0 \text{ iff } M_{i, j}^{(k)} \neq 0 \text{ for all $1 \leq i \leq W_k$, $1 \leq j \leq W_{k - 1}$},\\
    b_i^{(k)} \neq 0 \text{ iff } M_{i, W_{k - 1} + 1}^{(k)} \neq 0 \text{ for all $1 \leq i \leq W_k$},\\
    \|A^{(k)}\|_\infty \leq B \text{ and }  \|b^{(k)}\|_\infty \leq B \text{ for all $1 \leq k \leq K$}
    }\right\}
\end{equation}
the set of neural networks of depth $(K + 1)$ with architecture $W$ and the mask $\bbM$. The following proposition indicates that $\cF(K, W, \bbM, B)$ satisfies Assumption \ref{as:lipschitz_parameterization}.

\begin{Prop}
    \label{prop:lambda_bound}
    Let $f_\theta$ and $f_{\theta'}$ be arbitrary functions from $\cF(K, W, \bbM, B)$ corresponding to some vectors of non-zero weights $B\theta, B\theta' \in [-B, B]^{|\bbM|}$. If $\|W\|_\infty \geq 2$ and $B \geq 1$, then for any $x \in \R^d$ it holds that
    \[
        \left| f_\theta(x) - f_{\theta'}(x) \right|
        \leq \Lambda \left( 1 + \left\| \Sigma^{-1/2} (x - m) \right\|^2 \right) \|\theta - \theta'\|_\infty
    \]
    with
    \begin{equation}
        \label{eq:lambda_nn}
        \Lambda = \Lambda(K, W, B)
        = K B^{K} \|W\|_\infty^{K} \left(B + \|m\|_\infty + \frac{\sqrt{\|\Sigma\|}}2\right).
    \end{equation}
\end{Prop}

Let us note that the class $\NN(K, W, S, B)$, defined in \eqref{eq:nn_class}, admits a representation
\[
   \NN(K, W, S, B) = \bigcup\limits_{\bbM : |\bbM| \leq S} \cF(K, W, \bbM, B),
   \quad \text{where} \quad
   |\bbM| = \sum\limits_{k = 1}^K \sum\limits_{i = 1}^{W_k} \sum\limits_{j = 1}^{W_{k - 1} + 1} M_{i,j}^{(k)}.
\]
In other words, $\NN(K, W, S, B)$ is the union of parametric classes satisfying Assumption \ref{as:lipschitz_parameterization}. This fact will be enough for our purposes.

\subsection{Oracle inequality for bounded sub-Gaussian densities}

We proceed with the analysis of the excess KL-risk of the estimate $\widehat\varphi$. 

\begin{Th}
    \label{th:excess_kl_bound}
    Grant Assumptions \ref{as:base-process}--\ref{as:lipschitz_parameterization}.
    Let $\widehat \varphi$ be defined in \eqref{eq:erm} and let $\widehat\rho_T$ be the corresponding density of $X_T^{\widehat \varphi}$.
    Then, for any $\delta \in (0, 1)$, with probability at least $1 - \delta$, it holds that
    \begin{align}
        \label{eq:oracle_inequality}
        \KL(\rho_T^*, \widehat\rho_T) - \inf\limits_{\varphi \in \Phi} \KL( \rho_T^*, \rho_T^\varphi )
        &\notag
        \lesssim \opterr + \sqrt{(M + Ld + d) \Upsilon(n, \delta)  \, \inf\limits_{\varphi \in \Phi} \KL( \rho_T^*, \rho_T^\varphi)} 
        \\&\quad
        + (M + Ld + d) \Upsilon(n, \delta),
    \end{align}    
    where
    \[
        \Upsilon(n, \delta) = \frac{\log(en)}n \left( D \log\big(\Lambda (1 + M) n\big) + D \sqrt{d} + \log(e/\delta) \right).
    \]
    The hidden constant behind $\lesssim$ depends on $R$, $T$, $\|\Sigma\|$, $\|\Sigma^{-1}\|$, $m$, $b$, $\ttv$, and $\rho_{\max}$.
\end{Th}

Theorem~\ref{th:excess_kl_bound} provides an oracle inequality that separates approximation, statistical, and optimization errors.
In the realizable case, when $\rho_T^*$ belongs to the model class $\{\rho_T^\varphi:\varphi\in\Phi\}$, the approximation error vanishes, yielding a high-probability KL-risk bound of order $\cO(\log^2(en)/n + \opterr)$ for fixed model complexity and reference parameters.

Related statistical guarantees for Schr\"odinger bridges have been established in \citep{Korotin2024,Pooladian2024}. In \citep{Korotin2024}, the authors introduce Light Schr\"odinger Bridge, which employs a Gaussian-mixture parameterization of Schr\"odinger potentials and a variational objective associated with the KL divergence between endpoint couplings.
Their Theorem~A.1 establishes an expected statistical error bound of order $\cO(n^{-1/2})$ for a bounded-parameter mixture class, assuming compact support of both marginals.
In contrast, our oracle inequality controls the KL divergence
between terminal marginal densities and allows bounded sub-Gaussian target densities with unbounded support.
Moreover, our bound holds with high probability and exhibits faster convergence in the realizable case.
Nevertheless, the different statistical objectives prevent
a direct comparison of the convergence rates.

In \citep{Pooladian2024}, the authors analyze a plug-in Sinkhorn-based estimator of the Schr\"odinger bridge. 
Assuming that the target measure is supported on a compact subset of a smooth $k$-dimensional submanifold, they obtain
a bound of order $\cO(n^{-1/2}+(1-\tau)^{-k-2}n^{-1})$ on the expected squared total variation distance between path measures restricted to $[0,\tau]$, where $\tau < 1$ and the diffusion parameters are fixed. Unlike this
result, Theorem~\ref{th:excess_kl_bound} establishes a KL-risk guarantee directly at the terminal time.
The two approaches therefore provide complementary statistical guarantees, under different assumptions and for different notions of estimation error.

\subsection{Rates of convergence for smooth sub-quadratic log-densities}
\label{sec:relu_approximation_rates}

The oracle inequality in Theorem~\ref{th:excess_kl_bound} separates the statistical error from the approximation error $\inf_{\varphi\in\Phi}\KL(\rho_T^*,\rho_T^\varphi)$.
In this subsection, we elaborate on this term. In particular we are interested in capabilities of feedforward neural networks with ReLU activations to approximate log-densities satisfying the following assumption.

\begin{As}
    \label{as:log-density_envelope}
    The target density $\rho_T^*$ is strictly positive on $\R^d$ and satisfies
    \[
        -a_- \left\| \Sigma^{-1/2} (y - m) \right\|^2 - b_-
        \le \log\rho_T^*(y)
        \le -a_+ \left\| \Sigma^{-1/2} (y - m) \right\|^2 + b_+,
        \qquad y\in\R^d,
    \]
    for some real numbers $b_-, b_+ \geq 0$ and $a_- \geq a_+ > 0.5 \sigma_T^{-2}$.
\end{As}

\begin{As}
    \label{as:smoothess_class}
    Let $s\in\mathbb N$ and assume that $\log\rho_T^*\in C^s_{\mathrm{loc}}(\R^d)$ and that there exist constants $C_\rho,\beta_\rho<\infty$ such that, for all $r\ge1$,
    \[
        A_r(\log \rho_T^*)
        = \max_{|\alpha|\le s} \; \max\limits_{\|y\| \leq r} \left\{ r^{|\alpha|} \big| \partial^\alpha \log \rho_T^*(y) \big| \right\}
        \le C_\rho \, r^{\beta_\rho}.
    \]
\end{As}

Assumptions~\ref{as:log-density_envelope}
and~\ref{as:smoothess_class} hold for a broad class
of nondegenerate Gaussian mixture densities of the form
\begin{equation}
    \label{eq:gaussian_mixture}
    \rho_T^*(y)
    = \frac{\det(S)^{-1/2}}{\sqrt{2\pi}} \int_{\R^d} \exp\left\{-\frac12 \left\| S^{-1/2} (y - z) \right\|^2 \right\} \, \sfQ(\dd z),
\end{equation}
where $S$ is a symmetric positive definite matrix and $\sfQ$ is a compactly supported
probability measure. Such models include the nondegenerate Gaussian mixtures considered in \citep{yakovlev25,yakovlev25b}
and the Gaussian mixture example discussed in
\citep{puchkin26}, provided that the reference
process satisfies the above covariance condition.
Related Gaussian-mixture parameterizations of
Schr\"odinger potentials appear in
\citep{Korotin2024}. In addition, Gaussian convolutions with compactly supported or sub-Gaussian mixing distributions were also studied in \citep{jia23}.
Under the model \eqref{eq:gaussian_mixture}, $\log\rho_T^*$ is
infinitely differentiable, and its derivatives
have at most polynomial growth. In particular,
Assumption~\ref{as:smoothess_class} holds for
every $s\in\N$. Moreover,
\[
    \log\rho_T^*(y)
    = -\frac12 (y - m)^\top S^{-1}(y - m)
    + \cO(1 + \|y - m\|)
    \quad \text{as $\|y\|\to\infty$.}
\]
Consequently, Assumption~\ref{as:log-density_envelope} also
holds whenever $S\prec\sigma_T^2\Sigma$.

Before we proceed, let us describe the construction of the class of neural networks. For given $K, S \in \N$, $B > 0$, and a vector $W \in \N^{K + 1}$, let $\NN(K, W, S, B)$ be a class of neural networks defined in \eqref{eq:nn_class}. We fix some positive constants $L_0 \geq a_-$ and $M_0$ and introduce a truncation operator
\begin{equation}
    \label{eq:ttT}
    \ttT_{L_0, M_0}[g](x) = \max\left\{-L_0 \left\|\Sigma^{-1/2} (x - m)\right\|^2 - M_0, \, \min\{g(x), M_0\} \right\}.
\end{equation}
Finally, we define
\[
    \Psi_{L_0, M_0}(K, W, S, B)
    = \big\{ \psi(x) = \ttT_{L_0, M_0}[g](x) : g \in \NN(K, W, S, B) \big\}
\]
and
\begin{equation}
    \label{eq:Phi}
    \Phi_{L_0, M_0}(K, W, S, B)
    = \big\{ \psi(x) - \cT_\infty[\psi] : \psi \in \Psi_{L_0, M_0}(K, W, S, B) \big\}.
\end{equation}
The class $\Phi_{L_0, M_0}(K, W, S, B)$ has the following approximation power.

\begin{Th}
\label{th:approximation_error_relu}
Grant Assumptions~\ref{as:base-process}, \ref{as:bounded_support}, \ref{as:sub_gaussian_density}, \ref{as:log-density_envelope}, and \ref{as:smoothess_class}. Put $\beta = \beta_\rho \vee s \vee 2$. Then one can choose $L_0, M_0 > 0$, such that for any resolution level $N \geq 2$ there exists a class of truncated centered neural networks $\Phi_N = \Phi_{L_0, M_0}(K_N, W_N, S_N, B_N)$ with
\[
    K_N \lesssim \log N,
    \quad
    \|W_N\|_\infty \lesssim N \log N,
    \quad
    S_N \lesssim N \log N,
    \quad \text{and} \quad
    B_N \lesssim \log^{\beta / 2}(N),
\]
which fulfils Assumption \ref{as:log-potential_quadratic_growth} with 
$L = L_0$, $M = 2M_0 + dL_0 / (2b)$ and satisfies the inequality
\[
    \inf_{\varphi\in\Phi_N}
    \KL(\rho_T^*, \rho_T^\varphi)
    \lesssim \frac{\log^{\beta}(N)}{N^{2s / d}}.
\]
\end{Th}

Theorem~\ref{th:approximation_error_relu} extends the
approximation capabilities of deep ReLU networks,
well studied for smooth functions on bounded domains
\citep{schmidt-hieber20,yarotsky2017error,guehring2020error},
to terminal distributions induced by Schr\"odinger
log-potentials. Although the smoothness assumptions
are imposed on the target log-density rather than
directly on the potential, they imply global regularity
of the latter with controlled growth of its derivatives.
The proof combines neural network approximation
on expanding balls with truncation, centering,
and sub-Gaussian tail bounds to obtain a global
KL approximation guarantee. In particular, sparse
ReLU networks achieve an approximation error of
order $\widetilde{\mathcal O}(N^{-2s/d})$.
Moreover, the proof establishes the same bound
for the joint endpoint distributions, which is
stronger than the stated terminal-marginal guarantee.

\begin{Co}
\label{cor:combined_estimation_approximation_rate}
Under Assumptions \ref{as:base-process}, \ref{as:bounded_support}, \ref{as:sub_gaussian_density}, \ref{as:log-density_envelope}, and \ref{as:smoothess_class}, one can choose $L_0, M_0 > 0$, such that for any resolution level $n \in \mathbb N$ there exists a class of truncated centered neural networks $\Phi_n = \Phi_{L_0, M_0}(K_n, W_n, S_n, B_n)$ with $K_n \lesssim \log(en)$
\[
    \|W_n\|_\infty \lesssim n^{d / (2s + d)} \log(en),
    \quad
    S_n \lesssim n^{d / (2s + d)} \log(en),
    \quad \text{and} \quad
    B_n \lesssim \log^{\beta / 2}(en),
\]
where $\beta = \beta_\rho \vee s \vee 2$, such that, for any $\delta \in (0, 1)$, the endpoint density $\widehat \rho_T$, induced by the exact empirical risk minimizer $\widehat \varphi$ over the class $\Phi_n$, satisfies 
\[
    \KL(\rho_T^*, \widehat\rho_T) \lesssim  \frac{\log^{\beta \vee 4}(en)}{n^{2s / (2s + d)}} + \frac{\log(en) \log(1 / \delta)}n
\]
with probability at least $(1 - \delta)$.
\end{Co}

The rate
$\widetilde{\mathcal O}(n^{-2s/(2s+d)})$
in Corollary~\ref{cor:combined_estimation_approximation_rate}
matches, up to logarithmic factors, the classical
minimax convergence exponent for nonparametric density
estimation under KL divergence.
In the univariate setting, \citet{koo96}
establish the rate $n^{-2s/(2s+1)}$ for densities
whose log-densities belong to Besov classes on
$[0,1]$. Their assumptions for the upper bound
imply that the densities are uniformly bounded above
and away from zero. Multivariate KL-risk bounds
with the corresponding convergence exponent have
been established in \citep{yang99},
while \citet{wang22} obtain analogous
rates for transport-based maximum likelihood estimators.
These results concern densities on bounded domains,
whereas our analysis accommodates target distributions
with unbounded support.

The latter setting presents additional difficulties
for KL-risk analysis, since approximation errors in
the tails may contribute significantly to the
divergence even when the densities are well
approximated on bounded domains. In particular,
a density on $\mathbb R^d$ cannot be uniformly
bounded away from zero. Our analysis instead relies
on two-sided Gaussian bounds on the target density
and polynomial growth conditions on the derivatives
of its log-density. These assumptions control the
tail contributions to KL divergence and yield a
global terminal-marginal KL guarantee with the
classical nonparametric convergence exponent.

\section{Numerical Experiments}
\label{sec:numerical}

The parametrization of the loss $(\varphi, y) \mapsto \log \rho_T^\varphi(y)$ through the right Schr\"odinger log-potential poses not only theoretical but also practical challenges. Indeed, given a log-potential $\varphi : \R^d \rightarrow \R$, a statistician has to use double Monte-Carlo technique to approximate the value of $\cT_T[e^\varphi](x)$ and of the integral \eqref{eq:rho_T_psi}. This makes direct optimization of the empirical loss computationally and memory intensive.
The same issue arises for a slightly different
loss function considered in \cite{Korotin2024} and \cite{puchkin26}. 
In this section, we aim to suggest a practical way to find an approximate empirical risk minimizer \eqref{eq:erm} based on properties of the operator $\cT_t$. In particular, geometric ergodicity of the Ornstein--Uhlenbeck process implies the following two-sided bound. 

\begin{Lem}[simplified version of Lemma \ref{lem:log_ou_bound}]
    \label{lem:log_ou_bound_simplified}
    Let $\varphi : \R^d \rightarrow \R$ be such that $\varphi(y) \leq M$ for all $y \in \R^d$.
    Then for any $x \in \R^d$ and $t > 0$ it holds that
    \begin{equation}
        \label{eq:ou_two-side_bound_simplified}
        e^{-\cA(x, t) \cK(t)} \left( \frac{\cT_\infty[e^{\varphi}]}{e^M} \right)^{\cK(t)}
        \leq \frac{\cT_t [e^{\varphi}](x)}{e^M} \leq e^{\cA(x, t)} \left( \frac{\cT_\infty [e^{\varphi}]}{e^M} \right)^{1 / \cK(t)},
    \end{equation}
    where
    \begin{equation}
        \label{eq:cK}
        \cK(t) = \big(1 - e^{-2bt} \big)^{-5e^2 \sqrt{d}} \cdot \exp\left\{ 2e^2 \sqrt{d} \arcsin(e^{-bt}) \right\}
    \end{equation}
    and 
    \begin{equation}
        \label{eq:cA}
        \cA(x, t) = \frac{b e^2}{\sqrt{d}} \left\|\Sigma^{-1/2} (x - m) \right\|^2 \arcsin\left(e^{-bt} \right) + 2 \log \cK(t).
    \end{equation}
\end{Lem}

Lemma \ref{lem:log_ou_bound_simplified} helps us to construct a surrogate loss using the following log-density representation.

\begin{Prop}
    \label{prop:approx_erm}
    Under Assumptions \ref{as:base-process} and \ref{as:bounded_support},
    for any $\varphi : \R^d \rightarrow (-\infty, M]$ and any $y \in \R^d$, it holds that
    \begin{align}
        \label{eq:log_rho_T_representation}
        \log \rho_T^\varphi(y)
        &
        = \log \rho_T^0(y) + \varphi(y) - M + \frac1{\cK(T)} \log\frac{e^M}{\cT_\infty[e^\varphi]}
        \\&\quad
        + \cO\left( \left(\frac{R^2}d + 1 + \log\frac{e^M}{\cT_\infty[e^\varphi]} \right) \sqrt{d} \, e^{-bT} \right)
    \end{align}
    when $bT$ tends to infinity. The hidden constant behind $\cO(\cdot)$ is uniform with respect to $\varphi : \R^d \rightarrow (-\infty, M]$ and $y \in \R^d$.
    As a consequence, if a reference class $\Phi$ is such that $\cT_\infty[\varphi] = 0$ for all $\varphi \in \Phi$, then a surrogate empirical risk minimizer
    \begin{equation}
        \label{eq:surrogate_erm}
        \widehat \varphi \in \argmin\limits_{\varphi \in \Phi} \left\{
        -\frac1n \sum\limits_{i = 1}^n \varphi(Y_i) + \frac1{\cK(T)} \log \cT_\infty[e^\varphi] \right\}
    \end{equation}
    satisfies \eqref{eq:erm} with $\opterr = \cO\left( (R^2 / d + M + 1) \sqrt{d} \, e^{-bT} \right)$ as $bT \rightarrow \infty$.
\end{Prop}

We defer the proof of Proposition \ref{prop:approx_erm} to Appendix \ref{sec:prop_approx_erm_proof}. A careful look at the bound \eqref{eq:oracle_inequality} reveals that if we take $bT \gtrsim \log(en)$ (with a sufficiently large hidden constant), then the gap $\opterr$ becomes negligible compared to the statistical error (in particular, it will be smaller than $(M + Ld + d) \Upsilon(n, \delta)$).
At the same time, the surrogate objective \eqref{eq:surrogate_erm} is more tractable and convenient than the log-density loss as it requires only one sample of i.i.d. Gaussian random vectors $Z_1, \dots, Z_N \sim \cN \big(m, \Sigma / (2b) \big)$ to approximate $\cT_\infty[e^\varphi]$ using Monte-Carlo approach. Moreover, in special situations $\cT_\infty[e^\varphi]$ can be computed exactly (see, for instance \citep{Korotin2024, puchkin26}).

We illustrate a mixture-based implementation of the surrogate objective \eqref{eq:surrogate_erm} on synthetic and real-world data. In both cases, we  followed the idea of \cite{Korotin2024} and \cite{Rapakoulias2024} and considered log-potentials of the form
\begin{equation}
    \label{eq:light_potential}
    \varphi(y) = \frac1{2 \sigma_T^2} \left\|\Sigma^{-1/2} y \right\|^2 + \log \left[ \; \sum\limits_{k = 1}^K \alpha_k \sfp\left(y ; r_k, \sigma_T^2 \, \Sigma^{1/2} S_k \Sigma^{1/2} \right) \right],
\end{equation}
where, for any $k \in \{1, \dots, K\}$, $\sfp(\cdot \, ; r_k, \sigma_T^2 \, \Sigma^{1/2} S_k \Sigma^{1/2})$ stands for the density of the Gaussian distribution with mean $r_k$ and covariance $\sigma_T^2 \, \Sigma^{1/2} S_k \Sigma^{1/2}$, and the triplets $\{(\alpha_k, r_k, S_k) : 1 \leq k \leq K\}$ must be estimated from data. For the Gaussian experiments below, we restricted the eigenvalues of $S_k$ to $[0.05,0.95]$. Each fitted log-potential is then bounded above, and the Gaussian mixture permits analytic evaluation of $\cT_\infty[e^\varphi]$. We used an isotropic diffusion matrix $\Sigma=\eps I_d$ and sampled $\rho_0$ from a standard Gaussian truncated to its central ball of probability $0.9$.

We used three metrics to evaluate performance of our method: sliced Wasserstein distance, maximum mean discrepancy (MMD) and energy distance. The sliced Wasserstein distance $\mathbb{SW}_1(\xi, \eta)$ between random vectors $\xi$ and $\eta$ on $\R^d$ is the average Wasserstein-1 distance between their one-dimensional projections:
\[
    \mathbb{SW}_1(\xi, \eta)
    = \E_{\theta} \, \mathbb{W}_1(\xi^\top \theta, \eta^\top \theta),
\]
where $\theta \in \R^d$ is independent of $\xi$ and $\eta$ and follows the uniform distribution over a unit sphere. The squared maximum mean discrepancy $\MMD^2(\xi, \eta)$ equals the squared distance between the averaged images of $\xi$ and $\eta$ in the reproducing kernel Hilbert space induced by a Gaussian kernel $\mathscr K(x, y) = \exp\{-\gamma \|x - y\|^2\}$, where $\gamma > 0$ is a fixed parameter and $x, y \in \R^d$:
\[
	\MMD^2(\xi, \eta) = \E \mathscr K(\xi, \xi') + \E \mathscr K(\eta, \eta') - 2 \E \mathscr K(\xi, \eta).
\]
Here $\xi'$, $\eta'$ are independent copies of $\xi$ and $\eta$, respectively. Finally, the energy distance $\ED(\xi, \eta)$ between $\xi$ and $\eta$ is defined as
\[
    \ED(\xi, \eta) = 2\E \|\xi - \eta\| - \E \|\xi - \xi'\| - \E \|\eta - \eta'\|,
\]
where $\xi'$ and $\eta'$ are independent copies of $\xi$ and $\eta$, respectively.

\subsection{Gaussian mixtures}
\label{sec:gaussian} 

Each target was an equally weighted mixture of 25 Gaussians in $\R^2$. In the standard setup, the centers lay on the grid $\{-6,-3,0,3,6\}^2$. In the irregular setup, we drew the centers uniformly from $[-9,9]^2$. Both setups used covariance $0.1I_2$. The anisotropic setup used the standard grid, with two covariance eigenvalues drawn independently from $[0.05,0.5]$ and a rotation angle drawn uniformly from $[0,\pi]$ for each component. The target parameters were held fixed during tuning and evaluation.

We fitted $K=50$ diagonal shapes with $T=10$, $\eps=0.1$, and initial shape $0.5I_2$. Each fit used 5,000 Adam updates with learning rate $0.002$ and fresh target batches of size 128. We used the uncentered mixture log-potential in the surrogate objective and added a minibatch penalty: writing $q(y)=\|y\|^2/(2\eps\sigma_T^2)$, the penalty was five times the batch mean of $(\varphi(y)-3-\max_{z\ {\rm in\ batch}}q(z))_+$. Conditional on each initial draw, we sampled the endpoint directly from the fitted Gaussian mixture.

\begin{figure*}[tb]
    \noindent
    \centering
    \begin{minipage}[c]{0.32\linewidth}
        \centering
        \includegraphics[width=\linewidth, keepaspectratio]{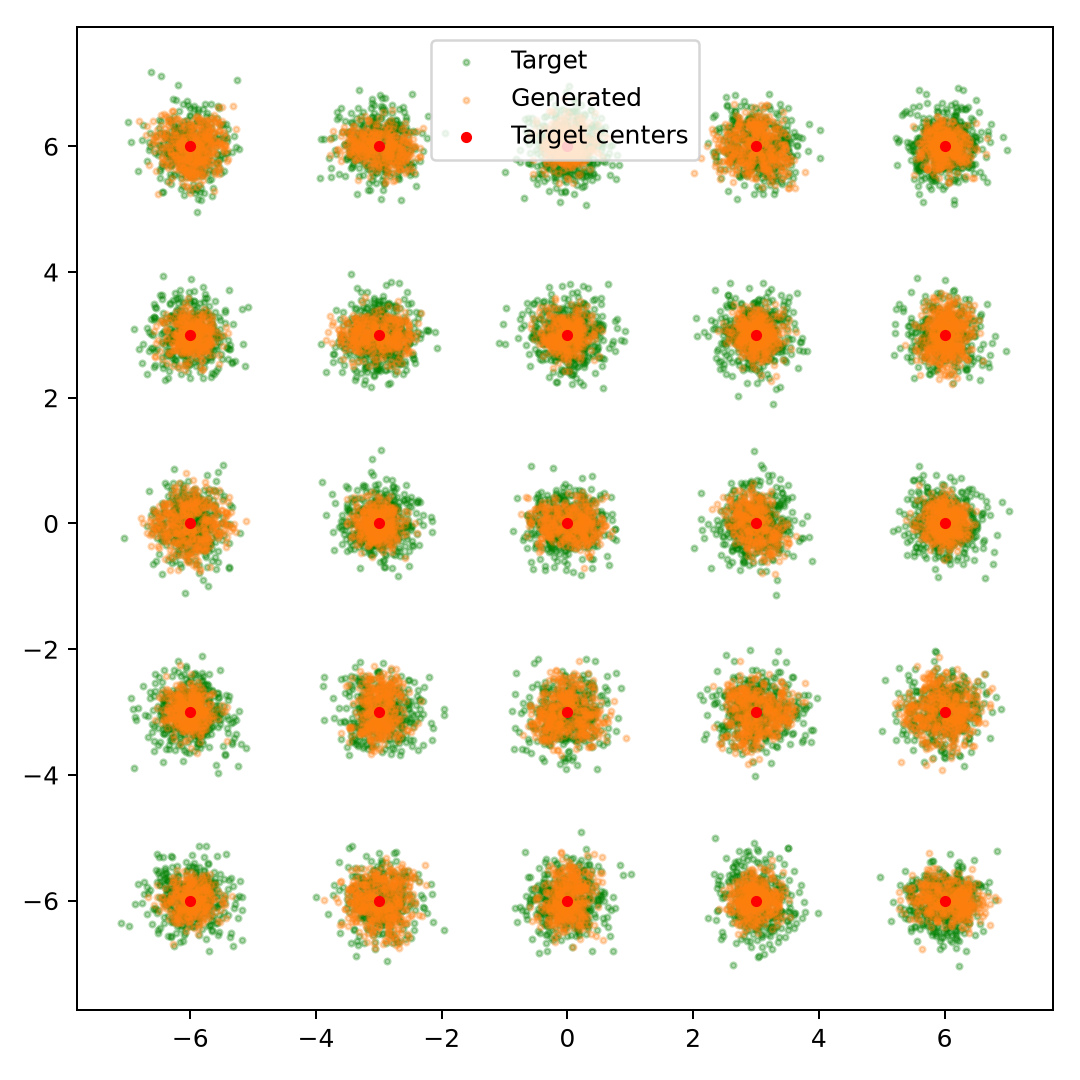}
    \end{minipage}
    \hfill
    \begin{minipage}[c]{0.32\linewidth}
        \centering
        \includegraphics[width=\linewidth, keepaspectratio]{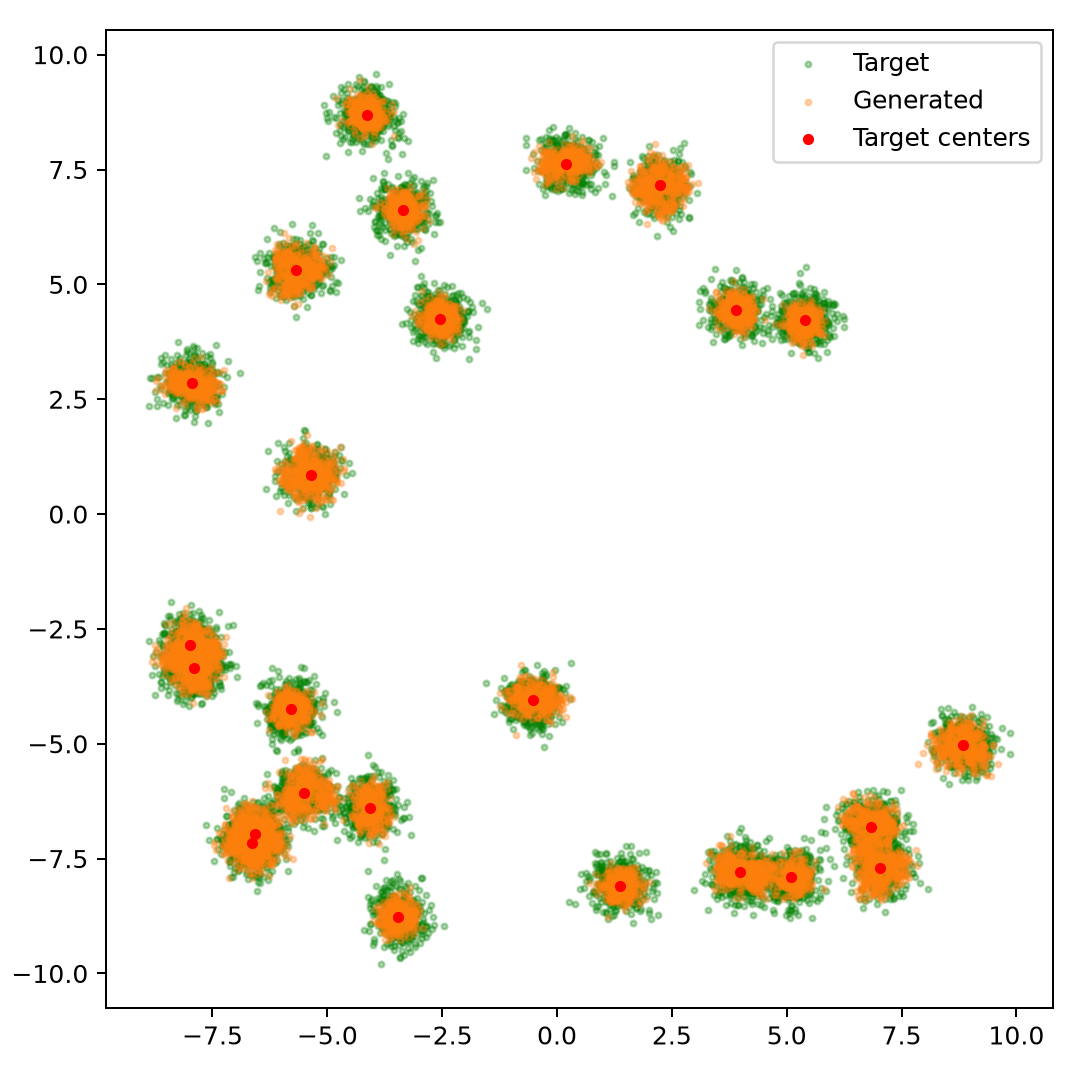}
    \end{minipage}
    \hfill
    \begin{minipage}[c]{0.32\linewidth}
        \centering
        \includegraphics[width=\linewidth, keepaspectratio]{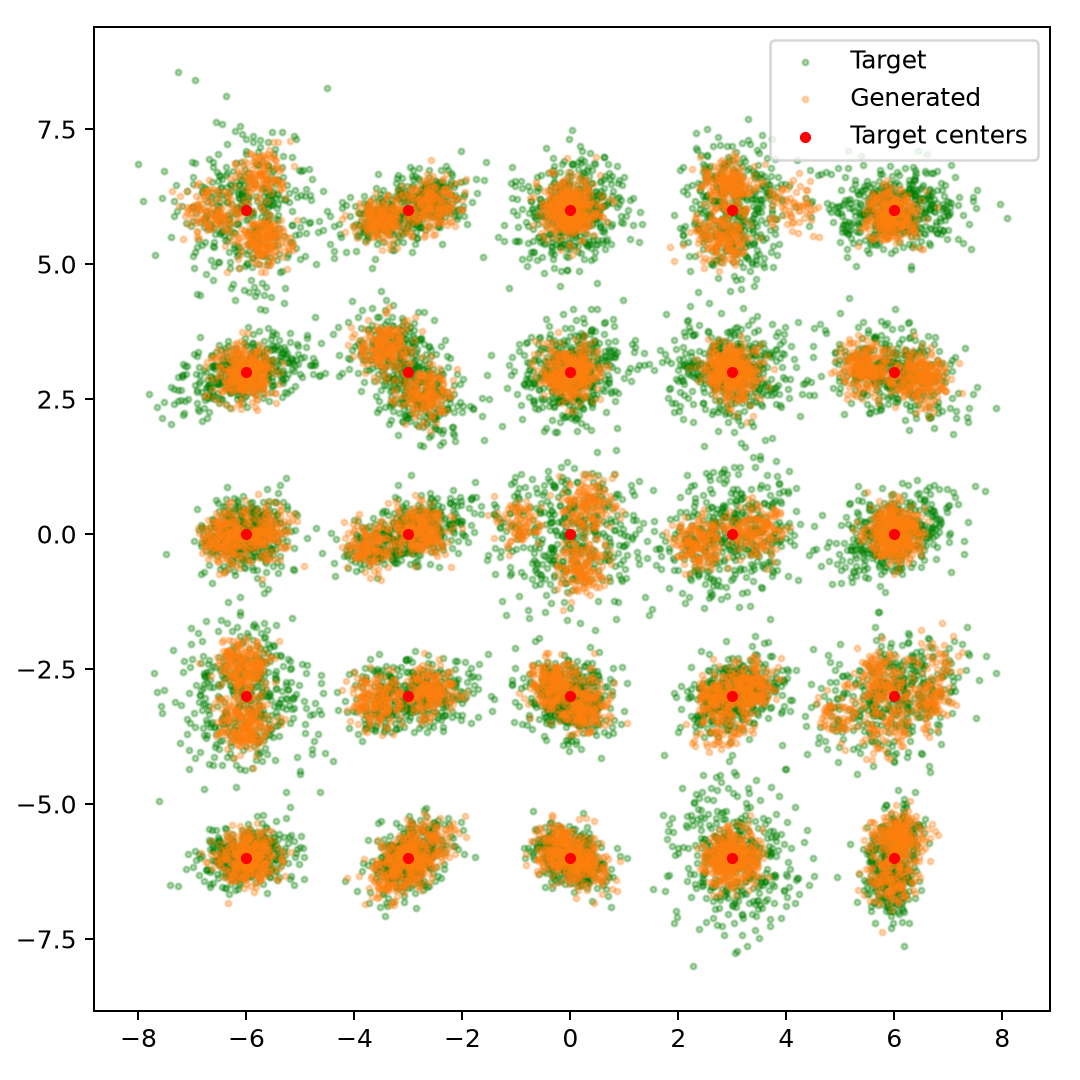}
    \end{minipage}
    \caption{Samples from the target distribution (green) and the estimated density (orange) in the experiments with Gaussian mixtures. From left to right, the panels show the standard and irregular targets with component covariance $0.1I_2$, followed by the anisotropic target. Red markers indicate target centers.}
    \label{fig:all_gauss}
\end{figure*}

For each setup, we ran 40 Optuna TPE trials \citep{Akiba2019} to select $(b,C)$ with $m=C(1,1)^\top$, minimizing the sum of sliced Wasserstein and energy distances on a common validation draw. The search ranges were $[1,3.5]\times[-0.05,0.05]$ for the standard and irregular targets and $[0.5,4]\times[-0.15,0.15]$ for the anisotropic target. The selected pairs were $(1.1415,0.0151)$, $(1.1644,-0.0278)$, and $(0.7260,-0.0710)$, respectively. We refitted each selected model with a separate seed. Figure~\ref{fig:all_gauss} shows its first evaluation draw.

We estimated sliced Wasserstein distance with 100 random unit directions. For each evaluation, we set $\gamma$ to the reciprocal of twice the median of the pooled within-sample and cross-sample squared distances. For $\MMD^2$, we excluded the within-sample kernel diagonals and used the corresponding off-diagonal pair counts. For energy distance, we used all empirical pairs.

\begin{table}[tb]
\centering
\begin{minipage}{0.95\linewidth}
\centering
\caption{Diagonal mixture fits with $K=50$. Entries give the mean and standard deviation (divisor five) across five fresh evaluations of one fit per setup, each with 10,000 generated and 10,000 target samples.}
\label{tab:gauss}
% One fit per setup; mean and SD (ddof=0) across five evaluations of 10,000 + 10,000 samples.
\begin{tabular}{lccc}
Setup & Sliced Wasserstein & $10^5\,\mathrm{MMD}^2$ & Energy distance \\
\hline
Standard & $0.090 \pm 0.016$ & $2.92 \pm 6.03$ & $0.0025 \pm 0.0009$ \\
Irregular & $0.098 \pm 0.040$ & $2.36 \pm 12.59$ & $0.0031 \pm 0.0028$ \\
Anisotropic & $0.118 \pm 0.020$ & $7.47 \pm 9.76$ & $0.0040 \pm 0.0015$ \\
\end{tabular}

\end{minipage}
\end{table}

\subsection{Photorealistic face generation}
\label{sec:Faces}

We fitted two bridges in the 512-dimensional latent space of the pretrained ALAE autoencoder \citep{Pidhorskyi2020}, using encodings of $1024\times1024$ FFHQ images \citep{Karras2019}. We used the first 60,000 of the 70,000 aligned FFHQ records as a common training pool. The child subset contained 10,762 records with a reported age below 18, excluding the missing-age code $-1$. The female-labeled subset contained all 32,816 records labeled female, irrespective of age. The two subsets shared 6,345 records.

For each subset we fitted a separate mixture with $K=50$ diagonal shapes, initialized at $0.5I_d$ and constrained to have eigenvalues in $[0.05,0.95]$. We initialized the component centers by $K$-means on 5,000 latent vectors from the subset. Both fits used $T=15$, $m=0$, and $\Sigma=0.2I_d$. We fixed $b=0.917004$ for the child subset and $b=3.363586$ for the female-labeled subset. We used the same penalized objective as in Section~\ref{sec:gaussian} and took 10,000 Adam updates with batch size 128 and learning rate $10^{-4}$.

\begin{figure*}[tb]
    \centering
    \includegraphics[width=0.485\linewidth]{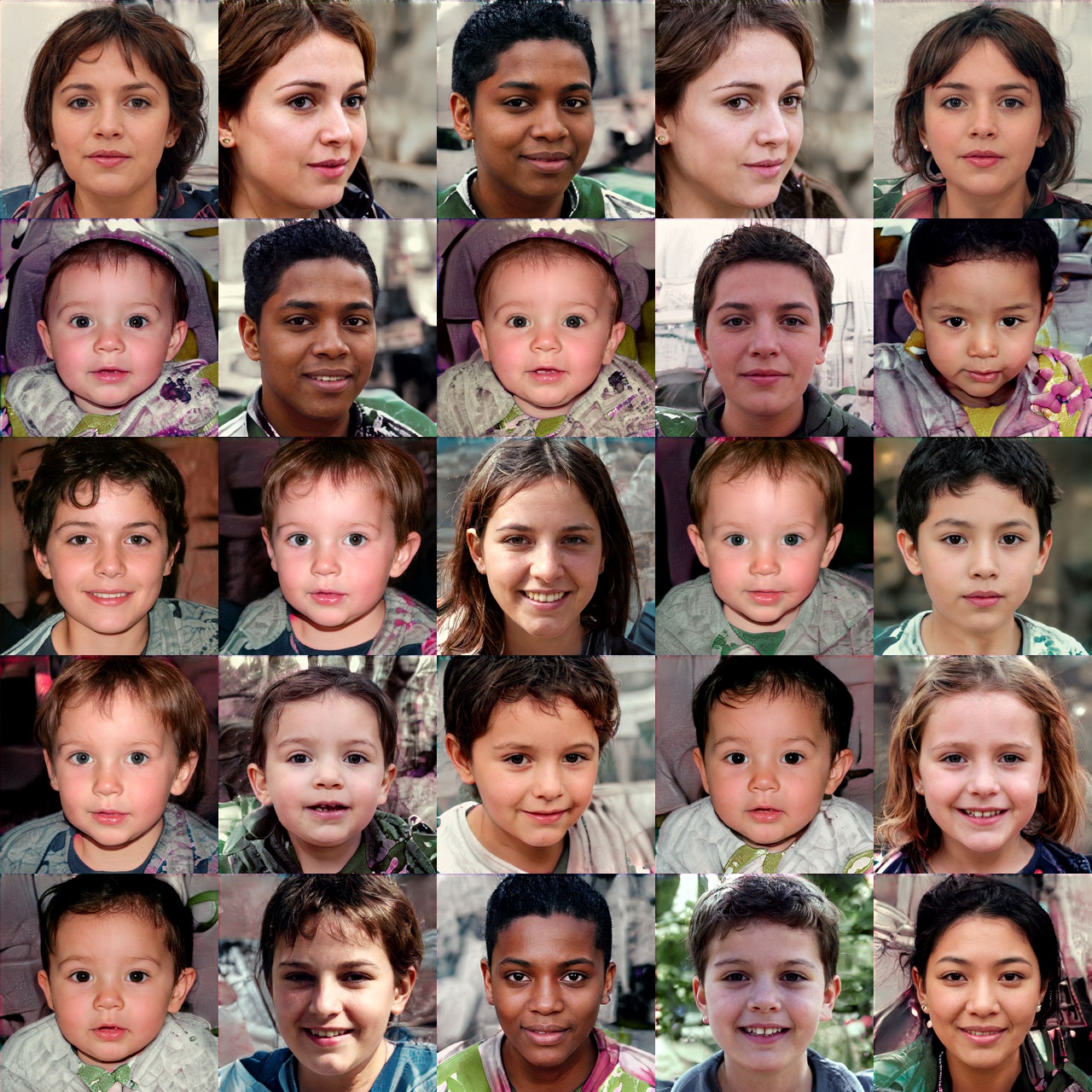}\hfill
    \includegraphics[width=0.485\linewidth]{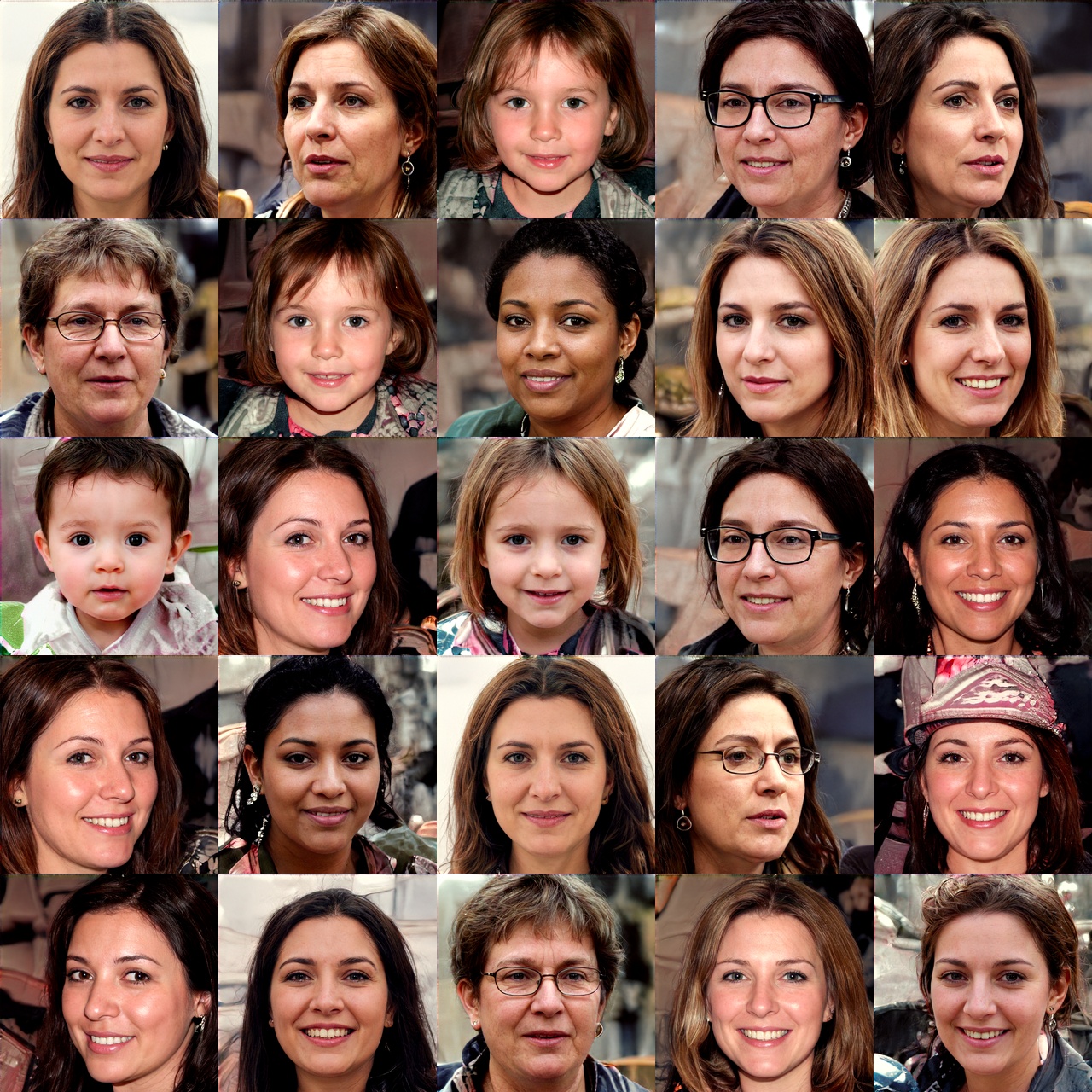}
    \caption{Generated FFHQ faces from the bridge fitted to the child subset (left) and the female-labeled subset (right). Each panel shows all 25 samples from one draw.}
    \label{fig:all_faces}
\end{figure*}

For each fit, we set the sampling seed to 42, drew 25 endpoints from the fitted bridge, and decoded them with the pretrained FFHQ checkpoint. Figure~\ref{fig:all_faces} shows both draws in full. The generated faces vary in pose and expression, although several samples within each panel look similar.

\section{Proof of Theorem \ref{th:excess_kl_bound}}
\label{sec:th_excess_kl_bound_proof}

The oracle inequality \eqref{eq:oracle_inequality} follows from the properties of the empirical process
\begin{equation}
    \label{eq:empirical_process}
    \widehat \KL\left(\rho_T^*, \rho_T^\varphi \right)
    - \KL\left(\rho_T^*, \rho_T^\varphi \right),
    \quad \varphi \in \Phi,
\end{equation}
where $\widehat \KL$ is an empirical counterpart of the KL-divergence defined in \eqref{eq:hat_kl}.
In particular, we will show that the excess loss class
\begin{equation}
    \label{eq:excess_loss_class}
    \cL = \left\{ \log \frac{\rho_T^*(Y_1)}{\rho_T^\varphi(Y_1)} : \varphi \in \Phi \right\}
\end{equation}
satisfies a Bernstein-type condition. This is a key ingredient allowing us to show that the statistical error can be smaller than $\cO(1/\sqrt{n})$ and, in favourable scenarios, the right-hand side of \eqref{eq:oracle_inequality} can be of order $\cO\big( \log^2(n) / n \big)$. However, we would like to note that the process \eqref{eq:empirical_process} is unbounded. For this reason, we cannot rely on the standard localization argument \citep{bartlett05}. Fortunately, in the recent paper \cite{yakovlev25b}, the authors extended this framework to the case of empirical processes with finite $\psi_1$-diameter (see Theorem D.3 in that paper). Despite the excess loss $\log\big( \rho_T^*(Y_1) / \rho_T^\varphi(Y_1) \big) - \KL\left(\rho_T^*, \rho_T^\varphi \right)$ has an unusual form for statistical learning, we carefully verify the conditions of Theorem D.3 from \cite{yakovlev25b}.
For convenience, we split the proof into several steps.

\smallskip

\noindent\textbf{Step 1: $\psi_1$-norm of an envelope.}
\quad
First, we have to ensure that the empirical process \eqref{eq:empirical_process} has a bounded $\psi_1$-diameter. 
We start with an observation that, according to Lemma \ref{lem:log_density_ratio_upper_bound} and Lemma \ref{lem:sub-gaussian_density}, the log-density ratio $\log\big( \rho_T^*(y) / \rho_T^\varphi(y) \big)$ admits a two-side bound
\[
    -\frac{d}2 \log\left(1 + 2 L \sigma_T^2 \right)
    - 2M - \frac{R^2 e^{-2bT}}{2 \sigma_T^2} + \log \frac{\rho_T^*(y)}{\rho_T^0(y)}
    \leq \log \frac{\rho_T^*(y)}{\rho_T^\varphi(y)}
    \leq \ttA \|y\|^2 + \ttB
\]
for all $y \in \R^d$ and all $\varphi \in \Phi$,
where the constants $\ttA = \cO(L + \sigma_T^{-2})$ and $\ttB = \cO(L + \sigma_T^{-2} + M + d + R^2 e^{-2bT} / \sigma_T^2)$ are defined in \eqref{eq:a} and \eqref{eq:b}, respectively.
It plays a significant role in the proof of Lemma \ref{lem:log_density_ratio_orlicz_norm_bound}, which claims that
\begin{align}
    \label{eq:psi_1_diameter_bound}
    \left\| \sup\limits_{\varphi \in \Phi} \left| \log \frac{\rho_T^*(Y)}{\rho_T^\varphi(Y)}  \right| \right\|_{\psi_1}
    &\notag
    \leq 2 \vee \left( 4 \max\{\ttB, 0\} + 4(d + 1) \ttA \ttv^2 \right)
    \\&\quad
    \vee \left( \frac{4M + d \log\left(1 + 2 L \sigma_T^2 \right)}{\log 2} + \frac{R^2 e^{-2bT}}{\sigma_T^2 \log 2} \right)
    \\&\notag
    \lesssim \cR + \cV,
\end{align}
where
\begin{equation}
    \label{eq:cR_cV}
    \cR = d \log(1 + 2L \sigma_T^2)
    \quad \text{and} \quad
    \cV = M + Ld + d + \frac{d + R^2 e^{-2bT}}{\sigma_T^2}.
\end{equation}
In \eqref{eq:psi_1_diameter_bound}, the hidden constant behind $\lesssim$ depends on $m$, $b$, $\rho_{\max}$, $\ttv^2$, $\|\Sigma\|$, and $\|\Sigma^{-1}\|$ only.

\smallskip

\noindent\textbf{Step 2: covering number bound.}
\quad
We proceed with an upper bound on the covering number of the excess loss class \eqref{eq:excess_loss_class} with respect to the empirical $L_2$-norm. We must ensure that it grows polynomially to apply Theorem D.3 from \cite{yakovlev25b}.  Let us fix arbitrary $\theta_0, \theta_1 \in \Theta$ and let $f_1 = \varphi_{\theta_1}$ and $f_0 = \varphi_{\theta_0}$ be the corresponding log-potentials. Let us elaborate on the empirical $L_2$-norm of the difference
\[
    \log \frac{\rho_T^*(y)}{\rho_T^{f_1}(y)} - \log \frac{\rho_T^*(y)}{\rho_T^{f_0}(y)}
    = \log \frac{\rho_T^{f_0}(y)}{\rho_T^{f_1}(y)}.
\]
Recalling that
\[
    \rho_T^\varphi(y) = e^{\varphi(y)} \int\limits_{\R^d} \frac{\sfq_T(y \,\vert\, x)}{\cT_T[e^\varphi](x)} \, \rho_0(x) \, \dd x
    \quad \text{for any $y \in \R^d$ and $\varphi \in \Phi$,}
\]
and applying the Cauchy-Schwarz inequality,
we observe that
\begin{align*}
    &
    \frac1n \sum\limits_{i = 1}^n \log^2 \left( \frac{\rho_T^{f_1}(Y_i)}{\rho_T^{f_0}(Y_i)} \right)
    \leq \frac2n \sum\limits_{i = 1}^n \big( f_1(Y_i) - f_0(Y_i) \big)^2
    \\&\qquad
    + \frac2n \sum\limits_{i = 1}^n \left( \log \int\limits_{\R^d} \frac{\sfq_T(Y_i \,\vert\, x)}{\cT_T[e^{f_0}](x)} \, \rho_0(x) \, \dd x - \log \int\limits_{\R^d} \frac{\sfq_T(Y_i \,\vert\, x)}{\cT_T[e^{f_1}](x)} \, \rho_0(x) \, \dd x \right)^2.
\end{align*}
Due to Assumption \ref{as:lipschitz_parameterization} and the Cauchy-Schwarz inequality, it holds that
\begin{align*}
    \frac2n \sum\limits_{i = 1}^n \big( f_1(Y_i) - f_0(Y_i) \big)^2
    &
    \leq \frac{2 \Lambda^2}n \sum\limits_{i = 1}^n \left(1 + \left\| \Sigma^{-1/2} (Y_i - m) \right\|^2 \right)^2 \|\theta_1 - \theta_0\|_\infty^2
    \\&
    \leq 4 \Lambda^2 \left(1 + \frac1n \sum\limits_{i = 1}^n \left\| \Sigma^{-1/2} (Y_i - m) \right\|^4 \right) \|\theta_1 - \theta_0\|_\infty^2.
\end{align*}
Moreover, according to Lemma \ref{lem:kl_bound},
\begin{align*}
    &
    \frac2n \sum\limits_{i = 1}^n \left( \log \int\limits_{\R^d} \frac{\sfq_T(Y_i \,\vert\, x)}{\cT_T[e^{f_0}](x)} \, \rho_0(x) \, \dd x - \log \int\limits_{\R^d} \frac{\sfq_T(Y_i \,\vert\, x)}{\cT_T[e^{f_1}](x)} \, \rho_0(x) \, \dd x \right)^2
    \\&
    \leq 8e^2 \Lambda^2 \left( 1 + 2 e^{-2bT} R^2 + 2(d + 2) \sigma_T^2 \right)^2
    \\&\quad
    \cdot \big(2 + M \cK(T) + \cA_{\max}(T) \cK(T) \big)^2 \|\theta_1 - \theta_0\|_\infty^2.
\end{align*}
Thus, introducing $D_n > 0$ such that
\begin{align}
    \label{eq:Dn}
    D_n^2
    &
    = 4\Lambda^2 + \frac{4 \Lambda^2}n \sum\limits_{i = 1}^n \left\| \Sigma^{-1/2} (Y_i - m) \right\|^4 
    \\&\quad\notag
    + 8e^2 \Lambda^2 \left( 1 + 2 e^{-2bT} R^2 + 2(d + 2) \sigma_T^2 \right)^2
    \Big[ 2 + \big(M + \cA_{\max}(T)\big) \cK(T) \Big]^2,
\end{align}
we deduce that
\[
    \frac1n \sum\limits_{i = 1}^n \log^2 \left( \frac{\rho_T^{f_1}(Y_i)}{\rho_T^{f_0}(Y_i)} \right)
    \leq D_n^2 \|\theta_1 - \theta_0\|_\infty^2.
\]
Since $\Theta \subseteq [-1, 1]^D$, this yields that
\[
    \sup\limits_{f_1, f_0 \in \Phi} \left\| \log \frac{\rho_T^*}{\rho_T^{f_1}} - \log \frac{\rho_T^*}{\rho_T^{f_0}} \right\|_{L_2(\sfP_n)}
    \leq 2 D_n
\]
and
\[
    \cN(\cL, L_2(\sfP_n), \eps)
    \leq \left\lceil \frac{2 D_n}{\eps} \right\rceil^D
    \leq \left(1 + \frac{2 D_n}{\eps} \right)^D
    \leq \left(\frac{4 D_n}{\eps} \right)^D
    \quad \text{for all $\eps \in (0, 2D_n]$.}
\]

\smallskip

\noindent\textbf{Step 3: bounding the loss variance.}
\quad
One of the key ingredients in the proof of Theorem \ref{th:excess_kl_bound}, which allows us to hope for faster rates of convergence than $\cO(n^{-1/2})$, is analysis of the second moment of $\log\big(\rho_T^*(Y_1) / \rho_T^\varphi(Y_1) \big)$, $\varphi \in \Phi$. On this step, we are going show that
\begin{align*}
    \E \left( \log\frac{\rho_T^*(Y_1)}{\rho_T^\varphi(Y_1)} \right)^2
    \lesssim \frac{\cV^2}n + \cV \, \KL\left(\rho_T^*, \rho_T^\varphi \right) \log(en)
    \quad \text{for all $\varphi \in \Phi$,}
\end{align*}
where $\cV$ is defined in \eqref{eq:cR_cV}.
This fact easily follows from Lemmata \ref{lem:squared_log_expectation_bound}, \ref{lem:log_density_ratio_upper_bound}, and \ref{lem:squared_log_mixture_bound} presented in Appendix \ref{sec:log-density_properties}. Indeed, Lemma \ref{lem:squared_log_expectation_bound} implies that
\begin{align}
    \label{eq:var_upper_bound}
    \E \left( \log\frac{\rho_T^*(Y_1)}{\rho_T^\varphi(Y_1)} \right)^2
    &
    \leq 2 \log(1/\omega) \KL\left(\rho_T^*, \rho_T^\varphi \right)
    \\&\quad
    + 2 \E \left( \log\frac{\omega \rho_T^*(Y_1) + (1 - \omega) \rho_T^\varphi(Y_1)}{\rho_T^\varphi(Y_1)} \right)^2
\end{align}
for any $\omega \in (0, 1/2]$. On the other hand, due to Lemma \ref{lem:log_density_ratio_upper_bound}, we have
\[
    \log \frac{\rho_T^*(y)}{\rho_T^\varphi(y)}
    \leq \ttA \|y\|^2 + \ttB
    \quad \text{for all $y \in \R^d$,}
\]
where $\ttA = \cO(L + \sigma_T^{-2})$ and $\ttB = \cO(L + \sigma_T^{-2} + M + d + R^2 e^{-2bT} / \sigma_T^2)$.
Then, according to Lemma \ref{lem:squared_log_mixture_bound}, it holds that
\[
    \E \left( \log\frac{\omega \rho_T^*(Y_1) + (1 - \omega) \rho_T^\varphi(Y_1)}{\rho_T^\varphi(Y_1)} \right)^2
    \leq 4 \omega^2 + 2e (1 + \alpha^2) e^{\ttB / \alpha} \omega^{1 / \alpha},
\]
where $\alpha = \big( (d + 2) \ttA \ttv^2 \big) \vee 1$.
Taking
\[
    \omega = \min\left\{ \frac{e^{-\ttB}}{(4n)^{\alpha}}, \frac{1}{2 \sqrt{n}} \right\},
\]
we obtain that
\[
    \max\left\{ 4 \omega^2, 4 e^{\ttB / \alpha} \omega^{1 / \alpha} \right\} \leq \frac1n,
\]
and then
\[
    \E \left( \log\frac{\omega \rho_T^*(Y_1) + (1 - \omega) \rho_T^\varphi(Y_1)}{\rho_T^\varphi(Y_1)} \right)^2
    \leq \frac{1 + e + 4 \alpha^2}n
    \lesssim \frac{d^2 (L + \sigma_T^{-2})^2}n
    \lesssim \frac{\cV^2}n.
\]
Furthermore, such choice of $\omega$ ensures that
\begin{align*}
    \log(1/\omega)
    &
    \lesssim \ttB + (1 \vee \alpha) \log(en)
    \\&
    \lesssim \left(L + \sigma_T^{-2} + M + d + \frac{R^2 e^{-2bT}}{\sigma_T^2} \right)
    + \left(L d + \frac{d}{\sigma_T^2} \right) \log(en)
    \\&
    \lesssim \cV \log(en).
\end{align*}
Summing up the last two inequalities and \eqref{eq:var_upper_bound}, we deduce that
\begin{equation}
    \label{eq:var_log-density_ratio_upper_bound}
    \E \left( \log\frac{\rho_T^*(Y_1)}{\rho_T^\varphi(Y_1)} \right)^2
    \lesssim \frac{\cV^2}n + \cV \, \KL\left(\rho_T^*, \rho_T^\varphi \right) \log(en).
\end{equation}

\smallskip

\noindent
\textbf{Step 4: uniform bound for the empirical process \eqref{eq:empirical_process}.}
\quad
Applying \cite[Theorem D.3]{yakovlev25b} to the empirical process
\[
    \frac1n \sum\limits_{i = 1}^n \log \frac{\rho_T^*(Y_i)}{\rho_T^\varphi(Y_i)} + \frac{\cV}n,
    \quad \varphi \in \Phi,
\]
we obtain that, for any $\eps \in (0, 1]$ and for any $\delta \in (0, 1)$, with probability at least $(1 - \delta)$, it holds that
\begin{align*}
    &
    \widehat \KL\left(\rho_T^*, \rho_T^\varphi \right) + \frac{\cV}n - (1 + \eps) \left( \KL\left(\rho_T^*, \rho_T^\varphi \right) + \frac{\cV}n \right)
    \lesssim \left( \frac{\cV}{\eps} + \cR \right) \Xi(n, \delta)
\end{align*}
and
\[
    \KL\left(\rho_T^*, \rho_T^\varphi \right) + \frac{\cV}n - (1 + \eps) \left( \widehat \KL\left(\rho_T^*, \rho_T^\varphi \right) + \frac{\cV}n \right)
    \lesssim \left( \frac{\cV}{\eps} + \cR \right) \Xi(n, \delta)
\]
simultaneously for all $\varphi \in \Phi$. The constants $\cV$ and $\cR$ are given by \eqref{eq:cR_cV} and
\[
    \Xi(n, \delta)
    = \frac{\log(en)}n \left( D \log(en) + D \left( \E \log^3(4D_n) \right)^{1/3} + \log(e/\delta) \right),
\]
where $D_n$ is the same as in \eqref{eq:Dn}. Let us note that, in view of Lemma \ref{lem:log_Dn_bound}, we have
\begin{align*}
    \left( \E \log^3(D_n) \right)^{1/3}
    &
    \lesssim \left\| \log D_n \right\|_{\psi_1}
    \lesssim \left( 1 \vee \log \E D_n^2 \right)
    \\&
    \lesssim 1 + \log \Lambda + \log \cK(T) + \log(d + R^2 e^{-2bT} + d \sigma_T^2)
    \\&\quad
    + \log\left( 1+ M + \frac{R^2 e^{-bT}}{\sqrt d} + \log \cK(T) \right).
\end{align*}
Since $\log\cK(T) \lesssim \sqrt{d} \log(e \vee \sigma_T^{-2}) + \sqrt{d} e^{-bT}$ due to \eqref{eq:cK}, we obtain that
\begin{align*}
    \left( \E \log^3(D_n) \right)^{1/3}
    &
    \lesssim 1 + \log \Lambda
    + \log\left( 1 + M + \frac{R^2 e^{-bT}}{\sqrt d} \right)
    + \sqrt{d} \log(e \vee \sigma_T^{-2}) + \sqrt{d} e^{-bT}
    \\&\quad
    + \log(d + R^2 e^{-2bT} + d \sigma_T^2)
    \\&
    \lesssim \log(e\Lambda) + \log(1 + M) + \sqrt{d}.
\end{align*}
Thus, we proved that
\[
    \Xi(n, \delta) \lesssim \frac{\log(en)}n \left( D \log\big(\Lambda (1 + M) n \big) + D \sqrt{d} + \log(e/\delta) \right)
    = \Upsilon(n, \delta).
\]
This yields that
\begin{equation}
    \label{eq:empirical_kl_uniform_bound}
    \widehat \KL\left(\rho_T^*, \rho_T^\varphi \right) + \frac{\cV}n - (1 + \eps) \left( \KL\left(\rho_T^*, \rho_T^\varphi \right) + \frac{\cV}n \right)
    \lesssim \left( \frac{\cV}{\eps} + \cR \right) \Upsilon(n, \delta)
\end{equation}
and
\begin{equation}
    \label{eq:kl_uniform_bound}
    \KL\left(\rho_T^*, \rho_T^\varphi \right) + \frac{\cV}n - (1 + \eps) \left( \widehat \KL\left(\rho_T^*, \rho_T^\varphi \right) + \frac{\cV}n \right)
    \lesssim \left( \frac{\cV}{\eps} + \cR \right) \Upsilon(n, \delta)
\end{equation}
simultaneously for all $\varphi \in \Phi$ on an event $\cE$ of probability at least $(1 - \delta)$.

\smallskip

\noindent
\textbf{Step 5: final bound.}
\quad
Let $\varphi^\circ$ minimize\footnote{If the minimum is not attained, one should consider a minimizing sequence.} $\KL(\rho_T^*, \rho_T^\varphi)$ over $\varphi \in \Phi$ and denote the corresponding density by $\rho_T^\circ$.
According to the definition of $\widehat \varphi$ (see \eqref{eq:erm}), it holds that
\begin{align*}
    &
    \KL\left(\rho_T^*, \widehat\rho_T \right) - (1 + \eps)^2 \, \KL\left(\rho_T^*, \rho_T^\circ \right)
    \\&
    \leq \KL\left(\rho_T^*, \widehat\rho_T \right) - (1 + \eps) \widehat \KL\left(\rho_T^*, \widehat\rho_T \right) + (1 + \eps) \, \widehat \KL\left(\rho_T^*, \rho_T^\circ \right)
    \\&\quad
    + (1 + \eps) \opterr - (1 + \eps)^2 \, \KL\left(\rho_T^*, \rho_T^\circ \right) 
    \\&
    \lesssim (1 + \eps) \opterr + (2 + \eps) \left( \frac{\eps \cV}{n} + \left( \frac{\cV}{\eps} + \cR \right) \Upsilon(n, \delta) \right)
\end{align*}
on the event $\cE$. Since $\eps \in (0, 1]$, we obtain that
\begin{align*}
    \KL\left(\rho_T^*, \widehat\rho_T \right) - \KL\left(\rho_T^*, \rho_T^\circ \right)
    &
    \lesssim \opterr + \eps \, \KL\left(\rho_T^*, \rho_T^\circ \right) +  \frac{\cV}{n} + \left( \frac{\cV}{\eps} + \cR \right) \Upsilon(n, \delta)
    \\&
    \lesssim \opterr + \eps \, \KL\left(\rho_T^*, \rho_T^\circ \right) +  \left( \frac{\cV}{\eps} + \cR \right) \Upsilon(n, \delta)
\end{align*}
on the same event.
Taking
\[
    \eps =
    \begin{cases}
        1 \land \sqrt{\frac{\cV \, \Upsilon(n, \delta)}{\KL\left(\rho_T^*, \rho_T^\circ \right)}},
        \quad \text{if $\KL\left(\rho_T^*, \rho_T^\circ \right) > 0$,}\\
        1, \quad \text{otherwise,}
    \end{cases}
\]
and using the inequalities $\cV \lesssim M + Ld + d$ and $\cR = d \log(1 + 2L\sigma_T^2) \leq 2 L d \sigma_T^2 \lesssim Ld$,
we conclude that
\begin{align*}
    \KL\left(\rho_T^*, \widehat\rho_T \right) - \KL\left(\rho_T^*, \rho_T^\circ \right)
    &
    \lesssim \opterr + \sqrt{(M + Ld + d) \, \KL\left(\rho_T^*, \rho_T^\circ \right) \, \Upsilon(n, \delta)} 
    \\&\quad
    + (M + Ld + d) \Upsilon(n, \delta)
\end{align*}
with probability at least $(1 - \delta)$. 
\myendproof

\section{Proof of Theorem \ref{th:approximation_error_relu}}
\label{sec:proof_approximation_error_relu}

In fact, we prove a stronger result. For a log-potential $\varphi$, let $\pi^\varphi$ denote the joint density of $(X_0^\varphi, X_T^\varphi)$, where the process $\{X_t^\varphi : 0 \leq t \leq T\}$ is defined in \eqref{eq:sde}. Similarly, $\pi^*$ stands for the joint density of $(X_0^*, X_T^*)$, corresponding to the optimally controlled process \eqref{eq:opt_control_sde}. 
We are going to show that, under the assumptions of the theorem, there exists a neural network $\varphi_N \in \Phi_{L_0, M_0}(K_N, W_N, S_N, B_N)$ such that
\[
    \KL(\pi^*, \pi^{\varphi_N})
    \lesssim \frac{\log^\beta(N)}{N^{2s/d}}.
\]
Then the desired claim will follow from the data-processing inequality.
Similarly to the proof of Theorem \ref{th:excess_kl_bound}, we perform our derivations into several steps for convenience.

\medskip

\noindent
\textbf{Step 1: well-posedness of $\varphi^*$ and the choice of $L_0$ and $M_0$.}
\quad
We start with an observation that, according to Lemma \ref{lem:quadratic_envelope_phi_star}, the log-potential $\varphi^*$ is well-defined, and there exists a constant $M_* \geq 0$ such that
\[
    -a_- \left\|\Sigma^{-1/2} (y - m) \right\|^2 - M_* \leq \varphi^*(y) \leq M_*
    \quad \text{for all $y \in \R^d$.}
\]
Let us take arbitrary $L_0 \geq a_-$ and $M_0 \geq M_*$ and show that the clipped and centered class $\Phi_{L_0, M_0}(K, W, S, B)$ is admissible. Indeed, by the construction,
$-L_0 \|\Sigma^{-1/2}(x - m)\|^2 - M_0 \le \psi(x) \le M_0$ for any $\psi \in \Psi_{L_0, M_0}(K, W, S, B)$. Since
$X_\infty^0 \sim \cN\big(m, \Sigma/(2b) \big)$, we have $\cT_\infty\big[ \|\Sigma^{-1/2}(\cdot - m)\|^2 \big] = d / (2b)$. Therefore
\[
    -\frac{L_0 d}{2b} - M_0
    \le
    \cT_\infty[\psi]
    \le
    M_0
    \quad \text{for all $\psi \in \Psi_{L_0, M_0}(K, W, S, B)$.}
\]
This yields that any $\varphi \in \Phi_{L_0, M_0}(K, W, S, B)$ fulfils
\[
    \varphi(x)
    =
    \psi(x) - \cT_\infty[\psi]
    \le
    2M_0 + \frac{L_0 d}{2b}
    \quad \text{and} \quad
    \varphi_\theta(x)
    \ge
    -L_0 \|\Sigma^{-1/2}(x - m)\|^2 - 2M_0.
\]
In other words, the class $\Phi_{L_0, M_0}(K, W, S, B)$ meets Assumption~\ref{as:log-potential_quadratic_growth} with $L = L_0$ and $M = 2M_0 + L_0 d / (2b)$.

\medskip

\noindent
\textbf{Step 2: approximation of $\varphi^*$.}
\quad
It remains to prove the approximation bound. By Lemma~\ref{lem:potential_smoothness}, for any $r \geq 1$, it holds that
\[
    A_r(\varphi^*)
    =
    \max_{|\alpha|\le s} \max_{\|x\| \leq r} \left\{
    r^{|\alpha|}
    \big|\partial^\alpha\varphi^*\big| \right\}
    \lesssim r^\beta,
    \quad \text{where} \quad
    \beta = \beta_\rho \vee s \vee 2.
\]
Let us fix a radius $r = r_N$, which will be specified a bit later, and apply Lemma~\ref{prop:schmidt_hieber_compact_approx} with $A_N = 1 \vee A_{r_N}(\varphi^*) \lesssim r_N^\beta$. Then there exist $K_N \lesssim \log N$, $S_N \lesssim N\log N$, $W_N \in \R^{K_N + 1}$ with $\|W_N\|_\infty \lesssim N \log N$, $B_N \lesssim 1 \vee r_N^\beta$, and a neural network $g_{N} \in \NN (K_N, W_N, S_N, B_N)$, such that
\[
    \sup_{\|x\| \leq r} |g_{N}(x) - \varphi^*(x)|
    \lesssim r_N^\beta N^{-s/d}.
\]
Let $\psi_N = \ttT_{L_0, M_0}[g_N]$ (see \eqref{eq:ttT} for the definition of $\ttT_{L_0, M_0}$) and define the clipped and centered log-potential $\varphi_N = \psi_N - \cT_\infty[\psi_N]$ corresponding to $g_{N}$. Note that our choice of $L_0$ and $M_0$ on the first step ensures that $\varphi^* = \ttT_{L_0, M_0}[\varphi^*]$. Taking into account that the operator $\ttT_{L_0, M_0}$ is non-expansive, we obtain that
\begin{align}
    \label{eq:psi_on-compact_bound}
    \max_{\|x\| \leq r_N}
    |\psi_{N}(x) - \varphi^*(x)|
    &
    = \max_{\|x\| \leq r_N} \big|\ttT_{L_0, M_0}[g_N](x) - \ttT_{L_0, M_0}[\varphi^*(x)] \big|
    \\&
    \leq \max_{\|x\| \leq r_N} \big|g_N(x) - \varphi^*(x) \big|
    \lesssim \frac{r_N^\beta}{N^{s / d}}.
\end{align}

\medskip

\noindent
\textbf{Step 3: approximation of $\rho_T^*$.}
\quad
Finally, we are going to quantify the difference between the couplings $\pi^{\varphi_N}$ and $\pi^*$, corresponding to the clipped neural network $\varphi_N$ and to the log-potential $\varphi^*$, respectively. Let us note that, due to Lemma \ref{lem:quadratic_envelope_phi_star}, they have a form
\[
    \pi^{\varphi_N}(x, y)
    =
    \frac{\rho_0(x) \, \sfq_T(y\mid x) \, e^{\varphi_N(y)}}{\cT_T[e^{\varphi_N}](x)}
    \quad \text{and} \quad
    \pi^*(x,y)
    =
    \frac{\rho_0(x) \, \sfq_T(y\mid x) \, e^{\varphi^*(y)}}{\cT_T[e^{\varphi^*}](x)}.
\]
In addition, since $\psi_N(y)$ and $\varphi_N(y)$ differ by the additive constant, $\cT_\infty[\psi_N]$, it also holds that $\pi^{\psi_N}(x, y) = \pi^{\varphi_N}(x, y)$.
For this reason, we can apply Lemma~\ref{lem:quadratic_kl_bound}, claiming that
\begin{align*}    
    \KL(\pi^*, \pi^{\varphi_{N}})
    = \KL(\pi^*, \pi^{\psi_{N}})
    &
    \le
    \|\psi_{N} - \varphi^*\|_{L_2(\rho_T^*)}^2
    +
    \|\psi_{N} -\varphi^*\|_{L_2(\rho_T^{\psi_{N}})}^2
    \\&
    = \|\psi_{N} - \varphi^*\|_{L_2(\rho_T^*)}^2
    +
    \|\psi_{N} -\varphi^*\|_{L_2(\rho_T^{\varphi_{N}})}^2.
\end{align*}
Let us elaborate on the second term in the right-hand side. According to Lemma \ref{lem:sub-gaussian_density}, for any $y \in \R^d$ we have
\[
    \rho_T^{\varphi_N}(y) \leq C_T  \rho_T^0(y),
    \quad \text{where} \quad
    C_T = \left(1 + 2 L \sigma_T^2 \right)^{d / 2}
    \exp\left\{ 2M + \frac{R^2 e^{-2bT}}{2 \sigma_T^2} \right\}.
\]
This implies that
\[
    \KL(\pi^*, \pi^{\varphi_{N}})
    \leq \|\psi_{N} - \varphi^*\|_{L_2(\rho_T^*)}^2
    + C_T \|\psi_{N} - \varphi^*\|_{L_2(\rho_T^0)}^2.
\]
In view of \eqref{eq:psi_on-compact_bound} and Assumption \ref{as:log-potential_quadratic_growth}, this yields that
\begin{align*}
    &
    \KL(\pi^*, \pi^{\varphi_{N}})
    \lesssim \frac{r_N^{2\beta}}{N^{2s / d}} + \int\limits_{\|x\| \geq r_N} \big|\psi_{N}(x) - \varphi^*(x)\big|^2 \big( \rho_T^*(x) + \rho_T^0(x) \big) \, \dd x
    \\&
    \lesssim \frac{r_N^{2\beta}}{N^{2s / d}} + \int\limits_{\|x\| \geq r_N} \left( M_0^2 + L_0^2 \left\| \Sigma^{-1/2}(x - m) \right\|^4 \right) \big( \rho_T^*(x) + \rho_T^0(x) \big) \, \dd x
    \\&
    \lesssim \frac{r_N^{2\beta}}{N^{2s / d}} + \int\limits_{\|x\| \geq r_N} \left( 1 + \|x\|^4 \right) \big( \rho_T^*(x) + \rho_T^0(x) \big) \, \dd x.
\end{align*}
It remains to note that the densities $\rho_T^*$ and $\rho_T^0$ are sub-Gaussian. Indeed, $\rho_T^*$ is sub-Gaussian due to Assumption. On the other hand, $\rho_T^0$ coincides with the density of $m_T(X_0^*) + Z$, where $X_0^*$ has a compactly supported density $\rho_0$ and $Z \sim \cN(0, \Sigma_T)$ is independent of $X_0^*$. Recalling Assumption \ref{as:bounded_support}, we obtain that
\begin{align*}
    \E_{\xi \sim \rho_T^0} e^{u^\top \xi}
    &
    = \E \exp\left\{ u^\top m_T(X_0^*) + u^\top Z \right\}
    \\&
    = \E \exp\left\{ u^\top m + e^{-bT} u^\top(X_0^* - m) + \frac{u^\top \Sigma_T u}2 \right\}
    \\&
    \leq \exp\left\{ u^\top \left( m + e^{-bT} (\E X_0^* - m) \right) + \left( \|\Sigma\| R^2 + \sigma_T^2 \|\Sigma\| \right)  \frac{\|u\|^2}2 \right\},
\end{align*}
where the last inequality holds due to Hoeffding's lemma and the fact that $\|\Sigma^{-1/2}(X_0^* - m)\| \leq R$ almost surely. This allows us to apply Lemma \ref{lem:integral_remainder_bound}, which yields that
\begin{align*}
    \int\limits_{\|x\| \geq r_N} \left( 1 + \|x\|^4 \right) \big( \rho_T^*(x) + \rho_T^0(x) \big) \, \dd x
    \lesssim (1 + r_N^4) \exp\left\{ -\frac{r_N^2}{4 (d + 1) \widetilde\ttv^2} \right\}
\end{align*}
for any $r_N \geq 2 \widetilde\ttv \sqrt{2 (d + 1)}$, where
\[
    \widetilde\ttv^2 = \ttv^2 \vee \left( \|\Sigma\| R^2 + \sigma_T^2 \|\Sigma\| \right).
\]
Taking
\[
    r_N = \max\left\{1, 2 \widetilde\ttv \sqrt{2 (d + 1)} \left(1 \vee \sqrt{\frac{s \log N}d} \right) \right\},
\]
we finally obtain that
\begin{align*}
    \KL(\pi^*, \pi^{\varphi_{N}})
    &
    \lesssim \frac{r_N^{2\beta}}{N^{2s / d}} + \int\limits_{\|x\| \geq r_N} \left( 1 + \|x\|^4 \right) \big( \rho_T^*(x) + \rho_T^0(x) \big) \, \dd x
    \\&
    \lesssim \frac{r_N^{2\beta}}{N^{2s / d}} + (1 + r_N^4) \exp\left\{ -\frac{r_N^2}{4 (d + 1) \widetilde\ttv^2} \right\}
    \lesssim \frac{\log^{\beta}(N)}{N^{2s / d}}.
\end{align*}
The assertion of the theorem follows from the data-processing inequality:
\[
    \KL(\rho_T^*, \rho_T^{\varphi_N})
    \le
    \KL(\pi^*, \pi^{\varphi_N})
    \lesssim \frac{\log^{\beta}(N)}{N^{2s / d}}.
\]
\myendproof

\section*{Statement on AI use}

We used large language models for proofreading, language editing, and proof verification only. AI tools were not used to formulate the ideas of the present research, to generate the proofs, or to develop the code for the numerical experiments.

\bibliographystyle{abbrvnat}
\bibliography{bibliography}

\newpage

\begin{appendix}

\section{Proof of Proposition \ref{prop:lambda_bound}}
\label{sec:prop_lambda_bound_proof}

Let
\[
    f_\theta(x) = -b^{(K)} + A^{(K)} \circ \relu_{b^{(K - 1)}} \circ A^{(K - 1)} \circ \ldots \circ \relu_{b^{(1)}} \circ A^{(1)} \circ x, 
    \quad x \in \R^{W_0}
\]
and
\[
    f_{\theta'}(x) = -\ttb^{(K)} + \ttA^{(K)} \circ \relu_{\ttb^{(K - 1)}} \circ \ttA^{(K - 1)} \circ \ldots \circ \relu_{\ttb^{(1)}} \circ \ttA^{(1)} \circ x, 
    \quad x \in \R^{W_0}
\]
be arbitrary neural networks from $\cF(K, W, \bbM, B)$, where $\theta$ and $\theta'$ are $|\bbM|$-dimensional vectors consisting of non-zero components of $A^{(1)} / B, b^{(1)} / B, \dots, A^{(K)} / B, b^{(K)} / B$ and of $\ttA^{(1)} / B, \ttb^{(1)} / B, \dots, \ttA^{(K)} / B, \ttb^{(K)} / B$, respectively.
For any $1 \leq i \leq j \leq K - 1$ let us introduce
\[
    f_\theta^{[i:j]}(y) = \relu_{b^{(j)}} \circ A^{(j)} \circ \ldots \circ \relu_{b^{(i)}} \circ A^{(i)} \circ y, 
    \quad y \in \R^{W_{i - 1}}.
\]
We also adopt the notation
\[
    f_\theta^{[i:K]}(y) =  -b^{(K)} + A^{(K)} \circ \relu_{b^{(K - 1)}} \circ A^{(K - 1)} \circ \ldots \circ \relu_{b^{(i)}} \circ A^{(i)} \circ y
\]
for all $y \in \R^{W_{i - 1}}$ and $1 \leq i \leq K - 1$
and
\[
    f_\theta^{[1:0]}(x) = x,
    \qquad
    f_\theta^{[K : K]}(u) = A^{(K)} u - b^{(K)},
    \qquad
    f_\theta^{[K + 1: K]}(y) = y.
\]
The functions $f_{\theta'}^{[i:j]}$ are defined in a similar way. We will rely on the following properties of these functions.

\begin{Lem}
    \label{lem:f_ij_properties}
    With the introduced notations, for any $j \in \{1, \dots, K\}$, $B \geq 1$, and $y, y' \in \R^{W_{j - 1}}$ it holds that
    \begin{equation}
        \label{eq:f_jk_Lipschitz_constant}
        \left\|f_\theta^{[j:K]}(y) - f_\theta^{[j:K]}(y') \right\|_\infty
        \leq B^{K - j + 1} \left( \prod\limits_{\ell = j - 1}^{K - 1} W_\ell \right) \|y - y'\|_\infty.
    \end{equation}
    Moreover, for any $k \in \{0, 1, \dots, K\}$ and any $x \in \R^{W_0}$ we have
    \begin{equation}
        \label{eq:f_1k_Lipschitz_constant}
        \left\|f_{\theta'}^{[1:k]}(x) \right\|_\infty \leq \frac{B (B^k \|W\|_\infty^k - 1)}{B\|W\|_\infty - 1} + B^k \|W\|_\infty^k \|x\|_\infty,
    \quad 0 \leq k \leq K.
    \end{equation}
\end{Lem}
    
We postpone the proof of Lemma \ref{lem:f_ij_properties} to Appendix \ref{sec:lem_f_ij_properties_proof} and focus on finishing the proof of the proposition. Let us define
\[
    h_{\theta, \theta'}^{[k]}(x) = f_\theta^{[k+1:K]}\left( f_{\theta'}^{[1:k]}(x) \right) - f_\theta^{[k:K]}\left( f_{\theta'}^{[1:k-1]}(x) \right),
    \quad
    1 \leq k \leq K.
\]
Lemma \ref{lem:f_ij_properties} implies that, for any $1 \leq k < K$,
\begin{align}
    \label{eq:h_k_infty_norm_bound}
    &\notag
    \left\|h_{\theta, \theta'}^{[k]}(x) \right\|_\infty
    = \left\| f_\theta^{[k+1:K]}\left( \relu_{b^{(k)}} \circ A^{(k)} f_{\theta'}^{[1:k-1]}(x) \right) - f_\theta^{[k+1:K]}\left( f_{\theta'}^{[1:k]}(x) \right) \right\|_\infty
    \\&\quad
    \leq \left( \prod\limits_{j = k}^{K - 1} B W_j \right) \left\| \relu_{b^{(k)}} \circ A^{(k)} f_{\theta'}^{[1:k-1]}(x) - \relu_{\ttb^{(k)}} \circ \ttA^{(k)} f_{\theta'}^{[1:k-1]}(x) \right\|_\infty.
\end{align}
Due to the triangle inequality, the second term in the right-hand side satisfies
\begin{align}
    \label{eq:relu_infty_norm_bound}
    &\notag
    \left\| \relu_{b^{(k)}} \circ A^{(k)} f_{\theta'}^{[1:k-1]}(x) - \relu_{\ttb^{(k)}} \circ \ttA^{(k)} f_{\theta'}^{[1:k-1]}(x) \right\|_\infty
    \\&
    \leq \|b^{(k)} - \ttb^{(k)}\|_\infty + \left\| \left( A^{(k)} - \ttA^{(k)} \right) f_{\theta'}^{[1:k-1]}(x) \right\|_\infty
    \\&\notag
    \leq \|b^{(k)} - \ttb^{(k)}\|_\infty + W_{k - 1} \left\| A^{(k)} - \ttA^{(k)} \right\|_\infty \left\| f_{\theta'}^{[1:k-1]}(x) \right\|_\infty.
\end{align}
Similarly, it holds that
\begin{align}
    \label{eq:h_kk_infty_norm_bound}
    \left\|h_{\theta, \theta'}^{[K]}(x) \right\|_\infty
    &\notag
    = \left\| A^{(K)} f_{\theta'}^{[1:K-1]}(x) - b^{(K)} - \ttA^{(K)} f_{\theta'}^{[1:K-1]}(x) + \ttb^{(K)} \right\|_\infty
    \\&
    \leq \|b^{(K)} - \ttb^{(K)}\|_\infty + W_{K - 1} \left\| A^{(K)} - \ttA^{(K)} \right\|_\infty \left\| f_{\theta'}^{[1:K-1]}(x) \right\|_\infty.
\end{align}
Summing up the inequalities \eqref{eq:h_k_infty_norm_bound}, \eqref{eq:relu_infty_norm_bound}, and \eqref{eq:h_kk_infty_norm_bound}, we obtain that
\begin{align*}
    &
    \left\| f_\theta(x) - f_{\theta'}(x) \right\|_\infty
    = \left\| \sum\limits_{k = 1}^K h_{\theta, \theta'}^{[k]}(x) \right\|_\infty
    \leq \sum\limits_{k = 1}^K \left\| h_{\theta, \theta'}^{[k]}(x) \right\|_\infty
    \\&
    \leq \sum\limits_{k = 1}^K B^{K - k} \left( \prod\limits_{j = k}^{K - 1} W_j \right) \left( \|b^{(k)} - \ttb^{(k)}\|_\infty + W_{k - 1} \left\| A^{(k)} - \ttA^{(k)} \right\|_\infty \left\| f_{\theta'}^{[1:k-1]}(x) \right\|_\infty \right)
    \\&
    \leq \|\theta - \theta'\|_\infty \sum\limits_{k = 1}^K B^{K - k} \left( \prod\limits_{j = k}^{K - 1} W_j \right) \left( B + B W_{k - 1} \left\| f_{\theta'}^{[1:k-1]}(x) \right\|_\infty \right).
\end{align*}
Invoking Lemma \ref{lem:f_ij_properties} once again, we conclude that
\begin{align*}
    &
    \left\| f_\theta(x) - f_{\theta'}(x) \right\|_\infty
    \\&
    \leq \|\theta - \theta'\|_\infty \sum\limits_{k = 1}^K B^{K - k} \|W\|_\infty^{K - k}
    \left( B + \frac{B^{k + 1} \|W\|_\infty^{k} - B^2 \|W\|_\infty}{B \|W\|_\infty - 1} + B^{k} \|W\|_\infty^{k} \|x\|_\infty \right)
    \\&
    = \|\theta - \theta'\|_\infty \sum\limits_{k = 1}^K B^{K - k} \|W\|_\infty^{K - k}
    \left( B \cdot \frac{B^{k} \|W\|_\infty^{k} - 1}{B \|W\|_\infty - 1} + B^{k} \|W\|_\infty^{k} \|x\|_\infty \right).
\end{align*}
Since $B \|W\|_\infty \geq 2$, this bound simplifies to
\begin{align*}
    \left\| f_\theta(x) - f_{\theta'}(x) \right\|_\infty
    &
    \leq \|\theta - \theta'\|_\infty \sum\limits_{k = 1}^K B^{K - k} \|W\|_\infty^{K - k}
    \\&\quad
    \cdot \left( B (B^{k} \|W\|_\infty^{k} - 1) + B^k \|W\|_\infty^{k} \|x\|_\infty \right)
    \\&
    \leq \|\theta - \theta'\|_\infty \sum\limits_{k = 1}^K B^{K} \|W\|_\infty^{K} \left( B + \|x\|_\infty \right)
    \\&
    = K \|\theta - \theta'\|_\infty \, B^{K} \|W\|_\infty^{K} \left( B + \|x\|_\infty \right).
\end{align*}
The second claim of the proposition follows from the inequality
\[
    \|x\|_\infty
    \leq \|x - m\|_\infty + \|m\|_\infty
    \leq \sqrt{\|\Sigma\|} \left\| \Sigma^{-1/2} (x - m) \right\| + \|m\|_\infty
\]
and the Cauchy-Schwarz bound
\begin{align*}
    &
    \sqrt{\|\Sigma\|} \left\| \Sigma^{-1/2} (x - m) \right\| + \|m\|_\infty + B
    \\&
    \leq \frac{\|\Sigma\|}{4B + 4\|m\|_\infty + 2\sqrt{\|\Sigma\|}} +  \left(B + \|m\|_\infty + \frac{\sqrt{\|\Sigma\|}}2 \right) \left\| \Sigma^{-1/2} (x - m) \right\|^2 + \|m\|_\infty + B
    \\&
    \leq \left(B + \|m\|_\infty + \frac{\sqrt{\|\Sigma\|}}2 \right) \left( 1 + \left\| \Sigma^{-1/2} (x - m) \right\|^2 \right).
\end{align*}
They yield that
\[
    \left( B + \|x\|_\infty \right)
    \leq \left(B + \|m\|_\infty + \frac{\sqrt{\|\Sigma\|}}2 \right) \left( 1 + \left\| \Sigma^{-1/2} (x - m) \right\|^2 \right),
\]
and then
\begin{align*}
    &
    \left\| f_\theta(x) - f_{\theta'}(x) \right\|_\infty
    \leq \Lambda \left( 1 + \left\| \Sigma^{-1/2} (x - m) \right\|^2 \right) \|\theta - \theta'\|_\infty,
\end{align*}
where
\[
    \Lambda = \Lambda(K, W, B) = K B^{K} \|W\|_\infty^{K} \left(B + \|m\|_\infty + \frac{\sqrt{\|\Sigma\|}}2\right).
\]
\myendproof

\subsection{Proof of Lemma \ref{lem:f_ij_properties}}
\label{sec:lem_f_ij_properties_proof}

The proof of the inequality \eqref{eq:f_jk_Lipschitz_constant} is quite obvious and is based on the following recurrent relation. For any $j \in \{1, \dots, K\}$ and $y, y' \in \R^{W_{j-1}}$ it holds that
\begin{align*}
    &
    \left\|f_\theta^{[j:K]}(y) - f_\theta^{[j:K]}(y') \right\|_\infty
    \\&
    \leq \left\|A^{(K)} \left( f_\theta^{[j:K-1]}(y) - f_\theta^{[j:K-1]}(y') \right) \right\|_\infty
    \\&
    = \max\limits_{1 \leq i \leq W_K} \left| \sum\limits_{\ell = 1}^{W_{K - 1}} A^{(K)}_{i\ell} \left( f_\theta^{[j:K-1]}(y) - f_\theta^{[j:K-1]}(y') \right)_\ell \right|
    \\&
    \leq \max\limits_{1 \leq i \leq W_K} \max\limits_{1 \leq \ell \leq W_{K - 1}} W_{K - 1} \left| A^{(K)}_{i\ell} \right| \left| \left( f_\theta^{[j:K-1]}(y) - f_\theta^{[j:K-1]}(y') \right)_\ell \right|
    \\&
    \leq B W_{K - 1} \left\| f_\theta^{[j:K-1]}(y) - f_\theta^{[j:K-1]}(y') \right\|_\infty.
\end{align*}
Repeating the same argument several times, we obtain that
\[
    \left\|f_\theta^{[j:K]}(y) - f_\theta^{[j:K]}(y') \right\|_\infty
    \leq B^{K - j + 1} \left( \prod\limits_{\ell = j - 1}^{K - 1} W_\ell \right) \|y - y'\|_\infty.
\]
To prove \eqref{eq:f_1k_Lipschitz_constant}, let us note that
\begin{align*}
    \left\| f_{\theta'}^{[1:k]}(x) \right\|_\infty
    &
    = \left\| \relu_{\ttb^{(k)}} \left( \ttA^{(k)} f_{\theta'}^{[1:k-1]}(x) \right) \right\|_\infty
    \leq \left\| \ttb^{(k)} \right\|_\infty + \left\| \ttA^{(k)} f_{\theta'}^{[1:k-1]}(x) \right\|_\infty
    \\&
    \leq B + B W_{k - 1} \left\| f_{\theta'}^{[1:k-1]}(x) \right\|_\infty
    \leq B + B \|W\|_\infty \left\| f_{\theta'}^{[1:k-1]}(x) \right\|_\infty
\end{align*}
for all $k < K$ and, similarly,
\begin{align*}
    \left\| f_{\theta'}^{[1:k]}(x) \right\|_\infty
    &
    = \left\| \ttA^{(k)} f_{\theta'}^{[1:k-1]}(x) - \ttb^{(k)} \right\|_\infty
    \leq \left\| \ttb^{(k)} \right\|_\infty + \left\| \ttA^{(k)} f_{\theta'}^{[1:k-1]}(x) \right\|_\infty
    \\&
    \leq B + B W_{k - 1} \left\| f_{\theta'}^{[1:k-1]}(x) \right\|_\infty
    \leq B + B \|W\|_\infty \left\| f_{\theta'}^{[1:k-1]}(x) \right\|_\infty.
\end{align*}
Since $\| f_{\theta'}^{[1:0]}(x)\|_\infty = \|x\|_\infty$, it is straightforward to check by the induction that
\[
    \left\{\| f_{\theta'}^{[1:k]}(x)\|_\infty : 0 \leq k \leq K \right\}
\]
is bounded above by $\{g_k : 0 \leq k \leq K \}$, where the latter sequence is defined by
\begin{equation}
    \label{eq:g_k}
    g_0 = \|x\|_\infty, 
    \quad g_k = B + B \|W\|_\infty \, g_{k - 1},
    \quad 1 \leq k \leq K.
\end{equation}
Solving the linear recurrence \eqref{eq:g_k}, we obtain that
\[
    \left\| f_{\theta'}^{[1:k]}(x) \right\|_\infty
    \leq g_k = \frac{B (B^k \|W\|_\infty^k - 1)}{B\|W\|_\infty - 1} + B^k \|W\|_\infty^k \|x\|_\infty,
    \quad 0 \leq k \leq K.
\]
\myendproof

\section{Proof of Proposition \ref{prop:approx_erm}}
\label{sec:prop_approx_erm_proof}

Since we are interested in the case when $bT \rightarrow \infty$, we can assume that $bT \geq \log d$ throughout the proof.
Let us fix arbitrary log-potential $\varphi : \R^d \rightarrow (-\infty, M]$ and $y \in \R^d$.
Then, according to Lemma \ref{lem:log_ou_bound_simplified}, for any $x \in \R^d$, we have a two-sided bound
\[
    e^{-\cA(x, T) \cK(T)} \left( \frac{\cT_\infty[e^{\varphi}]}{e^M} \right)^{\cK(T)}
    \leq \frac{\cT_T [e^{\varphi}](x)}{e^M} \leq e^{\cA(x, T)} \left( \frac{\cT_\infty [e^{\varphi}]}{e^M} \right)^{1 / \cK(T)},
\]
where the functions $\cK(t)$ and $\cA(x, t)$ are defined in \eqref{eq:cK} and \eqref{eq:cA}, respectively. This yields that
\begin{equation}
    \label{eq:rho_T_upper_bound}
    \rho_T^\varphi(y) \leq \frac{e^{\varphi(y)}}{e^M} \cdot \left( \frac{e^M}{\cT_\infty[e^\varphi]} \right)^{\cK(T)} \int\limits_{\R^d} e^{\cA(x, T) \cK(T)} \, \sfq_T(y \,\vert\, x) \, \rho_0(x) \, \dd x
\end{equation}
and, similarly,
\begin{equation}
    \label{eq:rho_T_lower_bound}
    \rho_T^\varphi(y) \geq \frac{e^{\varphi(y)}}{e^M} \cdot \left( \frac{e^M}{\cT_\infty[e^\varphi]} \right)^{1 / \cK(T)} \int\limits_{\R^d} e^{-\cA(x, T)} \, \sfq_T(y \,\vert\, x) \, \rho_0(x) \, \dd x.
\end{equation}
Let us introduce
\[
    F(\lambda) = \log \int\limits_{\R^d} e^{\lambda \cA(x, T)} \, \sfq_T(y \,\vert\, x) \, \rho_0(x) \, \dd x
\]
and note that
\[
    F(0)
    = \log \int\limits_{\R^d} \sfq_T(y \,\vert\, x) \, \rho_0(x) \, \dd x
    = \log \rho_T^0(y).
\]
Then, in view of \eqref{eq:rho_T_upper_bound} and \eqref{eq:rho_T_lower_bound}, we have
\[
    \log \frac{\rho_T^\varphi(y)}{\rho_T^0(y)} - \varphi(y) + M - \frac1{\cK(T)} \log \frac{e^M}{\cT_\infty[e^\varphi]}
    \geq F\left(-1 \right) - F(0)
\]
and
\begin{align*}
    \log \frac{\rho_T^\varphi(y)}{\rho_T^0(y)} - \varphi(y) + M - \frac1{\cK(T)} \log \frac{e^M}{\cT_\infty[e^\varphi]}
    &
    \leq \left(\cK(T) - \frac1{\cK(T)} \right) \log \frac{e^M}{\cT_\infty[e^\varphi]}
    \\&\quad
    + F\big(\cK(T) \big) - F(0).
\end{align*}
Since, due to the definition of $\cK(t)$,
\[
    \cK(T) - \frac1{\cK(T)}
    \lesssim \log \cK(T) \lesssim \sqrt{d} \, e^{-bT},
\]
we deduce that
\begin{align*}
    &
    \left| \log \frac{\rho_T^\varphi(y)}{\rho_T^0(y)} - \varphi(y) + M - \frac1{\cK(T)} \log \frac{e^M}{\cT_\infty[e^\varphi]} \right|
    \\&
    \lesssim \max\left\{ F\big(\cK(T) \big) - F(0), F(0) - F\left(-1 \right) \right\} + \sqrt{d} \, e^{-bT} \log \frac{e^M}{\cT_\infty[e^\varphi]}.
\end{align*}
Then the expansion \eqref{eq:log_rho_T_representation} follows from the analysis of derivatives of the function $F$. Let us note that, for any $\lambda \in \R$
\[
    F'(\lambda)
    = \left( \; \int\limits_{\R^d} \cA(x, T) e^{\lambda \cA(x, T)} \, \sfq_T(y \,|\, x) \, \rho_0(x) \, \dd x \right) \Big/ \left( \; \int\limits_{\R^d} e^{\lambda \cA(x, T)} \, \sfq_T(y \,|\, x) \, \rho_0(x) \, \dd x \right).  
\]
Since $\|\Sigma^{-1/2} (x - m)\| \leq R$ on the support of $\rho_0$, we have
\[
    0
    \leq F'(\lambda)
    \leq \max\limits_{x \in \supp(\rho_0)} \cA(x, T) 
    \leq \frac{b e^2 R^2}{\sqrt{d}} \arcsin\left( e^{-bT} \right) + 2 \log \cK(T)
    \lesssim \left( \frac{R^2}{\sqrt d} + \sqrt{d} \right) e^{-bT}.
\]
This immediately implies that
\[
    \max\left\{ F\big(\cK(T) \big) - F(0), F(0) - F\left(-1 \right) \right\}
    \lesssim \left( \frac{R^2}{\sqrt d} + \sqrt{d} \right) e^{-bT},
\]
and \eqref{eq:log_rho_T_representation} follows. The second claim of the proposition is a straightforward consequence of the representation \eqref{eq:log_rho_T_representation}. If $\cT_\infty[\varphi] = 0$ for all $\varphi \in \Phi$ and $\widehat\varphi$ satisfies \eqref{eq:surrogate_erm}, then the corresponding endpoint density $\widehat\rho_T$ fulfils
\begin{align*}
    \frac1n \sum\limits_{i = 1}^n \log\frac{\rho_T^\varphi(Y_i)}{\widehat\rho_T(Y_i)}
    &
    = -\frac1n \sum\limits_{i = 1}^n \widehat\varphi(Y_i) + \frac1{\cK(T)} \log \cT_\infty[e^{\widehat\varphi}] + \frac1n \sum\limits_{i = 1}^n \varphi(Y_i) - \frac1{\cK(T)} \log \cT_\infty[e^\varphi]
    \\&\quad
    + \cO\left( \left(\frac{R^2}d + 1 + \log\frac{e^M}{\cT_\infty[e^\varphi]} + \log\frac{e^M}{\cT_\infty[e^{\widehat\varphi}]} \right) \sqrt{d} \, e^{-bT} \right)
    \\&
    \leq \cO\left( (R^2 / d + M + 1) \sqrt{d} \, e^{-bT} \right).
\end{align*}
\myendproof

\section{Proof of Corollary \ref{cor:combined_estimation_approximation_rate}}

Let us fix $n \in \mathbb N$ and apply Theorem \ref{th:approximation_error_relu} with $N = 2 \vee n^{d / (2s + d)}$. According to this theorem, we can choose such $L_0, M_0 > 0$ and a class $\Phi_n = \Phi_{L_0, M_0}(K_n, W_n, S_n, B_n)$ with $K_n \lesssim \log(en)$
\[
    \|W_n\|_\infty \lesssim n^{d / (2s + d)} \log(en),
    \quad
    S_n \lesssim n^{d / (2s + d)} \log(en),
    \quad \text{and} \quad
    B_n \lesssim \log^{\beta / 2}(en),
\]
that $\Phi_n$ meets Assumption \ref{as:log-potential_quadratic_growth} with $L = L_0$, $M = 2 M_0 + d L_0 / (2b)$ and that there is $\varphi_n \in \Phi_n$ satisfying the inequality
\[
    \KL(\rho_T^*, \rho_T^{\varphi_n})
    \lesssim \frac{\log^{\beta}(en)}{n^{2s / (2s + d)}}.
\]
To establish the desired rate of convergence, we have to ensure that the estimation error is of the same order.

Let us note that
\[
   \NN(K_n, W_n, S_n, B_n) = \bigcup\limits_{\bbM : |\bbM| \leq S_n} \cF(K_n, W_n, \bbM, B_n),
\]
where the class $\cF(K, W, \bbM, B)$ is defined in \eqref{eq:nn_class_with_mask}, and the union is taken over all collections of matrices $\bbM = \{ M^{(k)} : 1 \leq k \leq K\}$, such that
\[
    M^{(k)} \in \{0, 1\}^{W_{n, k} \times (W_{n, k - 1} + 1)}
    \quad \text{for all $1 \leq k \leq K$}
    \quad \text{and} \quad 
    |\bbM| = \sum\limits_{k = 1}^K \|M_k\|_0 \leq S_n.
\]
For each mask $\bbM$ we can define the corresponding class of truncated and centered neural networks. Then the class $\Phi_n$ admits the following representation:
\[
    \Phi_n = \bigcup\limits_{\bbM : |\bbM| \leq S_n} \left\{ \ttT_{L_0, M_0}[f](x) - \cT_\infty\big[ \ttT_{L_0, M_0}[f] \big] : f \in \cF(K_n, W_n, \bbM, B_n) \right\}.
\]
The total number of such $\bbM$ equals
\[
    \cN_n = \sum\limits_{j = 0}^{S_n} \binom{P_n}{j} \lesssim \left( \frac{e P_n}{S_n} \right)^{S_n},
    \quad \text{where} \quad
    P_n = \sum\limits_{k = 1}^{K_n} W_{n, k} (W_{n, k - 1} + 1).
\]

Due to Proposition \ref{prop:lambda_bound}, for every mask $\bbM$ of size $|\bbM| \leq S_n$ the class $\cF(K_n, W_n, \bbM, B_n)$ fulfils Assumption \ref{as:lipschitz_parameterization} with $\Lambda(K_n, W_n, B_n)$ given by \eqref{eq:lambda_nn}. Since the operator $\ttT_{L_0, M_0}$ is non-expansive and $\cT_\infty$ corresponds to the expectation with respect to $\cN\big(m, \Sigma / (2b) \big)$, the set
\[
    \left\{ \ttT_{L_0, M_0}[f](x) - \cT_\infty\big[ \ttT_{L_0, M_0}[f] \big] : f \in \cF(K_n, W_n, \bbM, B_n) \right\}
\]
satisfies Assumption \ref{as:lipschitz_parameterization} as well with $\Lambda_n \leq \big(2 + d / (2b) \big) \Lambda(K_n, W_n, B_n)$, such that
\begin{align*}
    \log \Lambda_n
    &
    \leq \log\left( 2 + \frac{d}{2b} \right) + \log K_n + K_n \log \big( B_n \|W_n\|_\infty \big)
    \\&\quad
    + \log\left(B_n + \|m\|_\infty + \frac{\sqrt{\|\Sigma\|}}2\right)
    \\&
    \lesssim \log^2(en).
\end{align*}
Thus, we can carry out all the derivations as in the proof of Theorem \ref{th:excess_kl_bound}. In particular, similarly to \eqref{eq:empirical_kl_uniform_bound} and \eqref{eq:kl_uniform_bound}, we obtain that, for any $\bbM$, $|\bbM| \leq S_n$, and any $\delta \in (0, 1)$, with probability at least $(1 - \delta / \cN_n)$, it holds that
\begin{equation}
    \label{eq:empirical_kl_bound_masked}
    \widehat \KL\left(\rho_T^*, \rho_T^\varphi \right) - 2 \, \KL\left(\rho_T^*, \rho_T^\varphi \right)
    \lesssim \left( S_n \log(n \Lambda_n) + \log\frac{\cN_n}{\delta} \right) \frac{\log(en)}n
\end{equation}
and
\begin{equation}
    \label{eq:kl_bound_masked}
    \KL\left(\rho_T^*, \rho_T^\varphi \right) - 2 \, \widehat \KL\left(\rho_T^*, \rho_T^\varphi \right) 
    \lesssim \left( S_n \log(n \Lambda_n) + \log\frac{\cN_n}{\delta} \right) \frac{\log(en)}n
\end{equation}
simultaneously for all
\[
    \varphi \in \left\{ \ttT_{L_0, M_0}[f](x) - \cT_\infty\big[ \ttT_{L_0, M_0}[f] \big] : f \in \cF(K_n, W_n, \bbM, B_n) \right\}
\]
As before, $\widehat\KL$ stands for the empirical counterpart of the KL-divergence (see \eqref{eq:hat_kl}). Applying the union bound, we obtain that, for any $\eps \in (0, 1]$ and $\delta \in (0, 1)$, the inequalities \eqref{eq:empirical_kl_bound_masked} and \eqref{eq:kl_bound_masked} hold with probability $(1 - \delta)$ uniformly for all $\varphi \in \Phi_n$.
Hence, on this event, the endpoint marginal density $\widehat\rho_T$, corresponding to the exact empirical risk minimizer $\widehat\varphi$, satisfies
\begin{align*}
    \KL\left(\rho_T^*, \widehat\rho_T \right) - 4 \, \KL\left(\rho_T^*, \rho_T^{\varphi_n} \right)
    &
    \leq \KL\left(\rho_T^*, \widehat\rho_T \right) - 2 \, \widehat \KL\left(\rho_T^*, \widehat\rho_T \right)
    \\&\quad
    + 2 \, \widehat \KL\left(\rho_T^*, \rho_T^{\varphi_n} \right)
    - 4 \KL\left(\rho_T^*, \rho_T^{\varphi_n} \right) 
    \\&
    \lesssim \left( S_n \log(n \Lambda_n) + \log\frac{\cN_n}{\delta} \right) \frac{\log(en)}n.
\end{align*}
Since $\KL(\rho_T^*, \rho_T^{\varphi_n}) \lesssim n^{-2s / (2s + d)} \log^{\beta}(en)$
and
\begin{align*}
    S_n \log(n \Lambda_n) + \log\frac{\cN_n}{\delta}
    &
    \lesssim S_n \log(n \Lambda_n) + S_n \log\frac{e K_n \|W_n\|_\infty^2}{S_n} + \log\frac1{\delta}
    \\&
    \lesssim n^{d / (2s + d)} \log^3(en) + \log\frac1{\delta},
\end{align*}
we obtain that, with probability at least $(1 - \delta)$, it holds that
\[
    \KL(\rho_T^*, \widehat\rho_T) \lesssim  \frac{\log^{\beta \vee 4}(en)}{n^{2s / (2s + d)}} + \frac{\log(en) \log(1 / \delta)}n.
\]
\myendproof

\section{Properties of the Ornstein--Uhlenbeck operator}

This section contains auxiliary results on properties of the Ornstein--Uhlenbeck operator used in the proof of Theorem \ref{th:excess_kl_bound}. The first one helps us to establish that under Assumption \ref{as:log-potential_quadratic_growth} $-\log \rho_T^\varphi(y)$ grows as fast as $\cO(\|y\|^2)$.

\begin{Lem}
    \label{lem:log-density_quadratic_growth}
    Grant Assumption \ref{as:bounded_support}.
    Let $M \in \R$ and let $\varphi : \R^d \rightarrow \R$ be such that
    \[
        \cT_\infty[\varphi] = 0
        \quad \text{and} \quad
        \varphi(x) \leq M
        \quad \text{for all $x \in \R^d$.}
    \]
    Then the density
    \[
        \rho_T^\varphi(y) = \int\limits_{\R^d} \frac{e^{\varphi(y)} \sfq_T(y \,\vert\, x)}{\cT_T [e^{\varphi}](x)} \rho_0(x) \, \dd x,
    \]
    corresponding to the log-potential $\varphi$, satisfies
    \[
        \varphi(y) - \log \rho_T^\varphi(y)
        \leq \frac{1}{\sigma_T^2} \left\| \Sigma^{-1/2} (y - m) \right\|^2 + \Delta,
    \]
    where
    \[
        \Delta = M + \frac{d}2 \log(2\pi \sigma_T^2) + \frac{\log \det(\Sigma)}2 + \frac{R^2 e^{-2bT}}{\sigma_T^2}.
    \]
\end{Lem}
The proof of Lemma \ref{lem:log-density_quadratic_growth} is deferred to Appendix \ref{sec:lem_log-density_quadratic_growth_proof}. Under Assumptions \ref{as:sub_gaussian_density} and \ref{as:log-potential_quadratic_growth}, it helps us to conclude that $\log\big( \rho_T^*(Y_1) / \rho_T^\varphi(Y_1) \big)$ is a sub-exponential random variable. The next auxiliary lemma shows that $\log \rho_T^\varphi(y)$ changes smoothly with respect to $\varphi$. This is a key technical result which allows us to derive a uniform Bernstein-type inequality relating $\KL(\rho_T^*, \rho_T^\varphi)$ and its empirical counterpart.

\begin{Lem}
\label{lem:kl_bound}
    Grant Assumptions \ref{as:bounded_support}, \ref{as:log-potential_quadratic_growth}, and \ref{as:lipschitz_parameterization}. Let us take arbitrary $\theta_0, \theta_1 \in \Theta$, $\theta_0 \neq \theta_1$, and let
    $f_0 = \varphi_{\theta_0}$ and $f_1 = \varphi_{\theta_1}$ be the corresponding log-potentials from the reference class $\Phi$. Let $\sfq_T(y \,\vert\, x)$ stand for the density of the Gaussian distribution $\cN \big(m_t(x), \Sigma_T \big)$, where $m_T(x) = m + e^{-bT}(x - m)$ and $\Sigma_T = \sigma_T^2 \Sigma$.
    Then for any $y \in \R^d$ it holds that
    \begin{align*}
        &
        \left| \log \int\limits_{\R^d} \frac{\sfq_T(y \,\vert\, x)}{\cT_T[e^{f_1}](x)} \, \rho_0(x) \dd x - \log \int\limits_{\R^d} \frac{\sfq_T(y \,\vert\, x)}{\cT_T [e^{f_0}](x)} \, \rho_0(x) \dd x \right|
        \\&
        \leq 2e \Lambda \left( 1 + 2 e^{-2bT} R^2 + 2(d + 2) \sigma_T^2 \right)
        \\&\quad
        \cdot \big(2 + M \cK(T) + \cA_{\max}(T) \cK(T) \big) \|\theta_1 - \theta_0\|_\infty,
    \end{align*}
    where
    \begin{equation}
        \label{eq:cA_max}
        \cA_{\max}(t) = \frac{b e^2 R^2}{\sqrt{d}} \, \arcsin(e^{-bt}) + 2 \log \cK(t),
    \end{equation}
    and the function $\cK(t)$ is defined in \eqref{eq:cK}.
\end{Lem}

Let us note that  Lemma \ref{lem:kl_bound} provides a more subtle result than Lemma \ref{lem:log-density_quadratic_growth}. We provide the proof of Lemma \ref{lem:kl_bound} in Appendix \ref{sec:lem_kl_bound_proof}.
In contrast to Lemma \ref{lem:log-density_quadratic_growth}, the proof of Lemma \ref{lem:kl_bound} is more technical and relies on non-trivial properties of the Ornstein--Uhlenbeck operator $\cT_t$. In particular, it relies on the following result about asymptotic behaviour of $\cT_t$.

\begin{Lem}
    \label{lem:log_ou_bound}
    Let $f : \R^d \rightarrow \R$ and $g: \R^d \rightarrow \R$ be measurable functions. Assume that there exists $M \in \R$ and some non-negative constants $A$, $B$, $A + B > 0$, and $\alpha \geq 1$ such that
    \begin{equation}
        \label{eq:f_g_conditions}
        f(x) \leq M,
        \quad
        0 \leq g(x) \leq A \left\|\Sigma^{-1/2}(x - m) \right\|^\alpha + B
        \quad \text{for all $x \in \R^d$.}
    \end{equation}
    Let us fix arbitrary $x \in \R^d$ and $t > 0$ and introduce
    \begin{align}
        \label{eq:G}
        G(x)
        &\notag
        = B e^M + 2^{\alpha - 1} A e^M \left\| \Sigma^{-1/2}(x - m) \right\|^{\alpha}
        \\&\quad
        + 4^{\alpha - 1} A e^M (2b)^{-\alpha/2} \left((10 \alpha \sqrt{d})^\alpha + d^\alpha \right),
    \end{align}
    \[
        \cK(t) = \big(1 - e^{-2bt} \big)^{-5e^2 \sqrt{d}} \cdot \exp\left\{ 2e^2 \sqrt{d} \arcsin(e^{-bt}) \right\},
    \]
    and
    \[
        \cA(x, t) = \frac{b e^2}{\sqrt{d}} \left\|\Sigma^{-1/2} (x - m) \right\|^2 \arcsin(e^{-bt}) + 2 \log \cK(t).
    \]
    Then, it holds that
    \begin{equation}
        \label{eq:ou_two-side_bound}
        e^{-\cA(x, t) \cK(t)} \left( \frac{\cT_\infty \big[g e^{f} \big]}{G(x)} \right)^{\cK(t)}
        \leq \frac{\cT_t \big[g e^{f} \big](x)}{G(x)} \leq e^{\cA(x, t)} \left( \frac{\cT_\infty \big[g e^{f} \big]}{G(x)} \right)^{1 / \cK(t)}
    \end{equation}
\end{Lem}

The proof of Lemma \ref{lem:log_ou_bound} is postponed to Appendix \ref{sec:lem_log_ou_bound_proof}. The key ingredients of the proof are the Kolmogorov-Fokker-Planck equation, a Gronwall-type bound from Lemma \ref{lem:ode_bound} and a sharp bound on the $L_p$-norm of a centered chi-squared random variable (see Lemma \ref{lem:chi-squared_moments_bound}).

Finally, we conclude this section with Lemma \ref{lem:sub-gaussian_density}, claiming that Schr\"odinger potentials satisfying Assumption \ref{as:log-potential_quadratic_growth} and the Ornstein--Uhlenbeck kernel induce sub-Gaussian densities. Its proof follows immediately from Lemma \ref{lem:ou_lower_bound}. The details are postponed to Appendices \ref{sec:lem_sub-gaussian_density_proof} and \ref{sec:lem_ou_lower_bound_proof}.
\begin{Lem}
    \label{lem:sub-gaussian_density}
    Grant Assumptions \ref{as:bounded_support} and \ref{as:log-potential_quadratic_growth}. Then, for any $\varphi \in \Phi$, the probability density
    \[
        \rho_T^\varphi(y) = \int\limits_{\R^d} \frac{e^{\varphi(y)} \, \sfq_T(y \,\vert\, x)}{\cT_T[e^\varphi](x)} \, \rho_0(x) \, \dd x
    \]
    satisfies the inequality $\rho_T^\varphi(y) \leq C_T  \rho_T^0(y)$ for all $y \in \R^d$, where
    \[
        C_T = \left(1 + 2 L \sigma_T^2 \right)^{d / 2}
        \exp\left\{ 2M + \frac{R^2 e^{-2bT}}{2 \sigma_T^2} \right\}.
    \] 
\end{Lem}

\begin{Lem}
    \label{lem:ou_lower_bound}
    Under Assumptions \ref{as:bounded_support} and \ref{as:log-potential_quadratic_growth}, for any $\varphi \in \Phi$ and any $x \in \supp(\rho_0)$ it holds that
    \[
        \cT_T[e^\varphi](x) \geq (1 + 2L \sigma_T^{2})^{-d/2} \, \exp\left\{ -M - \frac{L R^2 \sigma_T^{-2} e^{-2bT}}{2L + \sigma_T^{-2}} \right\}.
    \]
\end{Lem}

\subsection{Proof of Lemma \ref{lem:log-density_quadratic_growth}}
\label{sec:lem_log-density_quadratic_growth_proof}

The bound \eqref{eq:ou_two-side_bound} is trivial when $g(x) = 0$ almost everywhere. For this reason, we restrict our attention on the situation when $g$ is positive on a set of non-zero measure. In this case, we have $\cT_t[g e^f](x) > 0$ for all $x \in \R^d$ and $t > 0$.
Let us note that
\[
    \cT_T [e^{\varphi}](x) \leq \cT_T [e^M] = e^M.
\]
This yields that
\[
    \varphi(y) - \log \rho_T^\varphi(y)
    = -\log \int\limits_{\R^d} \frac{\sfq_T(y \,\vert\, x)}{\cT_T [e^{\varphi}](x)} \rho_0(x) \, \dd x
    \leq M - \log \int\limits_{\R^d} \sfq_T(y \,\vert\, x) \rho_0(x) \, \dd x.
\]
Due to the Cauchy-Schwarz inequality and Assumption \ref{as:bounded_support}, for any $x \in \supp(\rho_0)$ it holds that
\begin{align*}
    \frac12 \left\| \Sigma_T^{-1/2} \big( y - m_T(x) \big) \right\|^2
    &
    \leq \left\| \Sigma_T^{-1/2} (y - m) \right\|^2
    + e^{-2bT} \left\| \Sigma_T^{-1/2} (x - m) \right\|^2
    \\&
    = \frac{1}{\sigma_T^2} \left\| \Sigma^{-1/2} (y - m) \right\|^2
    + \frac{e^{-2bT}}{\sigma_T^2} \left\| \Sigma^{-1/2} (x - m) \right\|^2
    \\&
    \leq \frac{1}{\sigma_T^2} \left\| \Sigma^{-1/2} (y - m) \right\|^2
    + \frac{R^2 e^{-2bT}}{\sigma_T^2}.
\end{align*}
This yields that
\begin{align*}
    \sfq_T(y \,\vert\, x)
    &
    = (2\pi)^{-d/2} \det(\Sigma_T)^{-1/2} \exp\left\{ - \frac12 \left\| \Sigma_T^{-1/2} \big( y - m_T(x) \big) \right\|^2 \right\} 
    \\&
    \geq (2\pi)^{-d/2} \det(\Sigma_T)^{-1/2} \exp\left\{ -\frac{1}{\sigma_T^2} \left\| \Sigma^{-1/2} (y - m) \right\|^2
    - \frac{R^2 e^{-2bT}}{\sigma_T^2} \right\},
\end{align*}
and then
\begin{align*}
    \varphi(y) - \log \rho_T^\varphi(y)
    &
    \leq M + \frac{d \log(2\pi)}2 + \frac{\log \det(\Sigma_T)}2
    + \frac{R^2 e^{-2bT}}{\sigma_T^2} + \frac{1}{\sigma_T^2} \left\| \Sigma^{-1/2} (y - m) \right\|^2
    \\&
    = M + \frac{d}2 \log(2\pi \sigma_T^2) + \frac{\log \det(\Sigma)}2
    + \frac{R^2 e^{-2bT}}{\sigma_T^2} + \frac{1}{\sigma_T^2} \left\| \Sigma^{-1/2} (y - m) \right\|^2.
\end{align*}

\myendproof

\subsection{Proof of Lemma \ref{lem:kl_bound}}
\label{sec:lem_kl_bound_proof}

Let us fix an arbitrary $y \in \R^d$ and consider
\[
    F(s) = \log \int\limits_{\R^d} \frac{\sfq_T(y \,\vert\, x)}{\cT_T[e^{f_s}](x)} \, \rho_0(x) \dd x,
    \quad s \in [0, 1],
\]
where, for any $s \in [0, 1]$, we introduced $f_s(x) = s f_1(x) + (1 - s) f_0(x)$ for brevity.
Then it is straightforward to observe that
\[
    \log \int\limits_{\R^d} \frac{\sfq_T(y \,\vert\, x)}{\cT_T[e^{f_1}](x)} \, \rho_0(x) \dd x - \log \int\limits_{\R^d} \frac{\sfq_T(y \,\vert\, x)}{\cT_T [e^{f_0}](x)} \, \rho_0(x) \dd x
    = F(1) - F(0).
\]

Let us fix an arbitrary $s \in [0, 1]$. Note that 
\[
    \frac{\dd F(s)}{\dd s}
    = - \left( \; \int\limits_{\R^d} \frac{\sfq_T(y \,\vert\, x) \cT_T \left[ (f_1 - f_0) e^{f_s} \right](x)}{\big(\cT_T[e^{f_s}](x) \big)^2} \, \rho_0(x) \dd x \right) \Bigg\slash
    \left( \; \int\limits_{\R^d} \frac{\sfq_T(y \,\vert\, x)}{\cT_T[e^{f_s}](x)} \, \rho_0(x) \dd x \right).
\]
Then the absolute value of $\dd F(s) / \dd s$ does not exceed
\[
    \left| \frac{\dd F(s)}{\dd s} \right|
    \leq \left( \; \int\limits_{\R^d} \frac{\sfq_T(y \,\vert\, x) \cT_T \left[ |f_1 - f_0| e^{f_s} \right](x)}{\big(\cT_T[e^{f_s}](x) \big)^2} \, \rho_0(x) \dd x \right) \Bigg\slash
    \left( \; \int\limits_{\R^d} \frac{\sfq_T(y \,\vert\, x)}{\cT_T[e^{f_s}](x)} \, \rho_0(x) \dd x \right).
\]
For any $y \in \R^d$, the expression in the right-hand side can be considered as an expectation with respect to the probability density
\[
    \sfp_y(x) \quad \propto \quad \frac{\sfq_T(y \,\vert\, x)}{\cT_T[e^{f_s}](x)} \, \rho_0(x) \dd x,
\]
supported on $\left\{x \in \R^d : \big\| \Sigma^{-1/2} (x - m) \big\| \leq R \right\}$. Hence, it is enough to bound
\[
    \frac{\cT_T \left[ |f_1 - f_0| e^{f_s} \right](x)}{\cT_T[e^{f_s}](x)}
    \quad \text{on} \quad
    \left\{x \in \R^d : \big\| \Sigma^{-1/2} (x - m) \big\| \leq R \right\}.
\]
The rest of the proof proceeds in three steps. First, we apply moment inequalities for sub-exponential random variables to $\cT_T \left[ |f_1 - f_0| e^{f_s} \right](x) / \cT_T[e^{f_s}](x)$ and show that it is proportional to the $\psi_1\big( \sfq_T(\cdot\,\vert\, x) \big)$-norm of $|f_1 - f_0|$. After that, we elaborate on $\big\| |f_1 - f_0| \big\|_{\psi_1(\sfq_T(\cdot\,\vert\, x))}$. Finally, we combine these results to derive the desired bound on $|F(1) - F(0)|$.

\medskip

\noindent
\textbf{Step 1: upper bound on $\cT_T \left[ |f_1 - f_0| e^{f_s} \right](x) / \cT_T[e^{f_s}](x)$.}
\quad
Let us fix an arbitrary $x \in \R^d$ such that $\big\| \Sigma^{-1/2} (x - m) \big\| \leq R$ and define $\eps=\eps(x)$ by
\begin{equation}
    \label{eq:eps_def}
    \frac1\eps = 1 + \log \left( \frac{e^M}{\cT_T [e^{f_s}](x)} \right).
\end{equation}
Since $f_s\le M$, we have $0<\cT_T[e^{f_s}](x)\le e^M$; hence $1/\eps\ge1$ and $\eps\in(0,1]$.
We are going to show that
\[
    \frac{\cT_T \left[ |f_1 - f_0| e^{f_s} \right](x)}{\cT_T[e^{f_s}](x)}
    \leq e \|f_1 - f_0\|_{\psi_1(\sfq_T( \cdot \,\vert\,x))} \big(2 + M \cK(T) + \cA(x, T) \cK(T) \big).
\]
Due to the conditions of the lemma, it holds that
\[
    f_s(y)
    = s f_1(y) + (1 - s) f_0(y)
    \leq s M + (1 - s) M
    = M
    \quad \text{for all $y \in \R^d$ and $s \in [0, 1]$.}
\]
Then, applying the H\"older inequality, we have
\begin{align*}
    \cT_T \big[ |f_1 - f_0| e^{f_s} \big](x)
    &
    \leq e^{\eps M} \cT_T \big[ |f_1 - f_0| e^{(1 - \eps) f_s} \big](x)
    \\&
    \leq e^{\eps M} \left( \cT_T \big[ |f_1 - f_0|^{1/\eps} \big](x) \right)^\eps \left( \cT_T [e^{f_s}](x) \right)^{1 - \eps}
\end{align*}   
for any $x \in \R^d$. This yields that
\[
    \frac{\cT_T \big[ |f_1 - f_0| e^{f_s} \big](x)}{\cT_T [e^{f_s}](x)}
    \leq \left( \frac{e^M}{\cT_T [e^{f_s}](x)} \right)^\eps \left( \cT_T \big[ |f_1 - f_0|^{1/\eps} \big](x) \right)^\eps
    \quad \text{for all $x \in \R^d$.}
\]
According to Lemma E.8 from \cite{puchkin26}, the value of $\cT_T \big[ |f_1 - f_0|^{1/\eps} \big](x)$, which is equal to the expectation of $|f_1 - f_0|^{1/\eps}$ with respect to the density $\sfq_T(\cdot\,\vert\, x)$, satisfies the bound
\[
    \left( \cT_T \big[ |f_1 - f_0|^{1/\eps} \big](x) \right)^\eps
    \leq 2^\eps \left(1 + \frac1\eps \right) \|f_1 - f_0\|_{\psi_1(\sfq_T(\cdot\,\vert\, x))}.
\]
Recalling the definition of $\eps \in (0, 1)$ (see \eqref{eq:eps_def}) that
\begin{align}
    \label{eq:ou_expectation_holder_bound}
    \frac{\cT_T \big[ |f_1 - f_0| e^{f_s} \big](x)}{\cT_T [e^{f_s}](x)}
    &\notag
    \leq \left(1 + \frac1\eps \right) \left( \frac{2 e^M}{\cT_T [e^{f_s}](x)} \right)^\eps \|f_1 -f_0\|_{\psi_1(\sfq_T(\cdot\,\vert\, x))}
    \\&
    \leq e \|f_1 - f_0\|_{\psi_1(\sfq_T(y\,\vert\,x))} \left(2 + \log \left( \frac{e^{M}}{\cT_T [e^{f_s}](x)} \right) \right).
\end{align}
The last line is due to the inequality
\[
    \left( \frac{2 e^M}{\cT_T [e^{f_s}](x)} \right)^\eps
    = \exp\left\{ \eps \log\frac{2 e^M}{\cT_T [e^{f_s}](x)} \right\}
    \leq \exp\left\{ \eps \left(1 + \log\frac{e^M}{\cT_T [e^{f_s}](x)} \right) \right\}
    \leq e.
\]
Let us elaborate on the expression in the right-hand side of \eqref{eq:ou_expectation_holder_bound}. Applying Lemma \ref{lem:log_ou_bound}, we obtain that
\[
    \frac{e^M}{\cT_T [e^{f_s}](x)}
    \leq e^{\cA(x, T) \cK(T)} \left( \frac{e^M}{\cT_\infty [e^{f_s}]} \right)^{\cK(T)},
\]
where the functions $\cA(x, T)$ and $\cK(T)$ are given in \eqref{eq:cA} and \eqref{eq:cK}. Taking into account that $\cT_\infty [e^{f_s}] \geq \exp\{ \cT_\infty [f_s] \} = 1$ and
\[
    \cA(x, T) \leq \cA_{\max}(T)
    \quad \text{on} \quad
    \left\{x \in \R^d : \big\| \Sigma^{-1/2} (x - m) \big\| \leq R \right\},
\]
we deduce that
\[
    \frac{e^M}{\cT_T [e^{f_s}](x)}
    \leq \exp\left\{ \big( M + \cA(x, T) \big) \cK(T) \right\}
    \leq \exp\left\{ \big( M + \cA_{\max}(T) \big) \cK(T) \right\}.
\]
Substituting this bound into \eqref{eq:ou_expectation_holder_bound}, we conclude that
\begin{equation}
    \label{eq:ou_expectation_final_bound}
    \frac{\cT_T \big[ |f_1 - f_0| e^{f_s} \big](x)}{\cT_T [e^{f_s}](x)}
    \leq e \|f_1 - f_0\|_{\psi_1(\sfq_T(\cdot\,\vert\,x))} \big(2 + M \cK(T) + \cA_{\max}(T) \cK(T) \big).
\end{equation}

\noindent
\textbf{Step 2: upper bound on $\big\| |f_1 - f_0| \big\|_{\psi_1(\sfq_T(\cdot\,\vert\,x))}$.}
\quad
Let us remind a reader that $\sfq_T(y \,\vert\, x)$ is the density of the Gaussian distribution $\cN\big( m_T(x), \Sigma_T\big)$, where $m_T(x) = m + e^{-bT}(x - m)$ and $\Sigma_T = \sigma_T^2 \Sigma$. Let us introduce $\xi \sim \cN\big( m_T(x), \Sigma_T \big)$, $\eta \sim \cN(0, I_d)$, and
\[
    u = \frac{\Lambda}{\log 2} \left( 1 + 2 e^{-2bT} R^2 + 2(d + 2) \sigma_T^2 \right) \|\theta_1 - \theta_0\|_\infty
\]
and show that 
\[
    \E \exp\left\{ \frac{|f_1(\xi) - f_0(\xi)|}{u} \right\}
    \leq 2.
\]
Note that, due to Assumption \ref{as:lipschitz_parameterization},
\[
    \big| f_1(y) - f_0(y) \big| \leq
    B \left\|\Sigma^{-1/2}(y - m) \right\|^2 + B,
    \quad \text{where} \quad
    B = \Lambda \|\theta_1 - \theta_0\|_\infty,
\]
for all $y \in \R^d$.
Then it holds that
\begin{align*}
    \E \exp\left\{ \frac{|f_1(\xi) - f_0(\xi)|}{u} \right\}
    &
    \leq \E \exp\left\{ \frac{B \left\|\Sigma^{-1/2}(\xi - m) \right\|^2 + B}{u} \right\}
    \\&
    \leq \E \exp\left\{ \frac{B \left\|\Sigma^{-1/2} \big(\Sigma_T^{1/2} \eta + m_t(x) - m \big) \right\|^2 + B}{u} \right\}
    \\&
    = \E \exp\left\{ \frac{B \left\|\sigma_T \eta + e^{-bT} \Sigma^{-1/2} (x - m) \right\|^2 + B}{u} \right\}.
\end{align*}
Applying the Cauchy-Schwarz inequality, we obtain that
\begin{align*}
    \E \exp\left\{ \frac{|f_1(\xi) - f_0(\xi)|}{u} \right\}
    &
    \leq \E \exp\left\{ \frac{2B \sigma_T^2 \|\eta\|^2 + 2e^{-2bT} B \left\|\Sigma^{-1/2} (x - m)  \right\|^2 + B}{u} \right\}
    \\&
    \leq \exp\left\{ \frac{2 e^{-2bT} B R^2 + B}u \right\}
    \left( 1 - \frac{4B \sigma_T^2}u \right)^{-d/2}.
\end{align*}
Since $-\log(1 - v) \leq v / (1 - v)$ for all $v\in (0,1)$ and $u \geq 2 (d + 2) B \sigma_T^2$, we have
\[
    -\frac{d}2 \log \left( 1 - \frac{4B \sigma_T^2}u \right)
    \leq \frac{d}2 \cdot \frac{4B \sigma_T^2 / u}{1 - 4B \sigma_T^2 / u}
    \leq \frac{d}2 \cdot \frac{4B \sigma_T^2 / u}{1 - 2 / (d + 2)}
    = \frac{2 (d + 2) B \sigma_T^2}u.
\]
This yields that
\begin{align*}
    \E \exp\left\{ \frac{|f_1(\xi) - f_0(\xi)|}{u} \right\}
    &
    \leq \exp\left\{ \frac{2 e^{-2bT} B R^2 + B}u \right\}
    \left( 1 - \frac{4B \sigma_T^2}u \right)^{-d/2}
    \\&
    \leq \exp\left\{ \frac{2 e^{-2bT} B R^2 + B + 2(d + 2) B \sigma_T^2}u \right\}
    = 2.
\end{align*}
The last equality holds due to the definitions of $u$ and $B$. Hence, we proved that
\begin{equation}
    \label{eq:orlicz_norm_bound}
    \big\| |f_1 - f_0| \big\|_{\psi_1(\sfq_T(\cdot\,\vert\,x))}
    \leq u
    \leq 2 \Lambda \left( 1 + 2 e^{-2bT} R^2 + 2(d + 2) \sigma_T^2 \right) \|\theta_1 - \theta_0\|_\infty.
\end{equation}

\noindent
\textbf{Step 3: final bound.}
\quad The inequalities \eqref{eq:ou_expectation_final_bound} and \eqref{eq:orlicz_norm_bound} yield that
\begin{align*}
    \frac{\cT_T \big[ |f_1 - f_0| e^{f_s} \big](x)}{\cT_T [e^{f_s}](x)}
    &
    \leq 2e \Lambda \left( 1 + 2 e^{-2bT} R^2 + 2(d + 2) \sigma_T^2 \right)
    \\&\quad
    \cdot \big(2 + M \cK(T) + \cA_{\max}(T) \cK(T) \big) \|\theta_1 - \theta_0\|_\infty
\end{align*}
for all $x \in \R^d$ such that $\big\| \Sigma^{-1/2} (x - m) \big\| \leq R$. Since
\[
    \supp(\rho_0) \subseteq \left\{x \in \R^d : \big\| \Sigma^{-1/2} (x - m) \big\| \leq R \right\},
\]
we obtain that
\begin{align*}
    \left| \frac{\dd F(s)}{\dd s} \right|
    &
    \leq \left( \; \int\limits_{\R^d} \frac{\sfq_T(y \,\vert\, x) \cT_T \left[ |f_1 - f_0| e^{f_s} \right](x)}{\big(\cT_T[e^{f_s}](x) \big)^2} \, \rho_0(x) \dd x \right) \Bigg\slash
    \left( \; \int\limits_{\R^d} \frac{\sfq_T(y \,\vert\, x)}{\cT_T[e^{f_s}](x)} \, \rho_0(x) \dd x \right)
    \\&
    \leq 2e \Lambda \left( 1 + 2 e^{-2bT} R^2 + 2(d + 2) \sigma_T^2 \right)
    \\&\quad
    \cdot \big(2 + M \cK(T) + \cA_{\max}(T) \cK(T) \big) \|\theta_1 - \theta_0\|_\infty.
\end{align*}
Hence,
\begin{align*}
    &
    \left| \log \int\limits_{\R^d} \frac{\sfq_T(y \,\vert\, x)}{\cT_T[e^{f_1}](x)} \, \rho_0(x) \dd x - \log \int\limits_{\R^d} \frac{\sfq_T(y \,\vert\, x)}{\cT_T [e^{f_0}](x)} \, \rho_0(x) \dd x \right|
    \\&
    = \left| F(1) - F(0) \right|
    \\&
    \leq 2e \Lambda \left( 1 + 2 e^{-2bT} R^2 + 2(d + 2) \sigma_T^2 \right)
    \\&\quad
    \cdot \big(2 + M \cK(T) + \cA_{\max}(T) \cK(T) \big) \|\theta_1 - \theta_0\|_\infty.
\end{align*}

\myendproof

\subsection{Proof of Lemma \ref{lem:log_ou_bound}}
\label{sec:lem_log_ou_bound_proof}

The case $g = 0$ is trivial, so we focus on the situation when $g(x) > 0$ on a set of positive measure. Since the proof is quite cumbersome, we split it into several steps for reader's convenience.

\smallskip

\noindent
\textbf{Step 1: Kolmogorov-Fokker-Planck equation.}
\quad
Let us note that $\cT_t \big[g e^{f} \big](x)$ satisfies the Kolmogorov-Fokker-Planck equation, that is,
\begin{equation}
    \label{eq:kolmogorov-fokker-planck}
    \frac{\partial \cT_t \big[g e^{f} \big](x)}{\partial t}
    = -b (x - m)^\top \nabla \cT_t \big[g e^{f} \big](x) + \frac12 \Tr\left(\Sigma \, \nabla^2 \cT_t \big[g e^{f} \big](x) \right).
\end{equation}
To simplify the expressions in the right-hand side, we represent both
\[
    (x - m)^\top \nabla \cT_t \big[g e^{f} \big](x) \quad \text{and} \quad
    \Tr\left(\Sigma \, \nabla^2 \cT_t \big[g e^{f} \big](x) \right)
\]
as expectations of functions of a Gaussian random vector. Indeed, direct computation yields that
\begin{align*}
    \nabla \cT_t \big[g e^{f} \big](x)
    &
    = \frac{e^{-bt}}{(2\pi)^{d/2} \sqrt{\det(\Sigma_t)}} \int\limits_{\R^d} \Sigma_t^{-1} \big(y - m_t(x) \big) g(y)
    \\& \hspace{1.5in}
    \cdot \exp\left\{ f(y) -\frac12 \left\|\Sigma_t^{-1/2}\big(y - m_t(x) \big) \right\|^2 \right\} \, \dd y
    \\&
    = e^{-bt} \, \E_{\eta \sim \cN(m_t(x), \Sigma_t)} \left[ \Sigma_{t}^{-1} \big(\eta - m_t(x) \big) g(\eta) e^{f(\eta)} \right].
\end{align*}
Similarly, for the Hessian of $\cT_t \big[g e^{f} \big](x)$ it holds that
\begin{align*}
    \nabla^2 \cT_t \big[g e^{f} \big](x)
    = \frac{e^{-2bt}}{(2\pi)^{d/2} \sqrt{\det(\Sigma_t)}}
    &
    \int\limits_{\R^d} \left( \Sigma_t^{-1} \big(y - m_t(x) \big) \big(y - m_t(x) \big)^\top \Sigma_t^{-1} - \Sigma_t^{-1} \right) g(y)
    \\&\quad
    \cdot \exp\left\{ f(y) -\frac12 \left\|\Sigma_t^{-1/2}\big(y - m_t(x) \big) \right\|^2 \right\} \, \dd y.
\end{align*}
Taking into account the relation $\Sigma_t = (1 - e^{-2bt}) \Sigma / (2b)$, it is straightforward to observe that $\Tr\big(\Sigma \, \nabla^2 \cT_t \big[g e^{f} \big](x) \big)$ equals to
\[
    \frac{2b e^{-2bt}}{1 - e^{-2bt}} \, \E_{\eta \sim \cN(m_t(x), \Sigma_t)} \left[ \left( \left\|\Sigma_t^{-1/2} \big(\eta - m_t(x) \big) \right\|^2 - d \right) g(\eta) e^{f(\eta)} \right].
\]
Then the Kolmogorov-Fokker-Planck equation \eqref{eq:kolmogorov-fokker-planck} and the triangle inequality imply that
\begin{align}
    \label{eq:ou_partial_t_bound}
    \left| \frac{\partial \cT_t \big( g(x)e^{f(x)} \big)}{\partial t} \right|
    \leq b e^{-bt} \left| \E_{\eta \sim \cN(m_t(x), \Sigma_t)} \left[ (x - m)^\top \Sigma_{t}^{-1} \big(\eta - m_t(x) \big) g(\eta) e^{f(\eta)} \right] \right|
    &\notag
    \\
    + \frac{b e^{-2bt}}{1 - e^{-2bt}} \left| \E_{\eta \sim \cN(m_t(x), \Sigma_t)} \left[ \left( \left\|\Sigma_t^{-1/2} \big(\eta - m_t(x) \big) \right\|^2 - d \right) g(\eta) e^{f(\eta)} \right] \right|
    &.
\end{align}

\smallskip

\noindent
\textbf{Step 2: H\"older's inequality.}
\quad
Let us fix arbitrary $x \in \R^d$ and $t > 0$ and let $p \geq 2$ be a parameter to be specified later.
Applying H\"older's inequality, we obtain that
\begin{align*}
    &
    \left| \E_{\eta \sim \cN(m_t(x), \Sigma_t)} \left[ (x - m)^\top \Sigma_{t}^{-1} \big(\eta - m_t(x) \big) g(\eta) e^{f(\eta)} \right] \right|
    \\&
    = \left| \E_{\eta \sim \cN(m_t(x), \Sigma_t)} \left[ (x - m)^\top \Sigma_{t}^{-1} \big(\eta - m_t(x) \big) \left(g(\eta) e^{f(\eta)}\right)^{\frac2p} \cdot \left(g(\eta) e^{f(\eta)}\right)^{1 - \frac2p} \right] \right|
    \\&
    \leq \left( \E \left| (x - m)^\top \Sigma_{t}^{-1} \big(\eta - m_t(x) \big) \right|^p \right)^{\frac1p} \left( \E \, g^2(\eta) e^{2f(\eta)} \right)^{\frac1p} \left( \E \, g(\eta) e^{f(\eta)}\right)^{1 - \frac2{p}}.
\end{align*}
The expression in the right-hand side is nothing but
\[
    \left( \E \left| (x - m)^\top \Sigma_{t}^{-1} \big(\eta - m_t(x) \big) \right|^p \right)^{\frac1p} \left( \cT_t \big[ g^2 e^{2f} \big](x) \right)^{\frac1p} \left( \cT_t \big[ g e^{f} \big](x) \right)^{1 - \frac2{p}}.
\]
Thus, we showed that
\begin{align}
    \label{eq:ou_gradient_holder_bound}
    &\notag
    \left| \E_{\eta \sim \cN(m_t(x), \Sigma_t)} \left[ (x - m)^\top \Sigma_{t}^{-1} \big(\eta - m_t(x) \big) g(\eta) e^{f(\eta)} \right] \right|
    \\&\notag
    \leq \left( \E_{\eta \sim \cN(m_t(x), \Sigma_t)} \left| (x - m)^\top \Sigma_{t}^{-1} \big(\eta - m_t(x) \big) \right|^p \right)^{\frac1p}
    \\&\quad
    \cdot \left( \cT_t \big[ g^2 e^{2f} \big](x) \right)^{\frac1p} \left( \cT_t \big[ g e^f \big](x) \right)^{1 - \frac2{p}}
    \\&\notag
    = \left( \E_{\xi \sim \cN(0, I_d)} \left| (x - m)^\top \Sigma_{t}^{-1/2} \xi \right|^p \right)^{\frac1p} \left( \cT_t \big[ g^2 e^{2f} \big](x) \right)^{\frac1p} \left( \cT_t \big[ g e^f \big](x) \right)^{1 - \frac2{p}}.
\end{align}
An upper bound on the absolute value of
\[
    \E_{\eta \sim \cN(m_t(x), \Sigma_t)} \left[ \left( \left\|\Sigma_t^{-1/2} \big(\eta - m_t(x) \big) \right\|^2 - d \right) g(\eta) e^{f(\eta)} \right]
\]
can be derived in a similar way. With the same $p \geq 2$, it holds that
\begin{align}
    \label{eq:ou_hessian_holder_bound}
    &\notag
    \left| \E_{\eta \sim \cN(m_t(x), \Sigma_t)} \left[ \left( \left\|\Sigma_t^{-1/2} \big(\eta - m_t(x) \big) \right\|^2 - d \right) g(\eta) e^{f(\eta)} \right] \right|
    \\&\notag
    \leq \left( \E_{\eta \sim \cN(m_t(x), \Sigma_t)} \left| \left\|\Sigma_t^{-1/2} \big(\eta - m_t(x) \big) \right\|^2 - d \right|^p \right)^{\frac1p}
    \\&\quad
    \cdot \left( \cT_t \big[ g^2 e^{2f} \big](x) \right)^{\frac1p} \left( \cT_t \big[ g e^f \big](x) \right)^{1 - \frac2{p}}
    \\&\notag
    = \left( \E_{\xi \sim \cN(0, I_d)} \left| \|\xi\|^2 - d \right|^p \right)^{\frac1p} \cdot \left( \cT_t \big[ g^2 e^{2f} \big](x) \right)^{\frac1p} \left( \cT_t \big[ g e^f \big](x) \right)^{1 - \frac2{p}}.
\end{align}
Summing up the inequalities \eqref{eq:ou_partial_t_bound}, \eqref{eq:ou_gradient_holder_bound}, and \eqref{eq:ou_hessian_holder_bound}, we conclude that
\begin{align}
    \hspace{-1.1cm}
    \label{eq:ou_partial_t_holder_bound}
    \left| \frac{\partial \cT_t \big( g(x)e^{f(x)} \big)}{\partial t} \right|
    &\notag
    \leq b e^{-bt} \left( \E_{\xi \sim \cN(0, I_d)} \left| (x - m)^\top \Sigma_{t}^{-1/2} \xi \right|^p \right)^{\frac1p}
    \\&\quad\notag
    \cdot \left( \cT_t \big[ g^2 e^{2f} \big](x) \right)^{\frac1p} \left( \cT_t \big[ g e^f \big](x) \right)^{1 - \frac2{p}}
    \\&\quad
    + \frac{b e^{-2bt}}{1 - e^{-2bt}} \left( \E_{\xi \sim \cN(0, I_d)} \left| \|\xi\|^2 - d \right|^p \right)^{\frac1p}
    \\&\quad\notag
    \cdot \left( \cT_t \big[ g^2 e^{2f} \big](x) \right)^{\frac1p} \left( \cT_t \big[ g e^f \big](x) \right)^{1 - \frac2{p}}.
\end{align}

\smallskip

\noindent
\textbf{Step 3: properties of Gaussian random vectors.}
\quad
Let us elaborate on
\[
    \E_{\xi \sim \cN(0, I_d)} \left| \|\xi\|^2 - d \right|^p
    \quad \text{and} \quad
    \E_{\xi \sim \cN(0, I_d)} \left| (x - m)^\top \Sigma_t^{-1/2} \xi \right|^p.
\]
Due to Lemma \ref{lem:chi-squared_moments_bound}, we have
\begin{equation}
    \label{eq:chi-squared_moments_bound}
    \left( \E \left|\|\xi\|^2 - d\right|^p \right)^{1/p}
    \leq 10p \sqrt{d}
    \quad \text{for all $p \geq 1$.}
\end{equation}
An upper bound on $\E_{\xi \sim \cN(0, I_d)} \left| (x - m)^\top \Sigma_t^{-1/2} \xi \right|^p$ follows from the properties of sub-Gaussian random variables. Note that $(x - m)^\top \Sigma_t^{-1/2} \xi \sim \cN\big(0, \|\Sigma_t^{-1/2} (x - m)\|^2 \big)$. Then it holds that
\[
    \E_{\xi \sim \cN(0, I_d)} \left| (x - m)^\top \Sigma_t^{-1/2} \xi \right|^p
    = \left\|\Sigma_t^{-1/2} (x - m) \right\|^p \E_{\z \sim \cN(0, 1)} |\z|^p.
\]
It is known that
\[
    \p_{\z \sim \cN(0, 1)} \big( |\z| \geq u \big) \leq 2 e^{-u^2 / 2}
    \quad \text{for all $u > 0$.}
\]
Then, according to \cite[Proposition 2.5.2]{Vershynin2018}\footnote{In the proof of the implication $1 \Rightarrow 2$ of Proposition 2.5.2, \citeauthor{Vershynin2018} shows that $K_2 = 2 K_1$ (see p. 24).},
\begin{equation}
    \label{eq:gaussian_moments_bound}
    \left( \E_{\z \sim \cN(0, 1)} |\z|^p \right)^{1/p}
    \leq 2 \sqrt{p}.
\end{equation}
The bounds \eqref{eq:ou_partial_t_holder_bound}, \eqref{eq:chi-squared_moments_bound}, and \eqref{eq:gaussian_moments_bound} yield that
\begin{align}
    \label{eq:ou_partial_t_sub-exp_bound}
    \left| \frac{\partial \cT_t \big( g(x)e^{f(x)} \big)}{\partial t} \right|
    &\notag
    \leq \left( 2b e^{-bt} \left\|\Sigma_t^{-1/2} (x - m) \right\| \sqrt{p} + \frac{10b p e^{-2bt} \sqrt{d}}{1 - e^{-2bt}} \right)
    \\&\quad
    \cdot \left( \cT_t \big[ g^2 e^{2f} \big](x) \right)^{\frac1p} \left( \cT_t \big[ g e^f \big](x) \right)^{1 - \frac2{p}}.
\end{align}

\smallskip

\noindent
\textbf{Step 4: upper bound on $\cT_t \big[ g^2 e^{2f} \big](x)$.}
\quad
Our next goal is to show that $\cT_t \big[ g^2 e^{2f} \big](x) \leq G^2(x)$ uniformly over $t > 0$, where $G(x)$ is defined in \eqref{eq:G}. Using the condition \eqref{eq:f_g_conditions}, we observe that
\begin{align*}
    \cT_t \big[ g^2 e^{2f} \big](x)
    &
    \leq e^{2M} \cT_t \big[ g^2 \big](x)
    = e^{2M} \; \E_{\eta \sim \cN(m_t(x), \Sigma_t)} \, g^2(\eta)
    \\&
    \leq e^{2M} \; \E_{\eta \sim \cN(m_t(x), \Sigma_t)} \left( A \left\|\Sigma^{-1/2}(\eta - m) \right\|^\alpha + B \right)^2.
\end{align*}
Taking into account the relations
\[
    m_t(x) - m = e^{-bt}(x - m)
    \quad \text{and} \quad
    \Sigma_t = \frac{1 - e^{-2bt}}{2b} \Sigma,
\]
we obtain that
\begin{align*}
    \cT_t \big[ g^2 e^{2f} \big](x)
    &
    \leq e^{2M} \; \E_{\eta \sim \cN(m_t(x), \Sigma_t)} \left( A \left\|\Sigma^{-1/2}(\eta - m) \right\|^\alpha + B \right)^2
    \\&
    = e^{2M} \; \E_{\xi \sim \cN(0, I_d)} \left( A \left\|\Sigma^{-1/2}\big(m_t(x) - m\big) + \Sigma^{-1/2} \Sigma_t^{1/2} \xi \right\|^\alpha + B \right)^2
    \\&
    = e^{2M} \; \E_{\xi \sim \cN(0, I_d)} \left( A \left\|e^{-bt} \, \Sigma^{-1/2}(x - m) + \sqrt{\frac{1 - e^{-2bt}}{2b}} \, \xi \right\|^\alpha + B \right)^2.
\end{align*}
Due to the triangle inequality, it holds that
\begin{align*}
    \sqrt{\cT_t \big[ g^2 e^{2f} \big](x)}
    &
    \leq e^{M} \sqrt{ \E_{\xi \sim \cN(0, I_d)} \left( A \left\|e^{-bt} \, \Sigma^{-1/2}(x - m) + \sqrt{\frac{1 - e^{-2bt}}{2b}} \, \xi \right\|^\alpha + B \right)^2}
    \\&
    \leq B e^{M} + A e^M \sqrt{ \E_{\xi \sim \cN(0, I_d)} \left\|e^{-bt} \, \Sigma^{-1/2}(x - m) + \sqrt{\frac{1 - e^{-2bt}}{2b}} \, \xi \right\|^{2\alpha}}.
\end{align*}
The inequality $(u + v)^{\alpha} \leq 2^{\alpha - 1} u^\alpha + 2^{\alpha - 1} v^\alpha$ holding for all non-negative $u$ and $v$ and all $\alpha\geq 1$ ensures that
\begin{align*}
    &
    \sqrt{ \E_{\xi \sim \cN(0, I_d)} \left\|e^{-bt} \, \Sigma^{-1/2}(x - m) + \sqrt{\frac{1 - e^{-2bt}}{2b}} \, \xi \right\|^{2\alpha}}
    \\&
    = \left( \left[\E_{\xi \sim \cN(0, I_d)} \left\|e^{-bt} \, \Sigma^{-1/2}(x - m) + \sqrt{\frac{1 - e^{-2bt}}{2b}} \, \xi \right\|^{2\alpha} \right]^{\frac1{2\alpha}} \right)^\alpha
    \\&
    \leq \left( e^{-bt} \left\| \Sigma^{-1/2}(x - m) \right\| + \sqrt{\frac{1 - e^{-2bt}}{2b}} \; \left[ \E_{\xi \sim \cN(0, I_d)} \, \|\xi\|^{2\alpha} \right]^{\frac1{2\alpha}} \right)^\alpha
    \\&
    \leq 2^{\alpha - 1} e^{-\alpha bt} \left\| \Sigma^{-1/2}(x - m) \right\|^{\alpha} + 2^{\alpha - 1} \left( \frac{1 - e^{-2bt}}{2b} \right)^{\alpha/2} \sqrt{\E_{\xi \sim \cN(0, I_d)} \, \|\xi\|^{2\alpha}}.
\end{align*}
Hence, we showed that
\begin{align*}
    \sqrt{\cT_t \big[ g^2 e^{2f} \big](x)}
    &
    \leq B e^M + 2^{\alpha - 1} A e^{M - \alpha bt} \left\| \Sigma^{-1/2}(x - m) \right\|^{\alpha}
    \\&\quad
    + 2^{\alpha - 1} A e^M \left( \frac{1 - e^{-2bt}}{2b} \right)^{\alpha/2} \sqrt{\E_{\xi \sim \cN(0, I_d)} \|\xi\|^{2\alpha}}
    \\&
    \leq B e^M + 2^{\alpha - 1} A e^M \left( \left\| \Sigma^{-1/2}(x - m) \right\|^{\alpha} + (2b)^{-\alpha/2} \sqrt{\E_{\xi \sim \cN(0, I_d)} \, \|\xi\|^{2\alpha}} \right).
\end{align*}
Finally, applying the inequality $(u + v)^{\alpha} \leq 2^{\alpha - 1} u^\alpha + 2^{\alpha - 1} v^\alpha$ once again and using Lemma \ref{lem:chi-squared_moments_bound}, we obtain that
\begin{align*}
    \E_{\xi \sim \cN(0, I_d)} \, \|\xi\|^{2\alpha}
    &
    \leq \E_{\xi \sim \cN(0, I_d)} \left( \left| \|\xi\|^2 - d\right| + d \right)^\alpha
    \\&
    \leq 2^{\alpha - 1} \E_{\xi \sim \cN(0, I_d)} \left| \|\xi\|^2 - d\right|^\alpha + 2^{\alpha - 1} d^\alpha
    \\&
    \leq 2^{\alpha - 1} (10 \alpha \sqrt{d})^\alpha + 2^{\alpha - 1} d^\alpha.
\end{align*}
This yields the bound
\begin{align}
    \label{eq:ou_partial_t_cauchy-schwarz_bound}
    \sqrt{\cT_t \big[ g^2 e^{2f} \big](x)}
    &\notag
    \leq B e^M + 2^{\alpha - 1} A e^M \left\| \Sigma^{-1/2}(x - m) \right\|^{\alpha}
    \\&\quad
    + 4^{\alpha - 1} A e^M (2b)^{-\alpha/2} \left((10 \alpha \sqrt{d})^\alpha + d^\alpha \right)
    \\&\notag
    = G(x),
\end{align}
which holds uniformly over $t > 0$.

\smallskip

\noindent
\textbf{Step 5: choice of $p$.}
\quad
The bounds \eqref{eq:ou_partial_t_sub-exp_bound}, \eqref{eq:ou_partial_t_cauchy-schwarz_bound} and the equality $\Sigma_t = (1 - e^{-2bt}) \Sigma / (2b)$ imply that
\begin{align*}
    &
    \left| \frac{\partial \cT_t \big[ g e^{f} \big](x) / \partial t}{\cT_t \big[ g e^{f} \big](x)} \right|
    \\&
    \leq \left( 2b e^{-bt} \left\|\Sigma_t^{-1/2} (x - m) \right\| \sqrt{p} + \frac{10b p e^{-2bt} \sqrt{d}}{1 - e^{-2bt}} \right)
    \cdot \left( \cT_t \big[ g^2 e^{2f} \big](x) \right)^{\frac1p} \left( \cT_t \big[ g e^{f} \big](x) \right)^{-\frac2{p}}
    \\&
    \leq \left(\frac{(2b)^{3/2} e^{-bt} \sqrt{p}}{\sqrt{1 - e^{-2bt}}} \left\|\Sigma^{-1/2} (x - m) \right\| + \frac{10 p b e^{-2bt} \sqrt{d}}{1 - e^{-2bt}} \right) \left( \frac{G(x)}{\cT_t \big[ g e^{f} \big](x)} \right)^{\frac2{p}}.
\end{align*}
In the last line we used $\Sigma_t = (1 - e^{-2bt}) \Sigma / (2b)$. Applying Young's inequality
\[
    (2b)^{3/2} \sqrt{p} \left\|\Sigma^{-1/2} (x - m) \right\|
    \leq \frac12 \left(
    \frac{2b^2 \left\|\Sigma^{-1/2} (x - m) \right\|^2}{ \sqrt{d}} + 4bp \sqrt{d} \right),
\]
we obtain that
\begin{align*}
    \left| \frac{\partial \cT_t \big[ g e^{f} \big](x) / \partial t}{\cT_t \big[ g e^{f} \big](x)} \right|
    &
    \leq \frac{b^{2} e^{-bt}}{\sqrt{1 - e^{-2bt}} \cdot \sqrt{d}} \left\|\Sigma^{-1/2} (x - m) \right\|^2 \left( \frac{G(x)}{\cT_t \big[ g e^{f} \big](x)} \right)^{\frac2{p}}
    \\&\quad
    + 2bp \sqrt{d} \left( \frac{e^{-bt}}{\sqrt{1 - e^{-2bt}}} + \frac{5 e^{-2bt}}{1 - e^{-2bt}}\right) \left( \frac{G(x)}{\cT_t \big[ g e^{f} \big](x)} \right)^{\frac2{p}}.
\end{align*}
Let us choose $p = 2 \vee \log\left(G(x) / \cT_t \big[ g e^{f} \big](x) \right)$. Then it is easy to observe that
\[
    \left( \frac{G(x)}{\cT_t[g e^f](x)} \right)^{\frac2{p}}
    \leq e^2,
\]
and, therefore,
\begin{align*}
    \left| \frac{\partial \cT_t \big[ g e^{f} \big](x) / \partial t}{\cT_t \big[ g e^{f} \big](x)} \right|
    &
    \leq \frac{b^{2} e^{2-bt}}{\sqrt{1 - e^{-2bt}} \cdot \sqrt{d}} \left\|\Sigma^{-1/2} (x - m) \right\|^2
    \\&\quad
    + 2be^2 \sqrt{d} \left( \frac{e^{-bt}}{\sqrt{1 - e^{-2bt}}} + \frac{5 e^{-2bt}}{1 - e^{-2bt}}\right) \left( 2 \vee \log\frac{G(x)}{\cT_t \big[ g e^{f} \big](x)} \right).
\end{align*}
We would like to recall that (see \eqref{eq:ou_partial_t_cauchy-schwarz_bound})
\[
    \sqrt{\cT_t \big[ g^2 e^{2f} \big](x)} \leq G(x).
\]
This means that
\begin{equation}
    \label{eq:log_is_non-negative}
    \log \frac{G(x)}{\cT_t \big[ g e^{f} \big](x)}
    \geq \log \frac{G(x)}{\sqrt{\cT_t \big[ g^2 e^{2f} \big](x)}} \geq 0,
\end{equation}
and, as a consequence, we have
\[
    2 \vee \log\frac{G(x)}{\cT_t \big[ g e^{f} \big](x)}
    \leq 2 + \log\frac{G(x)}{\cT_t \big[ g e^{f} \big](x)}.
\]
Thus, we obtain that
\begin{align}
    \label{eq:ou_partial_t_ode_bound_1}
    \left| \frac{\partial \cT_t \big[ g e^{f} \big](x) / \partial t}{\cT_t \big[ g e^{f} \big](x)} \right|
    &\notag
    \leq \frac{b^{2} e^{2-bt}}{\sqrt{1 - e^{-2bt}} \cdot \sqrt{d}} \left\|\Sigma^{-1/2} (x - m) \right\|^2
    \\&\quad
    + 2be^2 \sqrt{d} \left( \frac{e^{-bt}}{\sqrt{1 - e^{-2bt}}} + \frac{5 e^{-2bt}}{1 - e^{-2bt}}\right) \left( 2 + \log\frac{G(x)}{\cT_t \big[ g e^{f} \big](x)} \right).
\end{align}

\noindent
\textbf{Step 6: properties of ODEs.}
\quad Let us note that, due to \eqref{eq:ou_partial_t_ode_bound_1}, we have 
\begin{align}
    \label{eq:ou_partial_t_ode_bound_2}
    \left|\frac{\partial}{\partial t} \log\frac{G(x)}{\cT_t \big[ g e^{f} \big](x)} \right|
    &\notag
    = \left| \frac{\partial \cT_t \big[ g e^{f} \big](x) / \partial t}{\cT_t \big[ g e^{f} \big](x)} \right|
    \\&
    \leq \frac{b^{2} e^{2-bt}}{\sqrt{1 - e^{-2bt}} \cdot \sqrt{d}} \left\|\Sigma^{-1/2} (x - m) \right\|^2
    \\&\quad\notag
    + 2be^2 \sqrt{d} \left( \frac{e^{-bt}}{\sqrt{1 - e^{-2bt}}} + \frac{5 e^{-2bt}}{1 - e^{-2bt}}\right) \left( 2 + \log\frac{G(x)}{\cT_t \big[ g e^{f} \big](x)} \right).
\end{align}
In other words the partial derivative of $\log( G(x) / \cT_t \big[ g e^{f} \big](x)$ is bounded by its value. At this step, we use properties of ordinary differential equations to convert the inequality \eqref{eq:ou_partial_t_ode_bound_2} into an upper bound on the absolute value of
\[
    \log\frac{G(x)}{\cT_t \big[ g e^{f} \big](x)} - \log\frac{G(x)}{\cT_\infty \big[ g e^{f} \big]}.
\]
For this purpose, let us fix an arbitrary $x \in \R^d$ and apply Lemma \ref{lem:ode_bound} with
\[
    a(t) = \frac{b^{2} e^{2-bt}}{\sqrt{1 - e^{-2bt}} \cdot \sqrt{d}} \left\|\Sigma^{-1/2} (x - m) \right\|^2 + 4be^2 \sqrt{d} \left( \frac{e^{-bt}}{\sqrt{1 - e^{-2bt}}} + \frac{5 e^{-2bt}}{1 - e^{-2bt}}\right)
\]
and
\[
    \varkappa(t) = 2be^2 \sqrt{d} \left( \frac{e^{-bt}}{\sqrt{1 - e^{-2bt}}} + \frac{5 e^{-2bt}}{1 - e^{-2bt}}\right).
\]
Note that the function
\[
    \log\frac{G(x)}{\cT_t \big[ g e^{f} \big](x)}
\]
is always non-negative due to \eqref{eq:log_is_non-negative}.
Since
\[
    \int\limits_t^{+\infty} \frac{b e^{-bs} \, \dd s}{\sqrt{1 - e^{-2bs}}} = \arcsin\big(e^{-bt}\big)
    \quad \text{and} \quad
    \int\limits_t^{+\infty} \frac{2b e^{-2bs} \, \dd s}{1 - e^{-2bs}}
    = -\log\big(1 - e^{-2bt} \big),
\]
it holds that
\[
    \int\limits_t^{+\infty} \varkappa(\tau) \, \dd \tau
    = 2e^2 \sqrt{d} \arcsin(e^{-bt}) - 5e^2 \sqrt{d} \log\big(1 - e^{-2bt} \big)
    = \log \cK(t)
\]
and
\begin{align*}
    \int\limits_t^{+\infty} a(s) \, \dd s
    &
    = \left( \frac{b e^2}{\sqrt{d}} \left\|\Sigma^{-1/2} (x - m) \right\|^2
    + 4e^2 \sqrt{d}\right) \arcsin(e^{-bt})
    - 10e^2 \sqrt{d} \log\big(1 - e^{-2bt} \big)
    \\&
    = \frac{b e^2}{\sqrt{d}} \left\|\Sigma^{-1/2} (x - m) \right\|^2 \arcsin(e^{-bt})
    + 2 \log \cK(t)
    = \cA(x, t),
\end{align*}
where the functions $\cK(t)$ and $\cA(x, t)$ are defined in \eqref{eq:cK} and \eqref{eq:cA}, respectively. Then, according to Lemma \ref{lem:ode_bound}, it holds that
\[
    \frac1{\cK(t)} \log\frac{G(x)}{\cT_\infty \big[ g e^{f} \big]} - \cA(x, t)
    \leq \log\frac{G(x)}{\cT_t \big[ g e^{f} \big](x)}
    \leq \cK(t) \left( \log\frac{G(x)}{\cT_\infty \big[ g e^{f} \big]} + \cA(x, t) \right),
\]
and we finally obtain that
\[
    e^{-\cA(x, t) \cK(t)} \left( \frac{\cT_\infty \big[ g e^{f} \big]}{G(x)} \right)^{\cK(t)}
    \leq \frac{\cT_t \big[ g e^{f} \big](x)}{G(x)}
    \leq  e^{\cA(x, t)} \left( \frac{\cT_\infty \big[ g e^{f} \big]}{G(x)} \right)^{1 / \cK(t)}.
\]
\myendproof

\subsection{Proof of Lemma \ref{lem:sub-gaussian_density}}
\label{sec:lem_sub-gaussian_density_proof}

Invoking Lemma \ref{lem:ou_lower_bound} and recalling that $\varphi(y) \leq M$ for all $y \in \R^d$, we obtain that
\begin{align*}
    \rho_T^\varphi(y)
    &
    = \int\limits_{\R^d} \frac{e^{\varphi(y)} \, \sfq_T(y \,\vert\, x)}{\cT_T[e^\varphi](x)} \, \rho_0(x) \, \dd x
    \\&
    \leq (1 + 2L \sigma_T^{2})^{d/2} \, \exp\left\{ 2M + \frac{L \sigma_T^{-2} R^2 e^{-2bT}}{2L + \sigma_T^{-2}} \right\} \int\limits_{\R^d} \sfq_T(y \,\vert\, x) \, \rho_0(x) \, \dd x
    \\&
    = (1 + 2L \sigma_T^{2})^{d/2} \, \exp\left\{ 2M + \frac{L \sigma_T^{-2} R^2 e^{-2bT}}{2L + \sigma_T^{-2}} \right\} \rho_T^0(y)
    \\&
    \leq (1 + 2L \sigma_T^{2})^{d/2} \, \exp\left\{ 2M + \frac{R^2 e^{-2bT}}{2\sigma_T^2} \right\} \rho_T^0(y).
\end{align*}

\subsection{Proof of Lemma \ref{lem:ou_lower_bound}}
\label{sec:lem_ou_lower_bound_proof}

In view of Assumption \ref{as:log-potential_quadratic_growth}, we have
\begin{align*}
    \cT_T[e^\varphi](x)
    &
    = \int\limits_{\R^d} e^{\varphi(y)} \, \sfq_T(y \,|\, x) \, \dd y
    \\&
    \geq \int\limits_{\R^d} \exp\left\{ -L \left\|\Sigma^{-1/2} (y - m) \right\|^2 - M \right\} \, \sfq_T(y \,|\, x) \, \dd y.
\end{align*}
Taking into account the definition of the transition density $\sfq_T(y \,|\, x)$ (see \eqref{eq:ou_transition_density}), we obtain that
\begin{align}
    \label{eq:ou_initial_lower_bound}
    \cT_T[e^\varphi](x)
    &\notag
    \geq \frac{(2\pi)^{-d/2} e^{-M}}{\det(\Sigma_T)^{1/2}} \int\limits_{\R^d} \exp\left\{ -L \left\|\Sigma^{-1/2} (y - m) \right\|^2 \right\} 
    \\&\qquad\qquad\qquad\qquad\notag
    \cdot \exp\left\{-\frac12 \left\|\Sigma_T^{-1/2} \big(y - m_T(x) \big) \right\|^2 \right\} \, \dd y
    \\&
    = \frac{(2\pi \sigma_T^2)^{-d/2} e^{-M}}{\det(\Sigma)^{1/2}} \int\limits_{\R^d} \exp\left\{ -L \left\|\Sigma^{-1/2} (y - m) \right\|^2 \right\}
    \\&\qquad\qquad\qquad\qquad\notag
    \cdot \exp\left\{-\frac1{2\sigma_T^2} \left\|\Sigma^{-1/2} \big(y - m_T(x) \big) \right\|^2 \right\} \, \dd y.
\end{align}
The function $m_T(x)$ and $\sigma_T$ are defined in \eqref{eq:m_t_sigma_t}. We are going to elaborate on the expression in the right-hand side. For this purpose, we introduce
\[
    f(y) = L \left\|\Sigma^{-1/2} (y - m) \right\|^2 + \frac1{2\sigma_T^2} \left\|\Sigma^{-1/2} \big(y - m_T(x) \big) \right\|^2.
\]
It is straightforward to observe that $f(y)$ attains its minimum at
\[
    y^* = \frac{2Lm + \sigma_T^{-2} m_T(x)}{2L + \sigma_T^{-2}}
\]
and that $\nabla^2 f(y) = (2L + \sigma_T^{-2}) \Sigma^{-1}$. Using Taylor's expansion with the Lagrange remainder, we obtain that
\[
    f(y) - f(y^*) = \frac12 (2L + \sigma_T^{-2}) \left\| \Sigma^{-1/2} (y - y^*) \right\|^2.
\]
Recalling the definition of $m_T(x)$ (see \eqref{eq:m_t_sigma_t}), we note that
\[
    f(y^*) = \frac{L \sigma_T^{-2} e^{-2bT}}{2L + \sigma_T^{-2}} \left\| \Sigma^{-1/2} (x - m) \right\|^2.
\]
Hence, we showed that
\[
    f(y) = \frac12 (2L + \sigma_T^{-2}) \left\| \Sigma^{-1/2} (y - y^*) \right\|^2 + \frac{L \sigma_T^{-2} e^{-2bT}}{2L + \sigma_T^{-2}} \left\| \Sigma^{-1/2} (x - m) \right\|^2.
\]
Substituting this equality into \eqref{eq:ou_initial_lower_bound}, we obtain that
\begin{align*}
    \cT_T[e^\varphi](x)
    &
    \geq \frac{(2\pi \sigma_T^2)^{-d/2} e^{-M}}{\det(\Sigma)^{1/2}} \, \exp\left\{ -\frac{L \sigma_T^{-2} e^{-2bT}}{2L + \sigma_T^{-2}} \left\| \Sigma^{-1/2} (x - m) \right\|^2 \right\}
    \\&\quad
    \cdot \int\limits_{\R^d} \exp\left\{ -\frac12 (2L + \sigma_T^{-2}) \left\| \Sigma^{-1/2} (y - y^*) \right\|^2 \right\} \, \dd y
    \\&
    = e^{-M} (1 + 2L \sigma_T^{2})^{-d/2} \, \exp\left\{ -\frac{L \sigma_T^{-2} e^{-2bT}}{2L + \sigma_T^{-2}} \left\| \Sigma^{-1/2} (x - m) \right\|^2 \right\}.
\end{align*}
Since $\| \Sigma^{-1/2} (x - m) \| \leq R$ for all $x \in \supp(\rho_0)$, we obtain the desired bound.

\myendproof

\section{Properties of subquadratic log-densities}
\label{sec:log-density_properties}

In this section, we present some properties of probability densities $\sfp(x)$ such that $\log \sfp(x) = \cO(\|x\|^2)$. Such densities naturally arise in our setup in view of the following observation (see also Assumption \ref{as:log-potential_quadratic_growth} and Lemma \ref{lem:log-density_quadratic_growth}).
\begin{Lem}
    \label{lem:log_density_ratio_upper_bound}
    Grant Assumptions \ref{as:base-process}, \ref{as:bounded_support}, \ref{as:sub_gaussian_density}, and \ref{as:log-potential_quadratic_growth}. Then there are
    \begin{equation}
        \label{eq:a}
        \ttA = 2 \left( L + \sigma_T^{-2} \right) \left\| \Sigma^{-1} \right\|
    \end{equation}
    and
    \begin{align}
        \label{eq:b}
        \ttB
        &
        = \log \rho_{\max} + 2 \left( L + \sigma_T^{-2} \right) \left\|\Sigma^{-1/2} m \right\|^2 + 2M
        \\&\quad
        + \frac{d}2 \log(2\pi \sigma_T^2) + \frac{\log \det(\Sigma)}2 + \frac{R^2 e^{-2bT}}{\sigma_T^2}
    \end{align}
    such that
    \begin{equation}
        \label{eq:log-density_ratio_upper_bound}
        \log \frac{\rho_T^*(y)}{\rho_T^\varphi(y)}
        \leq \ttA \|y\|^2 + \ttB
        \quad \text{for all $y \in \R^d$ and all $\varphi \in \Phi$.}
    \end{equation}
\end{Lem}

We start with the following preliminary bound, which helps us to show that the class of log-potentials $\Phi$ satisfies a Bernstein-type condition. This is one of key moments in the proof of our main result, allowing us to derive rates of convergence possibly faster than $\cO(n^{-1/2})$.

\begin{Lem}
    \label{lem:squared_log_expectation_bound}
    For any two probability densities $\sfp$ and $\sfq$ on $\R^d$ such that
    \[
        \int\limits_{\R^d} \log^2 \left( \frac{\sfq(x)}{\sfp(x)} \right) \sfp(x) \, \dd x < + \infty
    \]
    and any $\omega \in (0, 1)$, it holds that
    \begin{align}
        \label{eq:squared_log_expectation_bound}
        \int\limits_{\R^d} \log^2 \left( \frac{\sfp(x)}{\sfq(x)} \right) \sfp(x) \, \dd x
        &\notag
        \leq 2 \log(1 / \omega) \, \KL(\sfp, \sfq)
        \\&\quad
        + 2 \int\limits_{\R^d} \log^2 \left( \frac{(1 - \omega) \sfq(x) + \omega \sfp(x)}{\sfq(x)} \right) \sfp(x) \, \dd x.
    \end{align}
\end{Lem}

The proof of Lemma \ref{lem:squared_log_expectation_bound} is postponed to Appendix \ref{sec:lem_squared_log_expectation_bound_proof}. The next lemma ensures that, under Assumption \ref{as:sub_gaussian_density}, the second term in the right-hand side of \eqref{eq:squared_log_expectation_bound} grows polynomially with $\omega$.

\begin{Lem}
    \label{lem:squared_log_mixture_bound}
    Let $\sfp$ and $\sfq$ be arbitrary probability densities on $\R^d$.
    Assume that $\sfp$ is a probability density of a centered sub-Gaussian distribution on $\R^d$ with a variance proxy $\ttv^2$, that is,
    \[
        \E_{\xi \sim \sfp} \; e^{u^\top \xi}
        \leq e^{\ttv^2 \|u\|^2 / 2}
        \quad \text{for all $u \in \R^d$.}
    \]
    Let $\sfp \ll \sfq$ and suppose that there are
    constants $A > 0$ and $B \in \R$ such that
    \[
        \log\frac{\sfp(x)}{\sfq(x)} \leq A \|x\|^2 + B
        \qquad \text{for all $x$ from the support of $\sfp(x)$.}
    \]
    Then, for any $\omega \in (0, 1/2]$, it holds that
    \[
        \int\limits_{\R^d} \log^2 \left( \frac{(1 - \omega) \sfq(x) + \omega \sfp(x)}{\sfq(x)} \right) \sfp(x) \, \dd x
        \leq 4 \omega^2 + 2e (1 + \alpha^2) e^{B / \alpha} \omega^{1 / \alpha},
    \]
    where $\alpha = \big( (d + 2) A \ttv^2 \big) \vee 1$.
\end{Lem}

The proof of Lemma \ref{lem:squared_log_mixture_bound} is moved to Appendix \ref{sec:lem_squared_log_mixture_bound_proof}. Lemma \ref{lem:squared_log_expectation_bound} and Lemma \ref{lem:squared_log_mixture_bound} imply that, if one takes $\omega = e^{-B} (1 + \alpha^2)^{-\alpha} (4n)^{-\alpha} \land (2 \sqrt{n})^{-1}$, then
\[
    \int\limits_{\R^d} \log^2 \left( \frac{\sfp(x)}{\sfq(x)} \right) \sfp(x) \, \dd x
    \lesssim \log(en) \, \KL(\sfp, \sfq) + \cO(1/n).
\]
In other words, under the conditions of these lemmata, the variance of $\log\big( \sfp(\xi) / \sfq(\xi) \big)$, where $\xi \sim \sfp$, is controlled by its expectation. Another important result is the following upper bound on the $\psi_1$-norm of $\log\big( \sfp(\xi) / \sfq(\xi) \big)$, $\xi \sim \sfp$. This upper bound on the Orlicz norm is necessary for understanding behaviour of distribution tails.

\begin{Lem}
    \label{lem:log_density_ratio_orlicz_norm_bound}
    Let $Y \sim \rho_T^*$. Under Assumptions \ref{as:base-process}, \ref{as:bounded_support}, \ref{as:sub_gaussian_density}, and \ref{as:log-potential_quadratic_growth}, it holds that
    \begin{align*}
        \left\| \sup\limits_{\varphi \in \Phi} \left| \log \frac{\rho_T^*(Y)}{\rho_T^\varphi(Y)}  \right| \right\|_{\psi_1}
        &
        \leq 2 \vee \left( 4 \max\{\ttB, 0\} + 4(d + 1) \ttA \ttv^2 \right)
        \\&\quad
        \vee \left( \frac{4M + d \log\left(1 + 2 L \sigma_T^2 \right)}{\log 2} + \frac{R^2 e^{-2bT}}{\sigma_T^2 \log 2} \right),
    \end{align*}
    where the constants $\ttA$ and $\ttB$ are given by \eqref{eq:a} and \eqref{eq:b}.
\end{Lem}

The proof of Lemma \ref{lem:log_density_ratio_orlicz_norm_bound} is deferred to Appendix \ref{sec:lem_log_density_ratio_orlicz_norm_bound_proof}.

\subsection{Proof of Lemma \ref{lem:log_density_ratio_upper_bound}}
\label{sec:lem_log_density_ratio_upper_bound_proof}

Let us fix a log-potential $\varphi \in \Phi$ and consider the corresponding marginal density $\rho_T^\varphi$. Since, according to Assumption \ref{as:log-potential_quadratic_growth}, $\varphi(x)$ does not exceed $M$ for all $x \in \R^d$, we can apply Lemma \ref{lem:log-density_quadratic_growth} claiming that
\[
    \varphi(y) - \log\rho_T^\varphi(y)
    \leq \sigma_T^{-2} \left\| \Sigma^{-1/2} \big(y - m \big) \right\|^2 + M + \frac{d}2 \log(2\pi \sigma_T^2) + \frac{\log \det(\Sigma)}2 + \frac{R^2 e^{-2bT}}{\sigma_T^2}.
\]
This and the upper bound $\rho_T^*(y) \leq \rho_{\max}$ yield that
\begin{align*}
    \log \frac{\rho_T^*(y)}{\rho_T^\varphi(y)}
    &
    \leq \log \rho_{\max} - \varphi(y) + \sigma_T^{-2} \left\| \Sigma^{-1/2} \big(y - m \big) \right\|^2 
    \\&\quad
    + M + \frac{d}2 \log(2\pi \sigma_T^2) + \frac{\log \det(\Sigma)}2 + \frac{R^2 e^{-2bT}}{\sigma_T^2}.
\end{align*}
Since $\varphi(y) \geq -L \|\Sigma^{-1/2} (y - m)\|^2 - M$ due to Assumption \ref{as:log-potential_quadratic_growth}, we obtain that
\begin{align*}
    \log \frac{\rho_T^*(y)}{\rho_T^\varphi(y)}
    &
    \leq \left( L + \sigma_T^{-2} \right) \|\Sigma^{-1/2} (y - m)\|^2 + \log \rho_{\max} + 2M
    \\&\quad
    + \frac{d}2 \log(2\pi \sigma_T^2) + \frac{\log \det(\Sigma)}2 + \frac{R^2 e^{-2bT}}{\sigma_T^2}.
\end{align*}
Finally, the inequality
\[
    \left\| \Sigma^{-1/2} (y - m) \right\|^2
    \leq 2 \left\|\Sigma^{-1/2} y \right\|^2 + 2 \left\|\Sigma^{-1/2} m \right\|^2
    \leq 2 \left\|\Sigma^{-1} \right\| \, \|y\|^2 + 2 \left\|\Sigma^{-1 / 2} m \right\|^2
\]
concludes the proof.

\myendproof

\subsection{Proof of Lemma \ref{lem:squared_log_expectation_bound}}
\label{sec:lem_squared_log_expectation_bound_proof}

According to the Newton-Leibniz formula, for any $u < 1$ we have
\[
    \log^2(1 - u)
    = \left( \int\limits_0^1 \frac{u \, \dd s}{1 - su} \right)^2.
\]
Using Young's inequality, we obtain that
\begin{align}
    \label{eq:squared_log_young_bound}
    \log^2(1 - u)
    &\notag
    \leq 2 \left( \int\limits_0^{1 - \omega} \frac{u \, \dd s}{1 - su} \right)^2 + 2 \left( \; \int\limits_{1 - \omega}^1 \frac{u \, \dd s}{1 - su} \right)^2
    \\&
    = 2 \left( \int\limits_0^{1 - \omega} \frac{u \, \dd s}{1 - su} \right)^2 + 2 \left( \log(1 - u) - \log\big(1 - (1 - \omega) u\big) \right)^2
    \\&\notag
    = 2 \left( \; \int\limits_0^{1 - \omega} \frac{u \, \dd s}{1 - su} \right)^2 + 2 \log^2 \left(1 + \frac{\omega u}{1 - u} \right).
\end{align}
The first term in the right-hand side does not exceed
\begin{align}
    \label{eq:squared_integral_cauchy-schwarz_bound}
    2 \left( \int\limits_0^{1 - \omega} \frac{1}{\sqrt{1 - s}} \cdot \frac{u \sqrt{1 - s}}{1 - su} \, \dd s \right)^2
    &\notag
    \leq 2 \int\limits_0^{1 - \omega} \frac{\dd s}{1 - s} \int\limits_0^{1 - \omega} \frac{u^2 (1 - s) \, \dd s}{(1 - su)^2}
    \\&
    = 2 \log(1/\omega) \, \int\limits_0^{1 - \omega} \frac{u^2 (1 - s) \, \dd s}{(1 - su)^2}
    \\&\notag
    \leq 2 \log(1/\omega) \, \int\limits_0^1 \frac{u^2 (1 - s) \, \dd s}{(1 - su)^2}.
\end{align}
On the other hand, due to Taylor's expansion with an integral remainder term, it holds that
\[
    \log(1 - u) = -u - \int\limits_0^1 \frac{u^2 (1 - s) \, \dd s}{(1 - su)^2}.
\]
In other words, the right-hand side in \eqref{eq:squared_integral_cauchy-schwarz_bound} is equal to
\[
    2 \log(1/\omega) \big( -u - \log(1 - u) \big).
\]
This, together with \eqref{eq:squared_log_young_bound}, yields that
\[
    \log^2(1 - u)
    \leq 2 \log(1/\omega) \big( -u - \log(1 - u) \big) + 2 \log^2 \left(1 + \frac{\omega u}{1 - u} \right).
\]
Substituting $u$ in the expression above with $(\sfp(x) - \sfq(x)) / \sfp(x)$, we observe that
\begin{align}
    \label{eq:squared_log_pointwise_bound}
    \log^2\left( \frac{\sfq(x)}{\sfp(x)} \right)
    &\notag
    \leq 2 \log(1/\omega) \left( \frac{\sfq(x) - \sfp(x)}{\sfp(x)} - \log\frac{\sfq(x)}{\sfp(x)} \right)
    \\&\quad
    + 2 \log^2 \left( \frac{(1 - \omega) \sfq(x) + \omega \sfp(x)}{\sfq(x)} \right)
\end{align}
for all $x$ such that $\sfp(x) > 0$. Let us note that the condition
\[
    \int\limits_{\R^d} \log^2 \left( \frac{\sfq(x)}{\sfp(x)} \right) \sfp(x) \, \dd x < + \infty
\]
implies that $\sfp \ll \sfq$. This means that $\sfq(x) > 0$ $\sfp$-almost everywhere.
Integrating \eqref{eq:squared_log_pointwise_bound}, we finally obtain that
\begin{align*}
    \int\limits_{\R^d} \log^2 \left( \frac{\sfp(x)}{\sfq(x)} \right) \sfp(x) \, \dd x
    &
    \leq 2 \log(1 / \omega) \, \KL(\sfp, \sfq)
    \\&\quad
    + 2 \int\limits_{\R^d} \log^2 \left( \frac{(1 - \omega) \sfq(x) + \omega \sfp(x)}{\sfq(x)} \right) \sfp(x) \, \dd x.
\end{align*}

\myendproof

\subsection{Proof of Lemma \ref{lem:squared_log_mixture_bound}}
\label{sec:lem_squared_log_mixture_bound_proof}

Due to the conditions of the lemma, for any $x$ belonging to the support of $\sfp$, it holds that 
\begin{align*}
    \log(1 - \omega)
    &
    \leq \log \left( \frac{(1 - \omega) \sfq(x) + \omega \sfp(x)}{\sfq(x)} \right)
    \\&
    \leq \log \left( 1 + \omega e^{\log(\sfp(x) / \sfq(x))} \right)
    \\&
    \leq \log \left( 1 + \omega e^{A \|x\|^2 + B} \right).
\end{align*}
Note that the expression in the left-hand side is at least $-2\omega$, since $\omega \in (0, 1/2]$. This yields that
\begin{align*}
    \int\limits_{\R^d} \log^2 \left( \frac{(1 - \omega) \sfq(x) + \omega \sfp(x)}{\sfq(x)} \right) \sfp(x) \, \dd x
    &
    \leq \int\limits_{\R^d} \max\left\{ 4\omega^2, \log^2 \left( 1 + \omega e^{A \|x\|^2 + B} \right) \right\} \sfp(x) \, \dd x
    \\&
    \leq \int\limits_{\R^d} \left(4\omega^2 + \log^2 \left( 1 + \omega e^{A \|x\|^2 + B} \right) \right) \sfp(x) \, \dd x
    \\&
    = 4 \omega^2 + \E_{X \sim \sfp} \log^2 \left( 1 + \omega e^{A \|X\|^2 + B} \right).
\end{align*}
Let us elaborate on the second term in the right-hand side. Introducing $\eps = e^B \omega$ for brevity, we observe that
\begin{align*}
    \E_{X \sim \sfp} \log^2 \left( 1 + \omega e^{A \|X\|^2 + B} \right)
    &
    = \E_{X \sim \sfp} \log^2 \left( 1 + \eps e^{A \|X\|^2} \right)
    \\&
    = \int\limits_0^{+\infty} \p_{X \sim \sfp} \left( \log^2 \left( 1 + \eps e^{A \|X\|^2} \right) > r \right) \, \dd r.
\end{align*}
Substituting $r$ with $s^2$, $s > 0$, we obtain that
\begin{align*}
    \int\limits_0^{+\infty} \p_{X \sim \sfp} \left( \log^2 \left( 1 + \eps e^{A \|X\|^2} \right) > r \right) \, \dd r
    &
    = 2 \int\limits_0^{+\infty} \p_{X \sim \sfp} \left( \log \left( 1 + \eps e^{A \|X\|^2} \right) > s \right) \, s \, \dd s
    \\&
    = 2 \int\limits_0^{+\infty} \p_{X \sim \sfp} \left( e^{A \|X\|^2} > \frac{e^s - 1}\eps \right) \, s \, \dd s.
\end{align*}
Let us take $\alpha = \big( (d + 2) A \ttv^2 \big) \vee 1$ and apply the Markov inequality:
\begin{align*}
    \E_{X \sim \sfp} \log^2 \left( 1 + \omega e^{A \|X\|^2 + B} \right)
    &
    = 2 \int\limits_0^{+\infty} \p_{X \sim \sfp} \left( e^{A \|X\|^2} > \frac{e^s - 1}\eps \right) \, s \, \dd s
    \\&
    \leq 2 \eps^{1/\alpha} \; \E \exp\left\{ \frac{\|X\|^2}{ (d + 2) \ttv^2} \right\} \int\limits_0^{+\infty} \frac{s \, \dd s}{(e^s - 1)^{1 / \alpha}}.
\end{align*}
We have to elaborate on the expression in the right-hand side. First, we focus our attention on the exponential moment of $\|X\|^2 / \big( (d + 2) \ttv^2 \big)$.
Let $\gamma \sim \cN(0, I_d)$ be a Gaussian random vector, which is independent of $X \sim \sfp$. Recalling that $X$ is a centered sub-Gaussian random vector, we deduce that
\[
    \E \exp\left\{ \frac{\|X\|^2}{ (d + 2) \ttv^2} \right\}
    = \E \exp\left\{ \frac{X^\top \gamma}{\ttv} \sqrt{\frac2{d + 2}} \right\}
    \leq \E e^{\|\gamma\|^2 / (d + 2)}
    = \left(1 - \frac{2}{d + 2} \right)^{-d/2}
    \leq e.
\]
The last inequality follows from the bound $-\log(1 - u) \leq u / (1 - u)$ holding for all $u \in (0, 1)$.
This yields that
\begin{equation}
    \label{eq:squared_log_expectation_markov_bound}
    \E_{X \sim \sfp} \log^2 \left( 1 + \omega e^{A \|X\|^2 + B} \right)
    \leq 2e \, \eps^{1/\alpha} \int\limits_0^{+\infty} \frac{s \, \dd s}{(e^s - 1)^{1 / \alpha}}.
\end{equation}
Finally, let us consider the integral
\[
    \int\limits_0^{+\infty} \frac{s \, \dd s}{(e^s - 1)^{1 / \alpha}}
    = \int\limits_0^{+\infty} \frac{s e^{-s / \alpha}\, \dd s}{(1 - e^{-s})^{1 / \alpha}}.
\]
Since $u^{1 / \alpha} \leq 1 + (u - 1) / \alpha$ for all $u \geq 0$ and $\alpha \geq 1$, it holds that
\[
	(1 - e^{-s})^{-1 / \alpha}
	\leq 1 + \frac{(1 - e^{-s})^{-1} - 1}{\alpha}
	= 1 + \frac1{\alpha (e^s - 1)}
\]
for any $s > 0$. Then 
\begin{align*}
    \int\limits_0^{+\infty} \frac{s e^{-s / \alpha}\, \dd s}{(1 - e^{-s})^{1 / \alpha}}
    &
    \leq \int\limits_0^{+\infty} s e^{-s / \alpha}\, \dd s
    + \frac1{\alpha} \int\limits_0^{+\infty} \frac{s e^{-s / \alpha}\, \dd s}{e^s - 1)}
    \\&
    \leq \int\limits_0^{+\infty} s e^{-s / \alpha}\, \dd s
    + \frac1{\alpha} \int\limits_0^{+\infty} e^{-s / \alpha}\, \dd s
    \\&
    = 1 + \alpha^2.
\end{align*}
Substituting this bound into \eqref{eq:squared_log_expectation_markov_bound}, we obtain that
\[
    \E_{X \sim \sfp} \log^2 \left( 1 + \omega e^{A \|X\|^2 + B} \right)
    \leq 2e(1 + \alpha^2) \eps^{1/\alpha}
    = 2e (1 + \alpha^2) e^{B / \alpha} \omega^{1 / \alpha},
\]
and then
\begin{align*}
    \int\limits_{\R^d} \log^2 \left( \frac{(1 - \omega) \sfq(x) + \omega \sfp(x)}{\sfq(x)} \right) \sfp(x) \, \dd x
    &
    \leq 4 \omega^2 + \E_{X \sim \sfp} \log^2 \left( 1 + \omega e^{A \|X\|^2 + B} \right)
    \\&
    \leq 4 \omega^2 + 2e (1 + \alpha^2) e^{B / \alpha} \omega^{1 / \alpha}.
\end{align*}
\myendproof

\subsection{Proof of Lemma \ref{lem:log_density_ratio_orlicz_norm_bound}}
\label{sec:lem_log_density_ratio_orlicz_norm_bound_proof}

Let us introduce
\[
    u = 1 \vee \log_2 C_T \vee \left( 2 \max\{\ttB, 0\} + 2(d + 1) \ttA \ttv^2 \right),
\]
where
\[
    C_T = \left(1 + 2 L \sigma_T^2 \right)^{d / 2}
    \exp\left\{ 2M + \frac{R^2 e^{-2bT}}{2 \sigma_T^2} \right\}
\]
and the constants $\ttA$ and $\ttB$ are defined in \eqref{eq:a} and \eqref{eq:b}.
We are going to show that
\[
    \E \sup\limits_{\varphi \in \Phi} \exp\left\{ \frac1{2u} \left| \log \frac{\rho_T^*(Y)}{\rho_T^\varphi(Y)} \right| \right\} \leq 2.
\]
We start with the following observation:
\begin{align*}
    &
    \E \sup\limits_{\varphi \in \Phi} \exp\left\{ \frac1{2u} \left| \log \frac{\rho_T^*(Y)}{\rho_T^\varphi(Y)} \right| \right\}
    \leq \left( \E \sup\limits_{\varphi \in \Phi} \exp\left\{ \frac1{u} \left| \log \frac{\rho_T^*(Y)}{\rho_T^\varphi(Y)}  \right| \right\} \right)^{1/2}
    \\&
    \leq \left( \E \sup\limits_{\varphi \in \Phi} \exp\left\{ \frac1{u} \log \frac{\rho_T^\varphi(Y)}{\rho_T^*(Y)} \right\} + \E \sup\limits_{\varphi \in \Phi} \exp\left\{ \frac1{u} \log \frac{\rho_T^*(Y)}{\rho_T^\varphi(Y)} \right\} \right)^{1/2}.
\end{align*}
Let us elaborate on the first term in the right-hand side. Since, due to the conditions of the lemma, $\cT_\infty[\varphi] = 0$ for all $\varphi \in \Phi$, Lemma \ref{lem:sub-gaussian_density} implies that
\[
    \frac{\rho_T^\varphi(y)}{\rho_T^*(y)}
    \leq \frac{C_T \rho_T^0(y)}{\rho_T^*(y)}
    \quad \text{for all $\varphi \in \Phi$ and all $y \in \R^d$.}
\]
In view of the Jensen's inequality and the condition $u \geq 1$, we obtain that
\[
    \E \sup\limits_{\varphi \in \Phi} \exp\left\{ \frac1{u} \log \frac{\rho_T^\varphi(Y)}{\rho_T^*(Y)} \right\}
    \leq C_T^{1/u} \; \E \exp\left\{ \frac1{u} \log \frac{\rho_T^0(Y)}{\rho_T^*(Y)} \right\}
    \leq C_T^{1/u} \; \left( \E_{Y \sim \rho_T^*} \frac{\rho_T^0(Y)}{\rho_T^*(Y)} \right)^{1/u}.
\]
Taking into account that $u \geq \log_2 C_T$ by the definition, we conclude that
\[
    \E \sup\limits_{\varphi \in \Phi} \exp\left\{ \frac1{u} \log \frac{\rho_T^\varphi(Y)}{\rho_T^*(Y)} \right\}
    \leq C_T^{1/u}
    \leq 2.
\]

On the other hand, Lemma \ref{lem:log_density_ratio_upper_bound} 
yields that
\[
    \log \frac{\rho_T^*(y)}{\rho_T^\varphi(y)}
    \leq \ttA \|y\|^2 + \ttB
    \quad \text{for all $y \in \R^d$ and all $\varphi \in \Phi$,}
\]
and then
\[
    \E \sup\limits_{\varphi \in \Phi} \exp\left\{ \frac1{u} \log \frac{\rho_T^*(Y)}{\rho_T^\varphi(Y)} \right\}
    \leq e^{\ttB / u} \; \E e^{\ttA \|Y\|^2 / u}.
\]
Let $\gamma \sim \cN(0, I_d)$ be a Gaussian random vector in $\R^d$ which is independent of $Y$. Then we can represent the exponential moment of $\ttA \|Y\|^2 / u$ in the following form:
\[
    \E e^{\ttA \|Y\|^2 / u}
    = \E_{Y \sim \rho^*} \E_{\gamma \sim \cN(0, I_d)} \exp\left\{ \sqrt{\frac{2\ttA}u} \, \gamma^\top Y \right\}.
\]
According to the conditions of the lemma, $\xi$ is a sub-Gaussian random vector with variance proxy $\ttv^2$. This yields that
\begin{align*}
    \E e^{\ttA \|Y\|^2 / u}
    &
    = \E_{Y \sim \rho^*} \E_{\gamma \sim \cN(0, I_d)} \exp\left\{ \sqrt{\frac{2\ttA}u} \, \gamma^\top Y \right\}
    \\&
    \leq \E e^{\ttA \ttv^2 \|\gamma\|^2 / u}
    = \left(1 - \frac{2\ttA\ttv^2}u \right)^{-d/2}.
\end{align*}
Since $u \geq 2 (\ttB \vee 0) + 2(d + 1) \ttA \ttv^2 \geq 2(d + 1) \ttA \ttv^2$, we obtain that
\[
    -\frac{d}2 \log\left(1 - \frac{2\ttA\ttv^2}u \right)
    \leq \frac{d}{2} \cdot \frac{2\ttA\ttv^2 / u}{1 - 2 / (2d + 2)}
    = \frac{(d + 1) \ttA \ttv^2}u.
\]
Hence, it holds that
\begin{align*}
    &
    \E \sup\limits_{\varphi \in \Phi} \exp\left\{ \frac1{u} \log \frac{\rho_T^*(Y)}{\rho_T^\varphi(Y)} \right\}
    \leq e^{\ttB / u} \; \E e^{\ttA \|Y\|^2 / u}
    \leq \exp\left\{ \frac{\ttB + (d + 1) \ttA \ttv^2}{u} \right\}.
\end{align*}
Due to the definition, $u$ is not less than $2\ttB + 2(d + 1) \ttA \ttv^2$. This yields that the expression in the right-hand side not exceed $e^{1/2}$. This implies that
\begin{align*}
    &
    \E \sup\limits_{\varphi \in \Phi} \exp\left\{ \frac1{2u} \left| \log \frac{\rho_T^*(Y)}{\rho_T^\varphi(Y)} \right| \right\}
    \\&
    \leq \left( \E \sup\limits_{\varphi \in \Phi} \exp\left\{ \frac1{u} \log \frac{\rho_T^\varphi(Y)}{\rho_T^*(Y)} \right\} + \E \sup\limits_{\varphi \in \Phi} \exp\left\{ \frac1{u} \log \frac{\rho_T^*(Y)}{\rho_T^\varphi(Y)} \right\} \right)^{1/2}
    \\&
    \leq \sqrt{2 + e^{1/2}}
    < 2.
\end{align*}
The proof is finished.

\myendproof

\section{Auxiliary results on existence and approximation of Schrödinger log-potentials}
\label{sec:approximation_error}

This section collects auxiliary result used in the proof of Theorem \ref{th:approximation_error_relu}. First of all, we have to ensure that the right Schrödinger log-potential exists and well defined under Assumptions \ref{as:base-process}, \ref{as:bounded_support}, and \ref{as:log-density_envelope}.

\begin{Lem}
    \label{lem:quadratic_envelope_phi_star}
    Let Assumptions \ref{as:base-process}, \ref{as:bounded_support}, and \ref{as:log-density_envelope} hold. Then the right Schrödinger log-potential $\varphi^*$ is well-defined and satisfies
    \begin{equation}
        \label{eq:quadratic_envelope_phi_star}
        -a_- \left\|\Sigma^{-1/2} (y - m) \right\|^2 - M_* \leq \varphi^*(y) \leq M_*
        \quad \text{for all $y \in \R^d$,}
    \end{equation}
    for some non-negative real number $M_*$.
    In addition, the target density $\rho_T^*$ admits a representation
    \begin{equation}
        \label{eq:rho_T_star}
        \rho_T^*(y) = e^{\varphi^*(y)} \int\limits_{\R^d} \frac{\sfq_T(y\,|\,x)}{\cT_T[e^{\varphi^*}](x)} \, \rho_0(x) \, \dd x.
    \end{equation}
\end{Lem}

The proof of Lemma \ref{lem:quadratic_envelope_phi_star} is deferred to Appendix \ref{sec:lem_quadratic_envelope_phi_star_proof}. It is based on the following auxiliary result.

\begin{Lem}
    \label{lem:nu_0_finite}
    Under the conditions of Lemma \ref{lem:quadratic_envelope_phi_star}, the measure $\cV_0$ from \eqref{eq:right_marginal_simplified} is finite, that is, there exists $\cM_0 \in (0, +\infty)$, such that
    \[
        0 < \int\limits_{\|\Sigma^{-1/2} (x - m)\| \leq R} \dd \cV_0(x)
        \leq \cM_0.
    \]
\end{Lem}

We provide the proof of Lemma \ref{lem:nu_0_finite} in Appendix \ref{sec:lem_nu_0_finite_proof}. Our next finding connects the KL-divergence between the terminal marginals with the $L_2$-distance between the corresponding log-potentials.

\begin{Lem}
\label{lem:quadratic_kl_bound}
Given right Schr\"odinger log-potentials $\psi$ and $\varphi$, such that $\cT_T[e^{\psi}](x)$ and $\cT_T[e^{\varphi}](x)$ take their values in the interval $(0, +\infty)$ for all $x \in \R^d$,
let $\pi^\psi$ and $\pi^\varphi$ stand for the joint densities of $(X_0^\psi, X_T^\psi)$ and $(X_0^\varphi, X_T^\varphi)$, respectively, that is,
\[
    \pi^\psi(x, y)
    =
    \frac{\rho_0(x) \, \sfq_T(y\mid x) \, e^{\psi(y)}}{\cT_T[e^{\psi}](x)},
    \qquad
    \pi^\varphi(x,y)
    =
    \frac{\rho_0(x) \, \sfq_T(y\mid x) \,e^{\varphi(y)}}{\cT_T[e^{\varphi}](x)}.
\]
Denote the terminal marginals of $\pi^\psi$ and $\pi^\varphi$ by $\rho_T^\psi$ and $\rho_T^\varphi$, respectively. Then it holds that
\[
     \KL(\pi^\psi, \pi^\varphi)
     \le
     \|\varphi - \psi\|_{L_2(\rho_T^\psi)}^2
     +
     \|\varphi - \psi\|_{L_2(\rho_T^\varphi)}^2.
\]
\end{Lem}

We carry out the proof of Lemma \ref{lem:quadratic_kl_bound} in Appendix \ref{sec:lem_quadratic_kl_bound_proof}. Thanks to this lemma, we reduce the study of $\smash{\inf\limits_{\varphi \in \Phi}} \KL(\rho_T^*, \rho_T^\varphi)$ to approximation of the corresponding Schrödinger log-potential:
\[
     \inf\limits_{\varphi \in \Phi} \KL(\rho_T^*, \rho_T^\varphi)
     \leq
     \inf\limits_{\varphi \in \Phi} \KL(\pi^*, \pi^\varphi)
     \le
     \inf\limits_{\varphi \in \Phi} \left\{ \|\varphi - \varphi^* \|_{L_2(\rho_T^*)}^2
     +
     \|\varphi - \varphi^*\|_{L_2(\rho_T^\varphi)}^2 \right\},
\]
where
\[
    \pi^*(x, y) = \frac{\rho_0(x) \, \sfq_T(y\mid x) \, e^{\varphi^*(y)}}{\cT_T[e^{\varphi^*}](x)}.
\]
The next two results are devoted to properties of the target log-potential $\varphi^*$. The regularity assumptions ensure that $\varphi^*$ is sufficiently smooth. To be more precise, the following statement holds.

\begin{Lem}
\label{lem:potential_smoothness}
Grant Assumptions~\ref{as:base-process}, \ref{as:bounded_support}, \ref{as:log-density_envelope}, and \ref{as:smoothess_class}. Then there exists $C > 0$ such that
\[
    A_r(\varphi^*)
    =
    \max_{|\alpha|\le s} \, \max_{\|x\| \leq r} \left\{ r^{|\alpha|} |\partial^\alpha\varphi^*(x)| \right\}
    \lesssim r^{\beta_\rho\vee s\vee2},
    \quad \text{for all $r \ge 1$.}
\]
\end{Lem}

Lemma \ref{lem:potential_smoothness} allows us to apply standard tools for approximation of smooth functions via feedforward neural networks.

\begin{Lem}
\label{prop:schmidt_hieber_compact_approx}
Let $s,d\in\mathbb N$. For $r\ge1$ and $f\in C^s_{\mathrm{loc}}(\R^d)$, define
\[
    A_r(f) = \max_{|\alpha|\le s} \, \max_{\|x\| \leq r} \left\{r^{|\alpha|}\|\partial^\alpha f\| \right\}.
\]
Then, for any resolution level $N \geq 2$, one can choose $K_N \lesssim \log N$, $S_N \lesssim N \log N$, and a suitable architecture $W_N$ with $\|W_N\|_\infty \lesssim N \log N$, such that, for every $f\in C^s_{\mathrm{loc}}(\R^d)$, every $r \ge 1$, and every $A \ge A_r(f) \vee 1$, there exists $g_N \in \NN(K_N + 1, W_N, S_N, 3/2 \vee A)$ satisfying
\[
    \sup_{\|x\| \leq r} |g_N(x) - f(x)|
    \lesssim A N^{-s/d}.
\]
The hidden constants behind $\lesssim$ depend only on $s$ and $d$.
\end{Lem}

The proofs of Lemmata \ref{lem:potential_smoothness} and \ref{prop:schmidt_hieber_compact_approx} are deferred to Appendices \ref{sec:lem_potential_smoothness_proof} and \ref{sec:prop_schmidt_hieber_compact_approx_proof}, respectively.

\subsection{Proof of Lemma \ref{lem:quadratic_envelope_phi_star}}
\label{sec:lem_quadratic_envelope_phi_star_proof}

We split the proof into several steps for convenience. 

\medskip

\noindent
\textbf{Step 1: well-posedness of $\varphi^*$.}
\quad
Let us first show that the right Schr\"odinger log-potential $\varphi^*(y) = \log \nu_T(y)$ is well-defined and finite for almost all $y \in \R^d$. 
According to Theorem 2.2 from \cite{Daipra1991}, there exists unique (up to normalizing constants) pair of sigma-finite measures $(\cV_0, \cV_T)$ such that the marginals of
\[
    \dd \Pi(x, y) = \sfq_T(y \,|\, x) \, \dd \cV_0(x) \, \dd \cV_T(y)
\]
have the densities $\rho_0$ and $\rho_T^*$, in particular,
\begin{equation}
    \label{eq:right_marginal}
    \rho_T^*(y) \, \dd y = \dd \cV_T(y) \, \int\limits_{\R^d} \sfq_T(y\,|\,x) \, \dd \cV_0(x)
    \quad \text{almost everywhere on $\R^d$.}
\end{equation}
Moreover, in view of \citet[eq. (3.13)]{Daipra1991}, $\cV_T$ admits a density $\nu_T(y)$. Furthermore, since $\rho_0$ is supported on $\big\{x \in \R^d : \big\|\Sigma^{-1/2} (x - m) \big\| \leq R \big\}$ and the endpoint marginal is absolutely continuous, we can conclude that
\[
    \supp(\cV_0) \subseteq \supp(\rho_0)
    \subseteq \left\{x \in \R^d : \big\|\Sigma^{-1/2} (x - m) \big\| \leq R \right\}.
\]
Thus, the identity \eqref{eq:right_marginal} simplifies to
\begin{equation}
    \label{eq:right_marginal_simplified}
    \rho_T^*(y) = \nu_T(y) \int\limits_{\|\Sigma^{-1/2} (x - m)\| \leq R} \sfq_T(y\,|\,x) \, \dd \cV_0(x).
\end{equation}
Let us note that the measure $\cV_0$ is finite due to Lemma \ref{lem:nu_0_finite}, that is, there exists $\cM_0 \in (0, +\infty)$ such that the total mass
\[
    \m = \int\limits_{\|\Sigma^{-1/2} (x - m)\| \leq R} \dd \cV_0(x)
\]
lies in the interval $(0, \cM_0)$.
Since the function $\sfq_T(y\,|\,x)$ is positive and continuous on the compact set $\{x \in \R^d : \|\Sigma^{-1/2} (x - m)\| \leq R \}$, we obtain that
\[
    0 < \int\limits_{\|\Sigma^{-1/2} (x - m)\| \leq R} \sfq_T(y\,|\,x) \, \dd \cV_0(x) < +\infty
    \quad \text{for almost all $y \in \R^d$.}
\]
Taking into account that $\rho_T^*(y) > 0$ for any $y \in \R^d$ 
due to the conditions of the Lemma, we deduce that
\[
    0 < \nu_T(y) \leq \rho_T^*(y) \Big\slash \int\limits_{\|\Sigma^{-1/2} (x - m)\| \leq R} \sfq_T(y\,|\,x) \, \dd \cV_0(x) < +\infty.
\]
For this reason, $\varphi^*(y) = \log \nu_T(y)$ is well-defined and finite for almost all $y \in \R^d$.

\medskip

\noindent
\textbf{Step 2: upper and lower bounds on $\varphi^*$.}
\quad
We proceed with the derivation of the inequality \eqref{eq:quadratic_envelope_phi_star}. Using the lower bound \eqref{eq:q_T_lower_bound} from the proof of Lemma \ref{lem:nu_0_finite}, we observe that
\[
    \rho_T^*(y) \geq \frac{\m \, \det(\Sigma)^{-1/2}}{\left(\sqrt{2\pi} \sigma_T \right)^d} \exp\left\{ \varphi^*(y) - a_+ \left\|\Sigma^{-1/2} \big(y - m \big) \right\|^2 - \frac{a_+ R^2 e^{-2bT}}{2 \sigma_T^2 a_+ - 1} \right\}.
\]
This is equivalent to
\begin{align*}
    \varphi^*(y)
    &
    \leq \log \rho_T^*(y) + d \log\left(\sqrt{2\pi} \sigma_T \right) + \frac12\log\det(\Sigma) - \log \m
    \\&\quad
    + a_+ \left\|\Sigma^{-1/2} \big(y - m \big) \right\|^2 + \frac{a_+ R^2 e^{-2bT}}{2 \sigma_T^2 a_+ - 1}
    \\&
    \leq b_+ + d \log\left(\sqrt{2\pi} \sigma_T \right) + \frac12\log\det(\Sigma) - \log \m + \frac{a_+ R^2 e^{-2bT}}{2 \sigma_T^2 a_+ - 1},
\end{align*}
where the last inequality follows from Assumption \ref{as:log-density_envelope}.
On the other hand, since
\[
    \sfq_T(y\,|\,x)
    \leq \left(\sqrt{2\pi} \sigma_T \right)^{-d} \det(\Sigma)^{-1/2},
\]
it holds that
\begin{align*}
    \varphi^*(y)
    &
    \geq \log \rho_T^*(y) + d \log\left(\sqrt{2\pi} \sigma_T \right) + \frac12\log\det(\Sigma) - \log \m
    \\&
    \geq -a_- \left\|\Sigma^{-1/2} (y - m) \right\|^2 - b_- + d \log\left(\sqrt{2\pi} \sigma_T \right) + \frac12\log\det(\Sigma) - \log \m.
\end{align*}
Hence, the two-sided bound \eqref{eq:quadratic_envelope_phi_star} holds with
\begin{align*}
    M_*
    &
    = 0 \vee \left( b_- - d \log\left(\sqrt{2\pi} \sigma_T \right) - \frac12\log\det(\Sigma) + \log \m \right)
    \\&\quad
    \vee \left( b_+ + d \log\left(\sqrt{2\pi} \sigma_T \right) + \frac12\log\det(\Sigma) - \log \m + \frac{a_+ R^2 e^{-2bT}}{2 \sigma_T^2 a_+ - 1} \right).
\end{align*}

\noindent
\textbf{Step 3: expression for $\rho_T^*$.}
\quad
Finally, we establish \eqref{eq:rho_T_star}. Let us recall that the measure
\[
    \dd \Pi(x, y) = \sfq_T(y \,|\, x) \, \dd \cV_0(x) \, \dd \cV_T(y)
\]
has the marginals $\rho_0(x)\,\dd x$ and $\rho_T^*(y) \, \dd y$. Thus, almost everywhere on $\R^d$, it holds that
\[
    \rho_0(x)\,\dd x
    = \left( \; \int\limits_{\R^d} \sfq_T(y \,|\, x) \, \dd \cV_T(y) \right) \dd \cV_0(x)
    = \left( \; \int\limits_{\R^d} \sfq_T(y \,|\, x) \, e^{\varphi^*(y)} \dd y \right) \dd \cV_0(x).
\]
Since $\varphi^*$ satisfies \eqref{eq:quadratic_envelope_phi_star} and $\sfq_T(y\,\vert\,x) > 0$ for all $x, y \in \R^d$, we have
\[
    \cT_T[e^{\varphi^*}](x) = \int\limits_{\R^d} \sfq_T(y \,|\, x) \, e^{\varphi^*(y)} \dd y > 0.
\]
This yields that the measure $\cV_0$ is absolutely continuous and
\[
    \frac{\dd \cV_0(x)}{\dd x} = \frac{\rho_0(x)}{\cT_T[e^{\varphi^*}](x)}.
\]
Substituting this expression into \eqref{eq:right_marginal_simplified}, we obtain \eqref{eq:rho_T_star}.

\myendproof

\subsection{Proof of Lemma \ref{lem:nu_0_finite}}
\label{sec:lem_nu_0_finite_proof}

The inequality
\[
    \int\limits_{\|\Sigma^{-1/2} (x - m)\| \leq R} \dd \cV_0(x) > 0
\]
easily follows from \eqref{eq:right_marginal_simplified} and the fact that $\rho_T^*(y)$ is a probability density. If we assume that
\[
    \int\limits_{\|\Sigma^{-1/2} (x - m)\| \leq R} \dd \cV_0(x) = 0,
\]
then
\[
    \int\limits_{\R^d} \nu_T(y) \, \dd y \int\limits_{\|\Sigma^{-1/2} (x - m)\| \leq R} \sfq_T(y\,|\,x) \dd \cV_0(x) = 0 \neq 1 = \int\limits_{\R^d} \rho_T^*(y) \, \dd y,
\]
which contradicts \eqref{eq:right_marginal_simplified}.

We move to the proof of the upper bound. It is easy to check that there exists $y_0 \in \R^d$ such that $\nu_T(y_0) > 0$. Indeed, if $\nu_T(y)$ was equal to zero for almost all $y \in \R^d$, we would again get a contradiction with \eqref{eq:right_marginal_simplified}. Then
\[
   \int\limits_{\|\Sigma^{-1/2} (x - m)\| \leq R} \sfq_T(y_0 \,|\, x) \, \dd \cV_0(x) = \frac{\rho_T^*(y_0)}{\nu_T(y_0)} < +\infty.
\]
Let us elaborate on the conditional density $\sfq_T(y\,|\,x)$. 
Introducing
\[
    \eps = 2 \sigma_T^2 \, a_+ - 1 > 0    
\]
and applying the Cauchy-Schwarz inequality, we note that
\begin{align*}
    \left\|\Sigma^{-1/2} \big(y - m_T(x) \big) \right\|^2
    &
    \leq (1 + \eps) \left\|\Sigma^{-1/2} \big(y - m \big) \right\|^2 + e^{-2bT} \left(1 + \frac1\eps \right) \left\|\Sigma^{-1/2} \big(x - m \big) \right\|^2
    \\&
    \leq (1 + \eps) \left\|\Sigma^{-1/2} \big(y - m \big) \right\|^2 + e^{-2bT} R^2 \left(1 + \frac1\eps \right)
    \\&
    = 2 \sigma_T^2 a_+ \left\|\Sigma^{-1/2} \big(y - m \big) \right\|^2 + \frac{2 \sigma_T^2 a_+ e^{-2bT} R^2}{\eps}
\end{align*}
for all $y \in \R^d$ and $x \in \supp(\cV_0)$.
This yields that
\begin{equation}
    \label{eq:q_T_lower_bound}
    \sfq_T(y\,|\,x)
    \geq \frac{\det(\Sigma)^{-1/2}}{\left(\sqrt{2\pi} \sigma_T \right)^d} \exp\left\{ -a_+ \left\|\Sigma^{-1/2} \big(y - m \big) \right\|^2 - \frac{a_+ e^{-2bT} R^2}{\eps} \right\},
\end{equation}
and then
\begin{align*}
    \frac{\rho_T^*(y_0)}{\nu_T(y_0)}
    &
    \geq \frac{\det(\Sigma)^{-1/2}}{\left(\sqrt{2\pi} \sigma_T \right)^d} \exp\left\{ -a_+ \left\|\Sigma^{-1/2} \big(y_0 - m \big) \right\|^2 - \frac{a_+ e^{-2bT} R^2}{\eps} \right\}
    \\&\quad
    \cdot \int\limits_{\|\Sigma^{-1/2} (x - m)\| \leq R} \dd \cV_0(x).
\end{align*}
This is equivalent to
\[
    \int\limits_{\|\Sigma^{-1/2} (x - m)\| \leq R} \dd \cV_0(x)
    \leq \cM_0,
\]
where
\[
    \cM_0 = \frac{\rho_T^*(y_0)}{\nu_T(y_0)} \cdot \frac{\left(\sqrt{2\pi} \sigma_T \right)^d}{\det(\Sigma)^{-1/2}} \, \exp\left\{ a_+ \left\|\Sigma^{-1/2} \big(y_0 - m \big) \right\|^2 + \frac{a_+ e^{-2bT} R^2}{\eps} \right\} < +\infty.
\]
\myendproof

\subsection{Proof of Lemma \ref{lem:quadratic_kl_bound}}
\label{sec:lem_quadratic_kl_bound_proof}

Let $\delta(y) = \psi(y) - \varphi(y)$ and
\[
    c(x) = \log\frac{\cT_T[e^\psi](x)}{\cT_T[e^\varphi](x)}.
\]
We use the elementary bound
\begin{align*}
    \KL(\pi^\psi, \pi^\varphi)
    &
    \leq \KL(\pi^\psi, \pi^\varphi) + \KL(\pi^\varphi, \pi^\psi)
    = \J(\pi^\psi, \pi^\varphi)
    \\&
    \leq
    \frac12 \int\limits_{\R^d} \int\limits_{\R^d}  \log^2\left(\frac{\pi^\psi(x, y)}{\pi^\varphi(x, y)} \right) \left( \pi^\psi(x, y) + \pi^\varphi(x, y) \right) \, \dd x \, \dd y,
\end{align*}
which follows from $u e^u - u \leq (u^2 + u^2 e^u) / 2$. 
With the introduced notations, we can rewrite the expression in the right-hand side as follows:
\[
    \KL(\pi^\psi, \pi^\varphi)
    \leq \frac12 \int\limits_{\R^d} \int\limits_{\R^d}  \big( \delta(y) - c(x) \big)^2 \left( \pi^\psi(x, y) + \pi^\varphi(x, y) \right) \, \dd x \, \dd y.
\]
Introducing the conditional densities
\[
    \pi^\psi(y\mid x)
    =
    \frac{\sfq_T(y\mid x) \, e^{\psi(y)}}{\cT_T[e^{\psi}](x)},
    \quad \text{and} \quad
    \pi^\varphi(y\mid x)
    =
    \frac{\sfq_T(y\mid x) \,e^{\varphi(y)}}{\cT_T[e^{\varphi}](x)},
\]
we obtain that
\[
    \KL(\pi^\psi, \pi^\varphi)
    \le
    \E_{X\sim\rho_0} I(X),
\]
where
\[
    I(x)
    =
    \frac12\int_{\R^d} \big( \pi^\psi(y\mid x) + \pi^\varphi(y\mid x) \big) \big( \delta(y) - c(x) \big)^2 \, \dd y.
\]
For a fixed $x \in \R^d$, let us denote
\[
    a(x) = \E_{\pi^\varphi(\cdot\mid x)}\delta(Y)
    \quad \text{and} \quad
    b(x)=\E_{\pi^\psi(\cdot\mid x)}\delta(Y).
\]
Since
\[
    c(x) = \log\E_{\pi^\varphi(\cdot\mid x)}e^{\delta(Y)},
    \quad \text{and} \quad
    -c(x) = \log\E_{\pi^\psi(\cdot\mid x)}e^{-\delta(Y)},
\]
Jensen's inequality gives $a(x)\le c(x)\le b(x)$. Expanding $I(x)$ and using
\[
    \big(c(x) - a(x)\big)\big(c(x) - b(x) \big) \le 0
    \quad \text{for all $x \in \R^d$,}
\]
we deduce that
\begin{align*}
    I(x)
    &
    = \frac12 \E_{\pi^\varphi(\cdot\mid x)} \delta^2(Y) + \frac12 \E_{\pi^\psi(\cdot\mid x)} \delta^2(Y) - a(x) c(x) - b(x) c(x) + c^2(x)
    \\&
    \leq \frac12 \E_{\pi^\varphi(\cdot\mid x)} \delta^2(Y) + \frac12 \E_{\pi^\psi(\cdot\mid x)} \delta^2(Y) - a(x) b(x).
\end{align*}
Applying the Cauchy-Schwarz bound and Jensen's inequality, we obtain that
\begin{align*}
    I(x)
    &
    \leq \frac12 \E_{\pi^\varphi(\cdot\mid x)} \delta^2(Y) + \frac12 \E_{\pi^\psi(\cdot\mid x)} \delta^2(Y) + \frac{a^2(x)}2 + \frac{b^2(x)}2
    \\&
    \leq \E_{\pi^\varphi(\cdot\mid x)} \delta^2(Y) + \E_{\pi^\psi(\cdot\mid x)} \delta^2(Y).
\end{align*}
Integrating over $X\sim\rho_0$ and using the properties of conditional expectations, we conclude that
\begin{align*}
    \KL(\pi^\psi, \pi^\varphi)
    \le
    \E_{X\sim\rho_0} I(X)
    &
    \leq \E_{X\sim\rho_0} \left( \E_{\pi^\varphi(\cdot\mid X)} \delta^2(Y) + \E_{\pi^\psi(\cdot\mid X)} \delta^2(Y) \right)
    \\&
    = \E_{Y\sim\rho_T^\varphi} \delta^2(Y) + \E_{Y\sim\rho_T^\psi} \delta^2(Y).
\end{align*}
\myendproof

\subsection{Proof of Lemma \ref{lem:potential_smoothness}}
\label{sec:lem_potential_smoothness_proof}

According to the formula \eqref{eq:rho_T_psi},
we have
\[
    \log \rho_T^*(y)
    =
    \varphi^*(y) + H(y),
    \quad \text{where} \quad
    \qquad
    H(y) = \log \int\limits_{\R^d} \frac{\sfq_T(y \,\vert\, x)}{\cT_T \big[e^{\varphi^*} \big](x)} \, \rho_0(x) \, \dd x.
\]
Thus $\varphi^*(y) = \log \rho_T^*(y) - H(y)$. It remains to study properties of $H(y)$. 
Recalling that $\sfq_T(y\,\vert\,x)$ is the density of the Gaussian distribution $\cN\big(m_T(x), \Sigma_T \big)$, where $m_T(x)$ and $\Sigma_T$ are given in \eqref{eq:m_t_sigma_t}, we obtain that $H(y)$ admits a decomposition
\begin{align}
    \label{eq:H}
    H(y)
    &
    = -\frac{d}2 \log(2\pi) - \frac12 \log \det(\Sigma_T) - \frac12 \|\Sigma_T^{-1/2} y\|^2
    \\&\quad
    + \log \int\limits_{\supp(\rho_0)} \exp\left\{ m_T(x)^\top \Sigma_T^{-1} y \right\} \, \dd \mu(x),
\end{align}
where we introduced a non-zero measure
\[
    \dd \mu(x) = \frac{\rho_0(x) e^{-\|\Sigma_T^{-1/2} m_T(x)\|^2 / 2}}{\cT_T \big[e^{\varphi^*} \big](x)} \, \dd x.
\]
Let us denote the last term in the right-hand side of \eqref{eq:H} by $\ell(y)$:
\[
    \ell(y) = \log \int\limits_{\supp(\rho_0)} \exp\left\{ m_T(x)^\top \Sigma_T^{-1} y \right\} \, \dd \mu(x).
\]
Let us denote $\Sigma_T^{-1} m_T(x)$ by $\widetilde m(x) = \big( \widetilde m_1(x), \dots, \widetilde m_d(x) \big)^\top$ for brevity.
Following the same argument as \cite{yakovlev25} in the proof of their Lemma 4.1, we obtain that, for any $s \in \mathbb N$, the $s$-th order derivative $\nabla^s \ell(y)$ admits a representation
\[
    \nabla^s \ell(y) = \frac{\int_{\supp(\rho_0)} \ldots \int_{\supp(\rho_0)} P_s \big(\widetilde m(x_1), \dots, \widetilde m(x_s) \big) \, \dd \widetilde\mu_y(x_1) \dots \dd \widetilde\mu_y(x_s)}{\int_{\supp(\rho_0)} \ldots \int_{\supp(\rho_0)} \dd \widetilde\mu_y(x_1) \dots \dd \widetilde\mu_y(x_s)},
\]
where 
\[
    \dd \widetilde\mu_y(x) = \exp\left\{ y^\top \widetilde m_T(x) \right\} \, \dd \mu(x)
\]
and $P_s$ is the $s$-th order tensor defined by the following recurrent relations: $P_1(u_1) = u_1$,
\[
    P_{k + 1}(u_1, \dots, u_{k + 1}) = P_k(u_1, \dots, u_k) \otimes (u_1 + \ldots + u_k - k u_{k + 1})
    \quad \text{for all $k \in \mathbb N$.}
\]
Let us note that for every $x_1, \dots, x_s \in \supp(\rho_0)$ the operator norm of $P_s \big(\widetilde m(x_1), \dots, \widetilde m(x_s) \big)$ does not exceed
\begin{align*}
    &
    \left\| P_s \big(\widetilde m(x_1), \dots, \widetilde m(x_s) \big) \right\|
    \\&
    \leq \left\|P_{s - 1} \big(\widetilde m(x_1), \dots, \widetilde m(x_{s - 1}) \big) \right\| \, \left\| \widetilde m(x_1) + \ldots + \widetilde m(x_{s - 1}) - (s - 1) \widetilde m(x_s) \right\|
    \\&
    \leq \left\|P_{s - 1} \big(\widetilde m(x_1), \dots, \widetilde m(x_{s - 1}) \big) \right\| \, \left\| \widetilde m(x_1) + \ldots + \widetilde m(x_{s - 1}) - (s - 1) \widetilde m(x_s) \right\|
    \\&
    \leq 2(s - 1) \left\|P_{s - 1} \big(\widetilde m(x_1), \dots, \widetilde m(x_{s - 1}) \big) \right\| \sup\limits_{x \in \supp(\rho_0)} \|\widetilde m_T(x)\|.  
\end{align*}
Repeating the same argument $(s - 2)$ times and recalling that $P_1\big(\widetilde m_1(x_1) \big) = \widetilde m_1(x_1)$ by the definition, we conclude that
\[
    \left\| P_s \big(\widetilde m(x_1), \dots, \widetilde m(x_s) \big) \right\|
    \leq 2^{s - 1} (s - 1)! \sup\limits_{x \in \supp(\rho_0)} \|\widetilde m_T(x)\|^{s}.
\]    
Since the vector $\widetilde m_T(x) = \Sigma_T^{-1} m_T(x)$ is bounded on $\supp(\rho_0)$ and, due to the definition of $m_T(x)$ and $\Sigma_T$ (see \eqref{eq:m_t_sigma_t}),
\[
    \|\widetilde m_T(x)\|
    \leq \left\|\Sigma_T^{-1} m \right\| + e^{-bT} \left\| \Sigma_T^{-1} (x - m) \right\|
    \leq \frac{1}{\sigma_T^2} \left\|\Sigma^{-1} m \right\| + \frac{e^{-bT} R}{\sigma_T^2} \left\| \Sigma^{-1/2} \right\| 
\]
for all $x \in \supp(\rho_0)$, the operator norm of $P_s \big(\widetilde m(x_1), \dots, \widetilde m(x_s) \big)$ satisfies the inequality
\[
    \left\| P_s \big(\widetilde m(x_1), \dots, \widetilde m(x_s) \big) \right\| \leq \frac{2^{s - 1} (s - 1)!}{\sigma_T^{2s}} \left( \left\|\Sigma^{-1} m \right\| + e^{-bT} R \left\| \Sigma^{-1/2} \right\| \right)^s.
\]
This yields that
\[
    |\partial^\alpha \ell(y)| \leq  \frac{2^{s - 1} (s - 1)!}{\sigma_T^{2s}} \left( \left\|\Sigma^{-1} m \right\| + e^{-bT} R \left\| \Sigma^{-1/2} \right\| \right)^s
\]
for any multi-index $\alpha$ of order $s$ and any $y \in \R^d$. As a consequence, it holds that $H\in C^\infty$ and all the derivatives $\partial^{\alpha} H$ for $|\alpha| \ge 2$ are globally bounded. In addition, $H$ and $\nabla H$ have at most quadratic and linear growth, respectively, that is, $\|H(y)\| \lesssim r^2$ for all $\|y\| \leq r$, $r\|\partial^\alpha H(y)\| \lesssim r^2$ for $|\alpha| = 1$, $\|y\| \leq r$, and the global boundedness yields
\[
    \max_{\|y\| \leq r} \left\{ r^{|\alpha|} |\partial^\alpha H(y)| \right\}
    \lesssim r^s 
    \quad \text{for all $2\le |\alpha|\le s$.}
\]
Therefore,
\[
    \max_{|\alpha|\le s} \, \max_{\|y\| \leq r} \left\{ r^{|\alpha|} |\partial^\alpha H(y)| \right\}
    \lesssim
    r^{s \vee 2},
\]
and the displayed bound for $A_r(\varphi^*)$ follows.

\myendproof

\subsection{Proof of Lemma \ref{prop:schmidt_hieber_compact_approx}}
\label{sec:prop_schmidt_hieber_compact_approx_proof}

Let us fix an arbitrary $f\in C^s_{\mathrm{loc}}(\R^d)$.
If $A_r(f) = 0$, then $f = 0$ on the ball $\cB(0, r)$, and the network with zero weights gives the result. Otherwise, let us set $\bar f_r(u) = f(ru) / A$, $\|u\| \leq 1$. Since $A\ge A_r(f)$, we have
\[
    \max_{|\alpha|\le s} \; \max_{\|u\| \leq 1} \left| \partial^\alpha \bar f_r(u) \right| \le 1.
\]
Due to Stein's extension theorem (see, for instance, Theorem 5.24 in \cite{adams03}), we can extend $\bar f_r$ to $[-1,1]^d$ by a bounded operator
\[
    \mathcal E:W^{s,\infty}(B_1)\to W^{s,\infty}([-1,1]^d).
\]
For $v\in[0,1]^d$, let us define $F_r(v)=(\mathcal E\bar f_r)(2v-\mathbf 1_d)$, where $\mathbf 1_d = (1, \dots, 1)^\top \in \R^d$. Then $\|F_r\|_{W^{s,\infty}([0,1]^d)} \le C_{s,d}$ for some $C_{s,d} > 0$ depending only on $s$ and $d$. According to Theorem 5 from \cite{schmidt-hieber20}, for each $N \geq 2$ there exist $K_N \lesssim \log N$, $\widetilde S_N \lesssim N \log N$, and an architecture $\widetilde W_N \in \N^{K_N + 1}$ with $\|\widetilde W_N\|_\infty \lesssim N \log N$, such that the class $\NN(K_N, \widetilde W_N, \widetilde S_N, 1)$, defined in \eqref{eq:nn_class}, contains a neural network $h_N$ satisfying
\[
    \sup_{v\in[0,1]^d} |h_N(v) - F_r(v)|
    \lesssim N^{-s/d},
\]
where the hidden constant depends on $s$ and $d$.
Setting $g_N(x) = A h_N(0.5x / r + 0.5\mathbf 1_d)$, $\|x\| \leq r$, and using the fact that
\[
    f(x) = A F_r\left( \frac{x}{2r} + \frac12 \mathbf 1_d\right)
    \quad \text{for all $\|x\| \leq r$,}
\]
we obtain that
\[
    \sup_{\|x\| \leq r} |g_N(x) - f(x)|
    \lesssim A N^{-s/d}.
\]
It remains to note that
\[
    g_N(x)
    = A h_N\left( \frac{x}{2r} + \frac12 \mathbf 1_d \right)
    = A h_N\left( \relu_{-0.5 \mathbf 1_d}\left( 0.5 x / r \right) \right),
    \quad x \in [-r , r]^d
\]
that is, $g_N$ is a neural network of depth $(K_N + 1)$ with the architecture $W_N = (d, \widetilde W_N)$, $S_N \leq \widetilde S_N + 2d \lesssim N \log N$ non-zero weights and its weight magnitude does not exceed $3/2 \vee A$.

\myendproof

\section{Auxiliary results}

This section contains auxiliary results used in the proofs of Lemma \ref{lem:kl_bound} and Lemma \ref{lem:log_ou_bound}. The first one is a Gronwall-type inequality helping us to relate the operators $\cT_t$ and $\cT_\infty$ (see Lemma \ref{lem:log_ou_bound}).

\begin{Lem}
    \label{lem:ode_bound}
    Let $\phi : (0, +\infty) \rightarrow \R_+$ be a non-negative differentiable function such that there exists
    \[
        \lim\limits_{t \rightarrow +\infty} \phi(t) = \phi(+\infty) \in \R.
    \]
    Let the functions $a(t)$ and $\varkappa(t)$ take non-negative values on $(0, +\infty)$ and assume that the integrals
    \[
        \int\limits_t^{+\infty} a(s) \, \dd s
        \quad \text{and} \quad
        \int\limits_t^{+\infty} \varkappa(s) \, \dd s
    \]
    are finite for any $t > 0$.
    If
    \[
        -\frac{\dd \phi(t)}{\dd t} \leq a(t) + \varkappa(t) \phi(t) \quad \text{for all $t > 0$,}
    \]
    then
    \[
        \phi(t)
        \leq \left( \phi(+\infty) + \int\limits_t^{+\infty} a(s) \, \dd s \right)  \exp\left\{ \int\limits_t^{+\infty} \varkappa(\tau) \, \dd \tau \right\}
        \quad \text{for any $t > 0$.}
    \]
    Otherwise, if
    \[
        \frac{\dd \phi(t)}{\dd t} \leq a(t) + \varkappa(t) \phi(t) \quad \text{for all $t > 0$,}
    \]
    then it holds that
    \[
        \phi(t) \geq \phi(+\infty) \exp\left\{ -\int\limits_t^{+\infty} \varkappa(\tau) \, \dd \tau \right\} - \int\limits_t^{+\infty} a(s) \, \dd s
        \quad \text{for any $t > 0$.}
    \]
\end{Lem}
The proof of Lemma \ref{lem:ode_bound} is quite similar to the one of the original Gronwall lemma. Nevertheless, we provide all the derivations in Appendix \ref{sec:lem_ode_bound_proof}. We move to the next auxiliary result.

Finally, we present a sharp bound on $L_p$-norm of a centered chi-squared random variable.

\begin{Lem}
    \label{lem:chi-squared_moments_bound}
    Let $\xi \sim \cN(0, I_d)$ be a Gaussian vector in $\R^d$. Then, for any $p \geq 1$, it holds that
    \[
        \left( \E \left|\|\xi\|^2 - d\right|^p \right)^{1/p}
        \leq 10p \sqrt{d}.
    \]
\end{Lem}

The proof of Lemma \ref{lem:chi-squared_moments_bound} is moved to Appendix \ref{sec:lem_chi-squared_moments_bound_proof}. Let us note that, unlike the $L_p$-norm of $\|\xi\|^2$, $\xi \sim \cN(0, I_d)$, which is of order $\Omega(d)$, the $L_p$-norm of $\|\xi\|^2 - d$ is much smaller and grows as fast as $\cO(\sqrt{d})$. We conclude this section with the proof of an auxiliary result, which helps us to bound the local metric entropy of the excess loss class.

\begin{Lem}
    \label{lem:log_Dn_bound}
    Let $D_n \geq 1$ be as defined in \eqref{eq:Dn}. Then, for any $p \geq 1$, it holds that
    \begin{align*}
        \left( \E \log^p(2D_n) \right)^{1 / p}
        &
        \leq 2^{1 / p} (p + 1) \left( 1 \vee \log_2(2 \E D_n) \right)
        \\&
        \lesssim p \left( 1 + \log \Lambda + \log \cK(T) + \log(1 + d + R^2 e^{-2bT} + d \sigma_T^2) \right)
        \\&\quad
        + p \log\left( 1 + M + \frac{R^2 e^{-bT}}{\sqrt d} + \log \cK(T) \right),
    \end{align*}
    where the function $\cK(t)$ is given by \eqref{eq:cK}.
\end{Lem}

\begin{Lem}
    \label{lem:integral_remainder_bound}
    Let $\xi$ be a random vector in $\R^d$ with the density $\rho$. Assume that there are $\mu \in \R^d$ and $\ttv > 0$ such that
    \[
        \E e^{u^\top \xi} \leq \exp\left\{ \mu^\top u + \frac{\ttv^2 \|u\|^2}2 \right\}.
    \]
    Then, for any $\alpha > 0$ and any $r \geq \ttv \sqrt{2 \alpha (d + 1)}$, it holds that
    \[
        \int\limits_{\|x\| \geq r} \|x\|^\alpha \, \rho(x) \, \dd x
        \leq r^\alpha \exp\left\{-\frac{r^2}{4 \ttv^2 (d + 1)} + \frac{\|\mu\|^2}{2 \ttv^2 (d + 1)} + \frac14 \right\}.
    \]
\end{Lem}

\subsection{Proof of Lemma \ref{lem:ode_bound}}
\label{sec:lem_ode_bound_proof}

We perform the proof in two steps starting with the upper bound.

\smallskip

\noindent
\textbf{Step 1: upper bound.}
\quad
Let us introduce
\[
    \Psi(t) = \phi(t) \exp\left\{ -\int\limits_t^{+\infty} \varkappa(s) \, \dd s \right\},
    \quad t > 0.
\]
Then it is easy to observe that for any $t > 0$ the derivative of $\varphi$ satisfies the inequality
\begin{align*}
    \frac{\dd \Psi(t)}{\dd t}
    &
    = \varkappa(t) \phi(t) \exp\left\{ -\int\limits_t^{+\infty} \varkappa(s) \, \dd s \right\} + \frac{\dd \phi(t)}{\dd t} \exp\left\{ -\int\limits_t^{+\infty} \varkappa(s) \, \dd s \right\}
    \\&
    \geq -a(t) \exp\left\{ -\int\limits_t^{+\infty} \varkappa(s) \, \dd s \right\}.
\end{align*}
Applying the Newton-Leibniz formula, we obtain that
\[
    \Psi(t) - \Psi(+\infty)
    = -\int\limits_t^{+\infty} \frac{\dd \Psi(s)}{\dd s} \, \dd s
    \leq \int\limits_t^{+\infty} a(s) \exp\left\{ -\int\limits_s^{+\infty} \varkappa(\tau) \, \dd \tau \right\} \, \dd s.
\]
This yields that
\[
    \phi(t) \exp\left\{ -\int\limits_t^{+\infty} \varkappa(\tau) \, \dd \tau \right\}
    \leq \phi(+\infty) + \int\limits_t^{+\infty} a(s) \exp\left\{ -\int\limits_s^{+\infty} \varkappa(\tau) \, \dd \tau \right\} \, \dd s.
\]
Taking into account that
\[
    \exp\left\{ \int\limits_t^{s} \varkappa(\tau) \, \dd \tau \right\}
    \leq \exp\left\{ \int\limits_t^{+\infty} \varkappa(\tau) \, \dd \tau \right\}
    \quad \text{for all $s \geq t$,}
\]
we finally deduce
\begin{align*}
    \phi(t)
    &
    \leq \phi(+\infty) \exp\left\{\int\limits_t^{+\infty} \varkappa(\tau) \, \dd \tau \right\} + \int\limits_t^{+\infty} a(s) \exp\left\{ \int\limits_t^s \varkappa(\tau) \, \dd \tau \right\} \, \dd s
    \\&
    \leq \phi(+\infty) \exp\left\{\int\limits_t^{+\infty} \varkappa(\tau) \, \dd \tau \right\} + \int\limits_t^{+\infty} a(s) \exp\left\{ \int\limits_t^{+\infty} \varkappa(\tau) \, \dd \tau \right\} \, \dd s
    \\&
    = \left(\phi(+\infty) + \int\limits_t^{+\infty} a(s) \, \dd s \right) \exp\left\{ \int\limits_t^{+\infty} \varkappa(\tau) \, \dd \tau \right\}.
\end{align*}

\smallskip

\noindent
\textbf{Step 2: lower bound.}
\quad
The proof of the lower bound is quite similar. The only difference is that we have to replace $\Psi(t)$ by
\[
    \chi(t) = \phi(t) \exp\left\{ \int\limits_t^{+\infty} \varkappa(s) \, \dd s \right\},
    \quad t > 0.
\]
Then the derivative of $\chi$ obeys the inequality
\begin{align*}
    \frac{\dd \chi(t)}{\dd t}
    &
    = -\varkappa(t) \phi(t) \exp\left\{ \int\limits_t^{+\infty} \varkappa(s) \, \dd s \right\} + \frac{\dd \phi(t)}{\dd t} \exp\left\{ \int\limits_t^{+\infty} \varkappa(s) \, \dd s \right\}
    \\&
    \leq a(t) \exp\left\{ \int\limits_t^{+\infty} \varkappa(s) \, \dd s \right\}.
\end{align*}
According to the Newton-Leibniz formula, it holds that
\[
    \chi(+\infty) - \chi(t)
    = \int\limits_t^{+\infty} \frac{\dd \chi(s)}{\dd s} \, \dd s
    \leq \int\limits_t^{+\infty} a(s) \exp\left\{ \int\limits_s^{+\infty} \varkappa(\tau) \, \dd \tau \right\} \, \dd s.
\]
This implies that
\[
    \phi(t) \exp\left\{ \int\limits_t^{+\infty} \varkappa(\tau) \, \dd \tau \right\}
    \geq \phi(+\infty) - \int\limits_t^{+\infty} a(s) \exp\left\{ \int\limits_s^{+\infty} \varkappa(\tau) \, \dd \tau \right\} \, \dd s.
\]
Since the function $\varkappa$ is non-negative, 
we obtain the desired bound:
\begin{align*}
    \phi(t)
    &
    \geq \phi(+\infty) \exp\left\{-\int\limits_t^{+\infty} \varkappa(\tau) \, \dd \tau \right\} - \int\limits_t^{+\infty} a(s) \exp\left\{ -\int\limits_t^s \varkappa(\tau) \, \dd \tau \right\} \, \dd s
    \\&
    \geq \phi(+\infty) \exp\left\{-\int\limits_t^{+\infty} \varkappa(\tau) \, \dd \tau \right\} - \int\limits_t^{+\infty} a(s) \, \dd s.
\end{align*}
\myendproof

\subsection{Proof of Lemma \ref{lem:chi-squared_moments_bound}}
\label{sec:lem_chi-squared_moments_bound_proof}

First, let us fix $\lambda \in (0, 1/2)$ and consider the exponential moment $\E e^{\lambda |\|\xi\|^2 - d|}$. Using explicit expressions for moment generating functions of the chi-squared distribution with $d$ degrees of freedom, we obtain that
\[
    \E e^{\lambda |\|\xi\|^2 - d|}
    \leq \E e^{\lambda \|\xi\|^2 - \lambda d} + \E e^{\lambda d - \lambda \|\xi\|^2}
    = \frac{e^{-\lambda d}}{(1 - 2\lambda)^{d/2}} + \frac{e^{\lambda d}}{(1 + 2\lambda)^{d/2}}.
\]
Let us note that
\[
    \frac{\dd^2}{\dd x^2} \left( -\frac12 \log(1 - 2x)\right) = \frac{2}{(1 - 2x)^2} \leq 8
\]
and
\[
    \frac{\dd^2}{\dd x^2} \left( -\frac12 \log(1 + 2x)\right) = \frac{2}{(1 + 2x)^2} \leq 2
\]
for all $0 \leq x \leq 1/4$. This and Taylor's expansion with a Lagrange remainder term imply that
\[
    -\frac12 \log(1 - 2x) \leq x + 4x^2 
    \quad \text{and} \quad
    -\frac12 \log(1 + 2x) \leq -x + x^2
    \quad \text{for all $0 \leq x \leq 1/4$.}
\]
Thus, we obtain that
\[
    \E e^{\lambda |\|\xi\|^2 - d|}
    \leq \frac{e^{-\lambda d}}{(1 - 2\lambda)^{d/2}} + \frac{e^{\lambda d}}{(1 + 2\lambda)^{d/2}}
    \leq e^{4\lambda^2 d} + e^{\lambda^2 d}
    \quad \text{for all $0 < \lambda \leq 1/4$.}
\]
We apply this inequality to bound the $L_p$-norm of $\|\xi\|^2 - d$ using a standard technique. To be more precise, for any $0 < \lambda \leq 1/4$, it holds that
\begin{align*}
    \E \left|\|\xi\|^2 - d\right|^p
    &
    = \int\limits_0^{+\infty} \p\left( \left|\|\xi\|^2 - d\right|^p \geq u \right) \, \dd u
    \\&
    \leq \int\limits_0^{+\infty} e^{-\lambda u^{1/p}} \E e^{\lambda |\|\xi\|^2 - d|} \, \dd u
    \\&
    \leq \left( e^{4\lambda^2 d} + e^{\lambda^2 d} \right) \int\limits_0^{+\infty} e^{-\lambda u^{1/p}} \, \dd u.
\end{align*}
Let us take $\lambda = 1 / (4 \sqrt{d})$. Then, substituting $u^{1/p} / (4 \sqrt{d})$ by $v$, we obtain that
\begin{align*}
    \E \left|\|\xi\|^2 - d\right|^p
    &
    \leq \left( e^{1/4} + e^{1/16} \right) \int\limits_0^{+\infty} e^{-\lambda u^{1/p}} \, \dd u
    \\&
    = (4 \sqrt{d})^p \left( e^{1/4} + e^{1/16} \right) \int\limits_0^{+\infty} p \, v^{p - 1} e^{-v} \, \dd v
    \\&
    = (4 \sqrt{d})^p \left( e^{1/4} + e^{1/16} \right) \cdot p \, \Gamma(p)
    \\&
    = (4 \sqrt{d})^p \left( e^{1/4} + e^{1/16} \right) \Gamma(p + 1).
\end{align*}
Since $e^{1/4} + e^{1/16} \leq 5/2$ and $\Gamma(p + 1) \leq p^p$ for all $p \geq 1$, the expression in the right-hand side does not exceed
\[
    (4 \sqrt{d})^p \left( e^{1/4} + e^{1/16} \right) \Gamma(p + 1)
    \leq \frac52 \cdot (4p \sqrt{d})^p
    \leq (10p \sqrt{d})^p
    \quad \text{for all $p \geq 1$.}
\]
The proof is finished.

\myendproof

\subsection{Proof of Lemma \ref{lem:log_Dn_bound}}
\label{sec:lem_log_Dn_bound_proof}

Let us take $t = 1 \vee \log_2(2\E D_n)$ and consider
\[
    \E \exp\left\{\frac{|\log D_n|}t \right\}.
\]
Since $D_n \geq 1$, it holds that
\[
    \E \exp\left\{ \frac{|\log(2 D_n)|}t \right\}
    = \E \exp\left\{ \frac{\log(2 D_n)}t \right\}
    = \E (2 D_n)^{1/t}.
\]
Due to Jensen's inequality, we have
\[
    \E \exp\left\{ \frac{|\log(2 D_n)|}t \right\}
    \leq \big( 2 \E D_n) \big)^{1/t} \leq 2.  
\]
This yields that $\| \log(2 D_n) \|_{\psi_1} \leq 1 \vee \log_2(2 \E D_n)$. Then, according to the properties of sub-exponential random variables (see, for instance, Lemma E.8 in \cite{puchkin26}), it holds that
\[
    \E \log^p(2D_n)
    \leq 2 (p + 1)^p \| \log(2 D_n) \|_{\psi_1}^p
    \leq 2 (p + 1)^p \left( 1 \vee \log_2(2 \E D_n) \right)^p.
\]
Note that, due to Assumption \ref{as:sub_gaussian_density}, we have
\[
    \E \; \frac1n \sum\limits_{i = 1}^n \left\| \Sigma^{-1/2} (Y_i - m) \right\|^4
    \lesssim \ttv^4 d^2 \lesssim d^2.
\]
This implies that
\[
    \E D_n^2
    \lesssim 4\Lambda^2 (1 + d^2)
    + 8e^2 \Lambda^2 \left( 1 + 2 e^{-2bT} R^2 + 2(d + 2) \sigma_T^2 \right)^2
    \Big[ 2 + \big(M + \cA_{\max}(T)\big) \cK(T) \Big]^2.
\]
Hence, we obtain that
\begin{align*}
    \left( \E \log^p(2D_n) \right)^{1/p}
    &
    \lesssim p \left( 1 + \log \Lambda + \log \cK(T) + \log(1 + d + R^2 e^{-2bT} + d \sigma_T^2) \right)
    \\&\quad
    + p \log\left( 1 + M + \frac{R^2 e^{-bT}}{\sqrt d} + \log \cK(T) \right).
\end{align*}
\myendproof

\subsection{Proof of Lemma \ref{lem:integral_remainder_bound}}
\label{sec:lem_integral remainder_bound_proof}

Let us take $\lambda = \big(2 \ttv^2 (d + 1) \big)^{-1}$ and note that
\[
    \int\limits_{\|x\| \geq r} \|x\|^\alpha \, \rho(x) \, \dd x
    = \int\limits_{\|x\| \geq r} \|x\|^\alpha e^{-\lambda \|x\|^2 / 2} \cdot e^{\lambda \|x\|^2 / 2} \, \rho(x) \, \dd x
    \leq M_r \int_{\R^d} e^{\lambda \|x\|^2 / 2} \, \rho(x) \, \dd x,
\]
where
\[
    M_r = \max\limits_{\|x\| \geq r} \left\{ \|x\|^\alpha e^{-\lambda \|x\|^2 / 2} \right\}.
\]
Since the function $u \mapsto u^{\alpha / 2} e^{-\lambda u / 2}$ is non-increasing whenever $u \geq \alpha / \lambda = 2 (d + 1) \alpha \ttv^2$, we obtain that
\[
    M_r = r^\alpha e^{-\lambda r^2 / 2} \quad \text{for all $r \geq \ttv \sqrt{2\alpha (d + 1)}$.}
\]
Hence, for any $r \geq \ttv \sqrt{2\alpha (d + 1)}$, it holds that
\[
    \int\limits_{\|x\| \geq r} \|x\|^\alpha \, \rho(x) \, \dd x
    \leq r^\alpha e^{-\lambda r^2 / 2} \int_{\R^d} e^{\lambda \|x\|^2 / 2} \, \rho(x) \, \dd x,
\]
and it remains to bound the integral in the right-hand side. For this purpose, let us introduce a Gaussian random vector $\gamma \sim \cN(0, I_d)$, which is independent of $\xi$ and note that
\[
    \int_{\R^d} e^{\lambda \|x\|^2 / 2} \, \rho(x) \, \dd x
    = \E e^{\lambda \|\xi\|^2 / 2} = \E e^{\sqrt{\lambda} \, \gamma^\top \xi}.
\]
Due to the conditions of the lemma, $\xi$ is sub-Gaussian. This yields that
\[
    \E e^{\sqrt{\lambda} \, \gamma^\top \xi}
    \leq \E \exp\left\{ \sqrt{\lambda} \gamma^\top \mu + \frac{\lambda \ttv^2 \|\gamma\|^2}2 \right\}
    \leq \sqrt{\E \exp\left\{2 \sqrt{\lambda} \gamma^\top \mu\right\} } \cdot \sqrt{\E e^{\lambda \ttv^2 \|\gamma\|^2}}.
\]
The exponential moments in the right-hand side can be computed explicitly. Substituting
\[
    \E \exp\left\{2 \sqrt{\lambda} \gamma^\top \mu\right\}
    = e^{2 \lambda \|\mu\|^2}
    \quad \text{and} \quad
    \E e^{\lambda \ttv^2 \|\gamma\|^2}
    = (1 - 2 \lambda \ttv^2)^{-d/2}
    = \left(1 + \frac1d \right)^{d/2}
    \leq \sqrt{e}
\]
into the expression above, we obtain that
\[
    \int_{\R^d} e^{\lambda \|x\|^2 / 2} \, \rho(x) \, \dd x
    = \E e^{\sqrt{\lambda} \, \gamma^\top \xi}
    \leq e^{1/4 + \lambda \|\mu\|^2}.
\]
Then, recalling the definition of $\lambda$, we deduce that
\[
    \int\limits_{\|x\| \geq r} \|x\|^\alpha \, \rho(x) \, \dd x
    \leq r^\alpha \exp\left\{-\frac{r^2}{4 \ttv^2 (d + 1)} + \frac{\|\mu\|^2}{2 \ttv^2 (d + 1)} + \frac14 \right\}
\]
for all $r \geq \ttv \sqrt{2 \alpha (d + 1)}$.

\myendproof

\clearpage
\tableofcontents

\end{appendix}
\end{document}